\documentclass[10pt,twocolumn,letterpaper]{article}

\usepackage{cvpr}              

\usepackage[dvipsnames]{xcolor}

\usepackage{subcaption}

\definecolor{cvprblue}{rgb}{0.21,0.49,0.74}
\usepackage[pagebackref,breaklinks,colorlinks,citecolor=cvprblue]{hyperref}
\usepackage{color}
\usepackage{tabularray}
\usepackage{pgfplots}
\usetikzlibrary{shapes.geometric}
\pgfplotsset{compat=newest}
\pgfdeclareplotmark{mystar}{
    \node[star,star point ratio=2.25,minimum size=6pt,
          inner sep=0pt,draw=black,solid,fill=yellow] {};
}
\usepackage{pgfplotstable}
\usepackage{float}
\usepackage{placeins}
\usepackage[moderate]{savetrees}
\pgfplotsset{compat=1.18}

\title{Toward Human Understanding with Controllable Synthesis}

\author{Hanz Cuevas-Velasquez$^{1}$ \,\,\, Priyanka Patel$^{2}$,\footnotemark  \,\,\,\, Haiwen Feng$^{1}$ \,\,\, Michael Black$^{1}$\\
	$^1$ Max Planck Institute for Intelligent Systems  \qquad $^2$ Meshcapade \qquad  \\
	\tt\small \{hanz.cuevas, haiwen.feng, black\}@tuebingen.mpg.de \,\,\, priyanka@meshcapade.com \vspace*{1.5ex}
}
\begin{document}
\newcommand{\teaserCaption}{
Toward Human Understanding with Controllable Synthesis
}

\twocolumn[{
    \renewcommand\twocolumn[1][]{#1}
    \maketitle
    \centering
    \vspace{-0.5em}
    \begin{minipage}{0.85\textwidth}
        \centering
        \includegraphics[trim=000mm 000mm 000mm 000mm, clip=False, width=\linewidth]{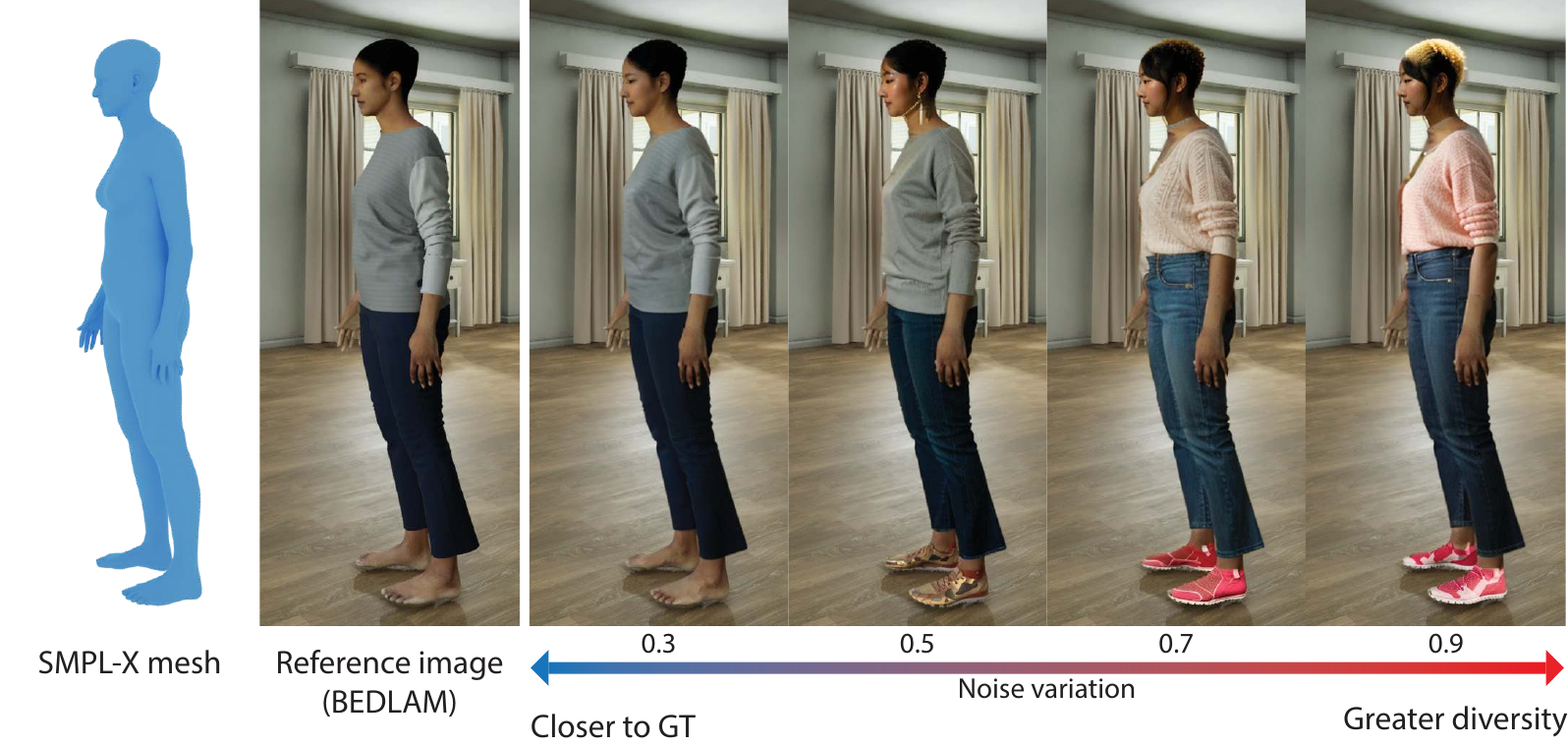}

    \end{minipage}
    \vspace{-0.5 em}
    \captionof{figure}{{\bf Generative BEDLAM (Gen-B)} is a dataset that takes traditionally rendered images with perfect ground truth 3D body shape and pose information and ``upgrades" their realism using a generative diffusion processes that remains faithful to the ground truth. 
    Specifically, we upgrade BEDLAM, a large-scale synthetic video dataset designed to train and test algorithms on the task of 3D human pose and shape estimation. 
    This is challenging because image generation methods produce realistic images, the resulting images may deviate from the ground truth, making them unusable for training or evaluation.
    To address this, we use metadata provided by BEDLAM to control the generative process. 
    Depending on the noise added during the diffusion step, we can produce more realistic images that preserve the pose and shape of the person. 
    We show, for the first time, that such a generative approach produces a training dataset that improves the accuracy of 3D human pose and shape estimation.
    }
    \label{fig:teaser}
    \vspace{2.2em}
}]
\maketitle
\begin{abstract}
Training methods to perform robust 3D human pose and shape (HPS) estimation requires diverse training images with accurate ground truth. 
While BEDLAM demonstrates the potential of traditional procedural graphics to generate such data, the training images are clearly synthetic. 
In contrast, generative image models produce highly realistic images but without ground truth.
Putting these methods together seems straightforward: use a generative model with the body ground truth as controlling signal. 
However, we find that, the more realistic the generated images, the more they deviate from the ground truth, making them inappropriate for training and evaluation.
Enhancements of realistic details, such as clothing and facial expressions, can lead to subtle yet significant deviations from the ground truth, potentially misleading training models. 
We empirically verify that this misalignment causes the accuracy of HPS networks to decline when trained with generated images. 
To address this, we design a controllable synthesis method that effectively balances image realism with precise ground truth.
We use this to create the Generative BEDLAM (Gen-B) dataset, which improves the realism of the existing synthetic BEDLAM dataset while preserving ground truth accuracy. 
We perform extensive experiments, with various noise-conditioning strategies, to evaluate the tradeoff between visual realism and HPS accuracy.
We show, for the first time, that generative image models can be controlled by traditional graphics methods to produce training data that increases the accuracy of HPS methods.
The code and dataset will be available for research purposes.

\end{abstract}

\renewcommand*{\thefootnote}{\fnsymbol{footnote}}
\footnotetext{$^*$This work was performed while Priyanka Patel was at the MPI for Intelligent Systems.}
\renewcommand*{\thefootnote}{\arabic{footnote}}
    
\section{Introduction}
\label{sec:intro}

Training neural networks to accurately estimate 3D human pose and shape requires large amounts of accurate training data.
Ideally, this data includes paired images with ground truth (GT) pose and shape parameters, e.g.~parameters of a parametric body model like SMPL~\cite{loper2015smpl}.
The diversity of the images and poses are critical for methods to generalize.
Capturing real images with ground truth data is hard, particularly with complex poses, so many methods rely on pseudo ground truth estimated from 2D data.
This can introduce errors and bias.
An alternative is synthetic data but, until recently, the domain gap between real and synthetic images meant that such data was insufficient to train networks without also using real data.
The BEDLAM dataset~\cite{black2023bedlam} demonstrated that traditional graphics methods can be used to produce images with ground truth data of sufficient complexity and realism that training only on the synthetic data produces state-of-the-art (SOTA) performance.
Despite this, images in the BEDLAM dataset are still very obviously synthetic.

In contrast, recent generative image models produce highly realistic images of people that are often indistinguishable from real images.
Generative models can synthesize an effectively infinite variation in clothing, hair, lighting, skin tone, etc.
To date, however, such imagery \cite{rombach2022high, ruiz2023dreambooth} could not be directly use for HPS methods because it lacks aligned ground truth 3D body shape and pose parameters.
The question then arises whether generative image methods can be leveraged to make synthetic training datasets with accurate ground truth and realistic imagery.

There have been several attempts \cite{weng2024diffusion, ge2024d} to obtain a pseudo-ground truth label from the generated images using ControlNet~\cite{zhang2023adding} and depth information to provide some control over the human body. However, they still do not manage to beat the BEDLAM dataset. This shows that controllable synthesis has the potential to produce highly realistic images, however, there is something missing to constrain them to retain accurate ground truth information. In this work, we combine both, generative models with traditional rendered images to enhance their photorealism while maintaining the accuracy of the ground truth pose and shape data.

This accomplishes what is desired in terms of reducing the visual domain gap between synthetic data and real images.
Unfortunately, it introduces a new kind of ``gap'' between the generated images and the ground truth pose and shape.
Making the images realistic changes them in ways that then deviate from the ground truth—the more realistic the imagery, the greater this gap, see \cref{fig:teaser} and \cref{fig:noise_examples}.
The changes can be subtle to the human eye, though clothing often changes in ways that imply a different body shape underneath.
We verify this effect quantitatively by evaluating HPS methods trained on the synthetic imagery and observe that as visual realism of the training data increases, the performance of networks trained on the data decreases.

What we seek at the end is a method to produce a strong {\em alignment} between visually realistic synthetic images and the ground truth.
Here we explore several methods to improve this alignment.
In particular, we condition the generative process on 2D keypoints, rendered depth maps, rendered surface normals, and image edges.
We use Stable Diffusion model~\cite{rombach2022high} and add varying amounts of noise to the original BEDLAM images.  
We explore using different noise levels for different parts of the body. Our pipeline changes things like clothing, without impacting the body shape.
We increase the realism of the faces and add hair and shoes, which are missing or often missing in BEDLAM.
Through these experiments, we find our pipeline increases realism while staying aligned with the ground truth, see Fig.~\ref{fig:control_differences}. The result is a new {\em Generative BEDLAM (Gen-B)} dataset that can be used to train HPS methods.

In addition to our pipeline and dataset, a key contribution of our work is that it explores and exposes the issues involved in using generative image models to create training data for HPS.
We expose the alignment issue as equally important as the well-known visual domain gap.
While it may seem obvious a priori that improving image realism is a useful goal, our work suggests caution.
Many HPS methods today use small image crops to analyze human pose and shape.
At low resolution, improvements in image realism may not be significant.
In the end, one wants an HPS method to be invariant to many things including unrealistic images.

Finally, we extensively evaluate state-of-the-art HPS networks to test how well our dataset generalizes to real-world images compared to BEDLAM. We train HMR~\cite{kanazawa2018end} with the original BEDLAM dataset and Gen-B. Our results show that training on Gen-B improves accuracy on real test datasets—3DPW, HMDB, and RICH—by $2.37\%$, $4.66\%$, and $1.95\%$, respectively. Similarly, when training CLIFF~\cite{li2022cliff}, Gen-B consistently outperforms BEDLAM, achieving lower error across all datasets. This consistent advantage extends to newer architectures like transformer-based ones, where we observe that HMR2.0~\cite{goel2023humans} trained on Gen-B show a reduction of error of $2.26\%$ on the RICH dataset and keep being marginally better on the 3DPW dataset. 

In summary, we 
(1) provide a process to upgrade synthetic data like BEDLAM using generative models while preserving alignment with the ground truth body shape and pose;
(2) provide a new dataset, Gen-B, that can be used to train HPS regressors to achieve SOTA accuracy;
(3) extensively evaluate existing HPS methods trained on Gen-B,
(4) explore and discuss how narrowing the domain gap is not enough and that alignment between the images and the ground truth is key.
The dataset, code to generate the dataset, trained models, and model training code will be available for research purposes.

\section{Related Work}
\label{sec:related_work}

There are three primary methods for generating synthetic data for HPS regressors: (1) Images created through a rendering process with perfect ground truth but low visual realism; (2) Images generated by a generative model with high realism but noisy ground truth; and (3) Real images that are augmented to increase diversity but are limited in pose variability. The ideal training data would combine the realism of (2) with the perfect ground truth of (1) to achieve high diversity and accuracy. None of the existing methods fully meets this goal. Below, we describe these approaches in more detail and provide an overview of real datasets.

\paragraph{Real data.} Real images offer complexity and diversity. Most data label 2D keypoints which are easy to be manually labeled at scale \cite{andriluka20142d, iqbal2017posetrack, martin2021jrdb}. However, as the data relies on human annotators, they are prone to error and might not be consistent. For better accuracy, other datasets are obtained from controlled environments with multiple calibrated cameras and motion capture equipment \cite{ben2021ikea, trumble2017total, sigal2010humaneva, bhatnagar2022behave, cai2022humman}. Some methods fit SMPL from images to obtain pseudo ground truth \cite{mahmood2019amass, joo2017panoptic, huang2022capturing}. Most of the real datasets focus on estimating the 3D pose of the person or 3D joints, but there are only few that address body shape like SSP-3D \cite{STRAPS2018BMVC} and Human Bodies in the Wild  (HBW) \cite{choutas2022accurate}, however, none of these datasets is large enough to train a body shape regressor. The major drawback of real data is the lack of diversity in shape, pose, gender and race, which can introduce a bias to the learning models. Synthetic data doesn't suffer such bias, as it can be balanced out when being generated.

\paragraph{Synthetic data from real images.}
The simplest form to create synthetic data is to capture real people with a constant background and then render them synthetically in new scenes \cite{gabeur2019moulding, mehta2018single}. Another way to augment the data is through changing the texture of clothes like in MPI-INF-3DHP \cite{mehta2017monocular}. However, these have to be simple to keep the appearance of the clothes. Other approaches blend multiple images to create a new one that is visually appealing \cite{zanfir2020human} or to augment the poses \cite{rogez2016mocap}. However, these methods have not proven sufficient for HPS learning.

\paragraph{Synthetic data from classical rendering.} Using a classical rendering pipeline allows for the generation of a large number of image and ground truth pairs, with high diversity in poses, shapes, and backgrounds. Datasets like SURREAL \cite{varol17_surreal}, AGORA \cite{patel2021agora}, and SPECS \cite{kocabas2021spec} are useful for pre-training but are not sufficient to achieve SOTA performance without additional real training data. This limitation is likely due to a lack of realism in these datasets.

Two datasets have demonstrated state-of-the-art performance with synthetic data: the body model SOMA \cite{hewitt2023procedural} and BEDLAM. \citet{hewitt2023procedural} achieved state-of-the-art egocentric pose estimation with SimpleEgo \cite{cuevas2024simpleego} by training solely on SOMA. This dataset includes synthetic people with a wide variation in poses, shapes, skin, cloth textures, and hair, unfortunately the pipeline is not publicly available. BEDLAM is a dataset available for research and offers a large diversity of body and hand poses with physically simulated clothes, hair, and variations in gender and race. However, the realism, especially of skin texture and faces, remains a bottleneck, and simulating new garments for new body shapes and poses requires significant time.

\paragraph{Synthetic data from generative models.}
Recently, diffusion models \cite{rombach2022high} have attracted the attention for their easy trainable capabilities and output realism \cite{
ho2020denoising, song2020denoising, 
zhao2023uni, rombach2022high, zhang2023adding}. However, the question remains, \textit{``how can we use generative models to generate data for HPS regressors?''} DatasetDM \cite{wu2023datasetdm} was one of the first works to demonstrate that diffusion models can be used to generate synthetic data by training a decoder that is capable of generating the ground truth data from the input image for various tasks like image segmentation, depth estimation and human pose estimation. However, they still needed to train on real images to beat the baseline and train a decoder on the desired label.

Diffusion-HPC~\cite{weng2024diffusion} was among the first to use diffusion models to create synthetic data for HPS estimation. They use Stable Diffusion~\cite{rombach2022high} to generate images of people and an off-the-shelf HMR regressor to obtain pseudo-ground truth pose. Then, they improve the image using ControlNet \cite{zhang2023adding} and the depth of the mesh. This approach shows that photorealistic generated images can be used for HPS regressor. However, the mesh ground truth is still noisy and they only test their method on sports datasets. \citet{ge2024d} take a similar approach, incorporating a 2D keypoint detector and a refinement step to obtain mesh ground truth. While effective for pre-training, it did not surpass the performance of networks trained on BEDLAM.

To date, generative models have proven useful for pre-training but cannot be used alone without real data \cite{he2022synthetic}. The lack of a ground truth pair between the mesh and the generated image necessitates a refinement step, which can introduce noise and adversely affect training.
\section{Method}
\label{sec:method}
\subsection{Preliminaries}
\paragraph{Image diffusion models.}
Image-based latent diffusion models~\cite{rombach2022high} are text-to-image generative models capable of generating photorealistic images given text input. They learn to gradually denoise a latent representation of the image obtained from a variational autoencoder (VAE), starting from random noise until it resembles the target image described by the text prompt. This process is guided by the conditional information provided by the text input, ensuring that the generated image aligns with the given description.

\paragraph{Image inpainting.} It involves filling parts of an image in a coherent and visually consistent manner. In the context of stable diffusion, the inpainting network is trained by adding an extra input to the U-Net denoiser network~\cite{rombach2022high}. Although fewer models are specifically trained for inpainting, a practical workaround involves sampling the known region from the input image at each denoising step $t$ and combining it with the inpainted part~\cite{lugmayrinpainting}.

\paragraph{ControlNet.}
ControlNet introduces conditional controls to the Stable Diffusion network~\cite{rombach2022high} by leveraging additional inputs such as edges, depth, segmentation, and 2D keypoints. This is achieved by training a small auxiliary network that takes a conditioning image \textit{c} as input, and the outputs are added to the stable diffusion U-Net. The resulting image tries to follow the condition signals, see \cref{fig:control_differences}.

\paragraph{BEDLAM dataset.}
The dataset was rendered using Unreal Engine 5 (EU5)~\cite{unreal} consists of 10450 images sequences of size $1290\times720$ pixels, with physically simulated clothes. The GT data consists of depth maps, segmentation masks for body, cloth, and environment, as well as SMPL-X pose and shape parameters. 

\subsection{Generative BEDLAM}
\begin{figure*}[!htbp]
\begin{center}
   \includegraphics[width=0.72\linewidth]{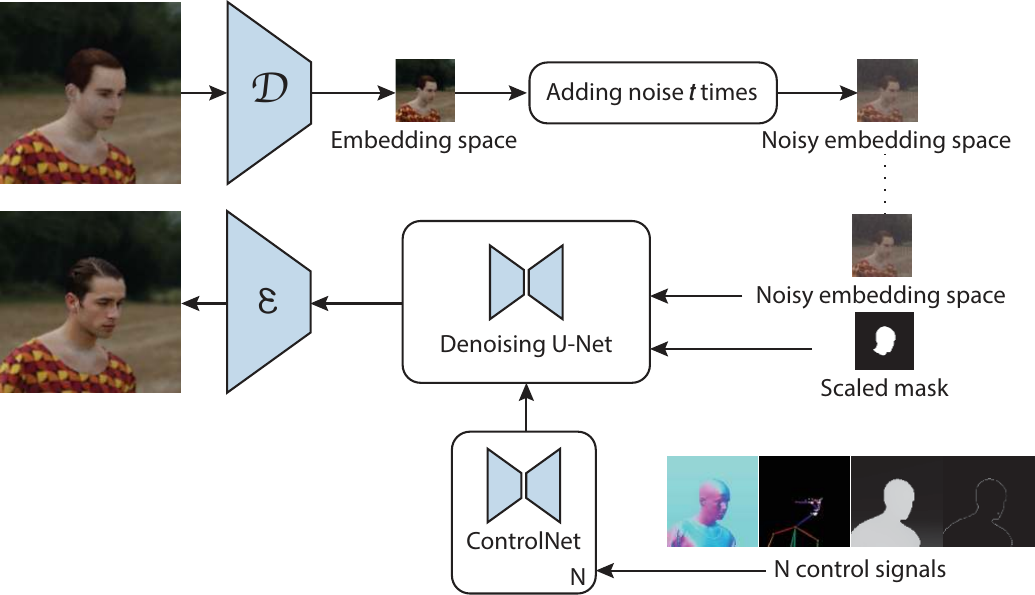}
\end{center}
   \caption{\textbf{Gen-B pipeline applied to ``head''.} We use the GT color image from BEDLAM and add noise for $t$ steps. The mask of the head and the embeddings are used to inpaint the region of the head. To preserve the shape and pose we use the surface normals, pose, depth, and edges as control signals.}
\label{fig:gen_bedlam_model}
\end{figure*}

\begin{figure*}
\begin{center}
   \includegraphics[width=.9\linewidth]{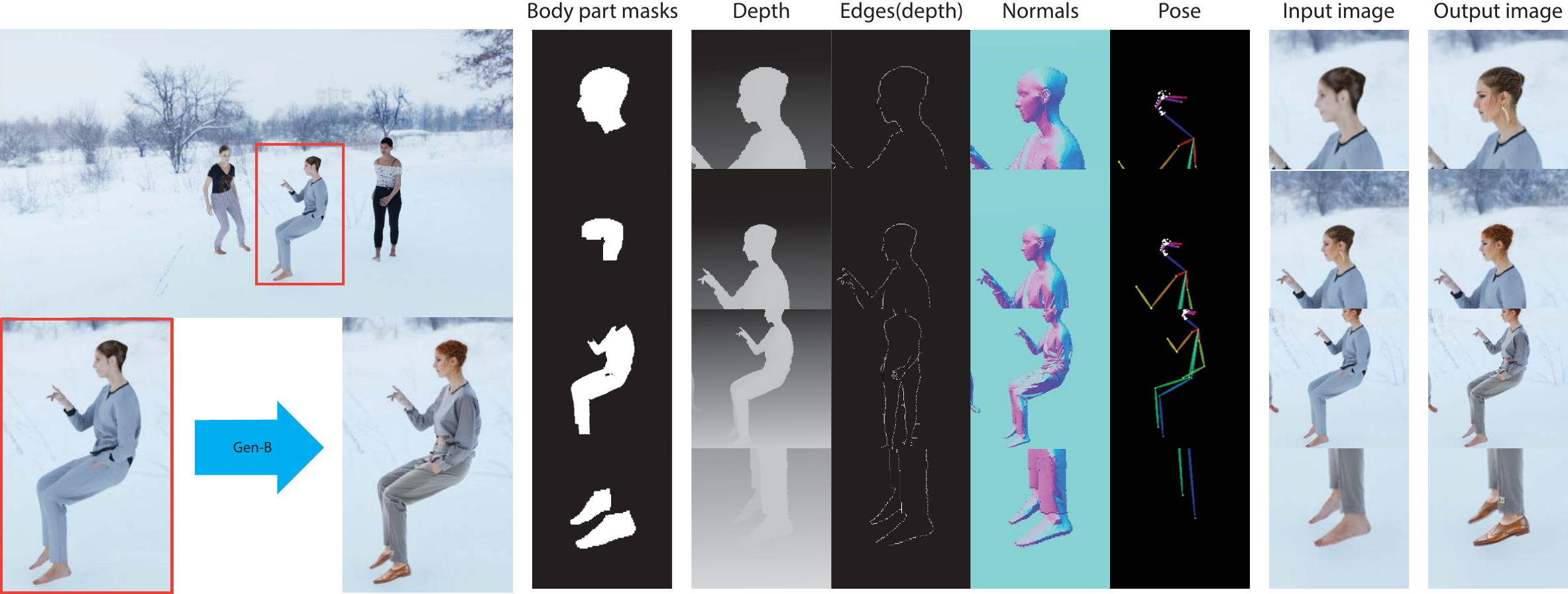}
\end{center}
   \caption{\textbf{Gen-B process.} Our method takes as input a BEDLAM image and processes each synthetic body in the image. For each person, it crops around it using the body part masks and uses the cropped region as input for our pipeline. Our pipeline uses depth, edges from depth, surface normals and 2D poses as control signal for the multi-ControlNet network. Our pipeline prcesses sequentially the head, then the hair, body and finally the feet of the person. Once the generation is done, it continues to the next person.}
\label{fig:bedlam_in_images}
\end{figure*}

Our method lies between generative dataset generation and the rendering process, leveraging the advantages of both. It consists of two crucial components: ground truth data from BEDLAM, which provides synthetic images that we use as noisy images (regions we aim to ``improve'' or modify), and a diffusion model that processes the ground truth image (reference image) to transform it into a photorealistic one while preserving shape and appearance. We control these transformations using the metadata provided by BEDLAM and ControlNet networks. Our pipeline and outputs are shown in \cref{fig:gen_bedlam_model} and \cref{fig:bedlam_in_images}.

\begin{figure}
    \centering
    \includegraphics[width=0.8\linewidth]{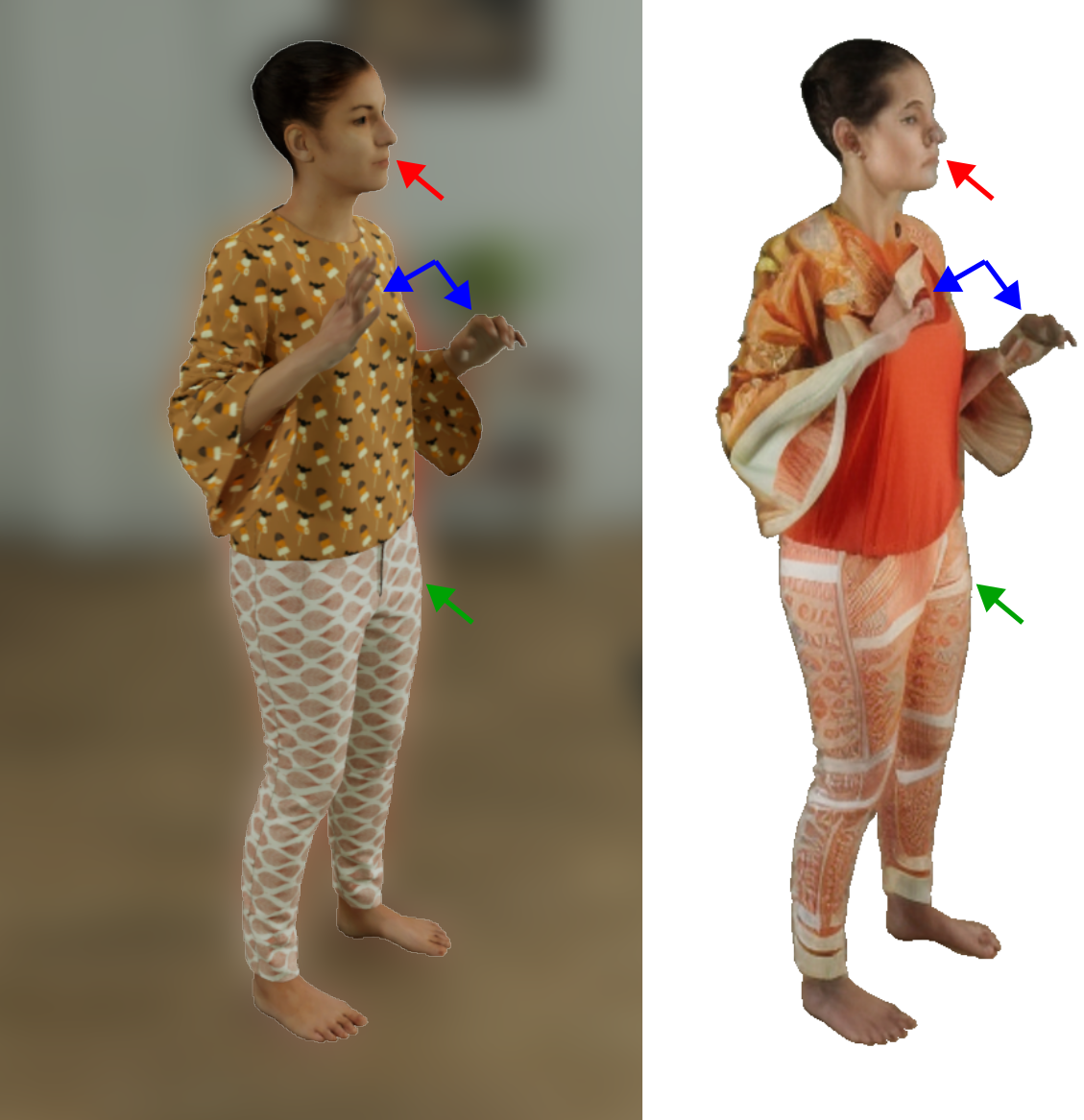}
    \caption{\textbf{Inpainting errors.} When inpainting is performed on the masked body, the body shape, hand, and face changes.}
    \label{fig:mask_body_inpainting}
\end{figure}

\paragraph{Synthetic-to-real.} BEDLAM employs HDRI and 3D backgrounds and renders people within these scenes. There are various methods to enhance the photorealism of these images. A naive approach involves using an image-to-image process, where the reference image is input into the Stable-Diffusion (SD) network and then denoised with prompts such as, ``photorealistic'', ``8K'', or ``Real people''. However, this approach generally modifies the images completely and alters the shape of the people. A more effective approach is to encode the reference image, introduce a small amount of noise, and then perform reverse diffusion starting from a later timestep $t$ rather than the initial timestep $T$, similar to SDEdit~\cite{meng2021sdedit}. Another viable method involves masking out the people and applying image inpainting. This approach keeps the realistic background unchanged, but it still modifies the face and shape of the synthetic people, \cref{fig:mask_body_inpainting}. However, even when we use ControlNet on top apply additional guidance, we find that the more regions the network has to attend for the inpainting, the more difficult it is to preserve the details of the original image.

SD generates photorealistic images of scenes and places, or people, but if the image is too complex, it tends to lose details in favor of producing an overall visually appealing image~\cite{ruiz2023dreambooth}. For example, if it has to generate a crowded city, the people will have fewer details, or if it generates a meadow with trees and river, the trees will lose high frequency details~\cite{ruiz2023dreambooth}. However, if it has to generate only a single building, or a face, it does a better job at recovering these small details.

Based on this observation, we decided to improve the realism of the BEDLAM people in the image by enhancing individual parts separately to generate detailed images. We modify the head, hair, clothes and feet. The head is a critical area where naive approaches fail to maintain consistency with the ground truth data. Most of the BEDLAM dataset lacks hair, so we focused on adding realistic hair. For clothes, we aimed to increase realism while preserving the shape consistent with the ground truth, in \cref{sec:ablation} we show there exist a trade-off between diversity and shape and pose consistency and it is hard for current diffusion models to achieve both. Additionally, since the BEDLAM dataset does not include shoes, we incorporated realistic shoes into the images.

For each part, we use an image inpainting approach, where we encode the image into a latent space, then, similar to SDEdit~\cite{meng2021sdedit}, we initialize the reverse diffusion process from an intermediate time $t_0 \in (0,1)$. However, unlike SDEdit, we don't do it over the whole image; we only do it on the masked region. Our inpainting process is similar to RePaint~\cite{lugmayrinpainting}, where for each denoising step, the known region is sampled from the input image, and it is combined with the inpainted part. Since we utilize Stable Diffusion, the merging of the two outputs is performed in the latent space, unlike RePaint, which operates at the pixel level. 

\begin{figure*}[!t]
    \centering
    \includegraphics[width=0.95\linewidth]{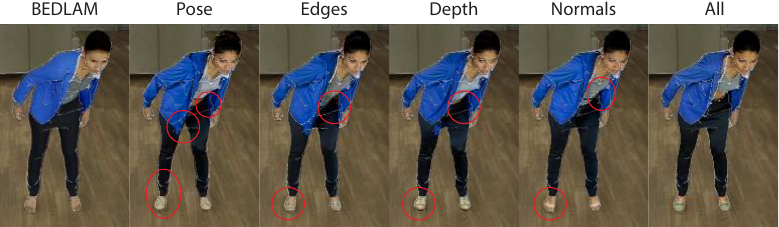}
    \caption{\textbf{ControlNet-generated images}. (better seen when zoomed-in). At first glance, it looks like the generated images with different conditioning images successfully convert BEDLAM into a well-aligned photorealistic image. However, if we look closely, we can observe that they modify parts of the body (red circles), which creates a mismatch with the GT mesh data. When we combine all the control signals, we manage to enforce the shape and pose consistency. We overlapped the images with the edges of the body to highlight the changes.}
\label{fig:control_differences}
\end{figure*}

\paragraph{Additional control.} When adding noise to the latent space of the image, we lose important information that the synthetic image can provide, such as head and body pose, as well as body shape appearance (see image \cref{fig:control_differences}). To preserve this information, we apply multi-ControlNet, and include depth, surface normals, edges, and 2D joints as control signals. We also provide text guidance for each part. We take advantage that BEDLAM~\cite{black2023bedlam} provides the gender and race metadata and use it as part of the text prompt. Including race and gender helps mainly to keep the skin texture consistent with the GT image. The prompts we use are the following. Hair prompt: ``Realistic \{hair\_color\} \{hair\_type\} \{race\} \{gender\}''. Where \{hair\_color\} can be: blond, brunette, redhead, and \{hair\_type\} is any of: straight, wavy, curly, coily, mullet, afro.
Head prompt: ``Realistic '\{race\} \{gender\} face''.
Body prompt: ``Realistic \{gender\} clothes''. 
Feet prompt: ``Realistic \{gender\} \{shoes\_type\}''. The \{shoes\_type\} are: oxford shoes, boots, sneakers, sandals, crocs.
We decided on a more generic prompt as we rely on the random noise initialization to provide variety, especially on the clothes.

\paragraph{Conditioning images.} We use depth, surface normals, 2D pose, and edges as control signals. Previous methods~\cite{weng2024diffusion,ge2024d} obtain the conditioning images by passing the photorealistic generated images through networks that predict depth, surface normals, and 2D poses, resulting in pseudo-GT. However, this approach inherently introduces noise into the control signal. To avoid noisy inputs, we directly process the metadata provided by BEDLAM. Specifically, we utilize the depth maps and SMPL-X body parameters. For ControlNet conditioning inputs, we derive the camera surface from the depth map and obtain the 2D pose by projecting the SMPL-X body joints into the 2D image. We also generate edges by applying a Canny edge detector to the depth map, ensuring that high-frequency details, such as small folds in clothing, are controlled by surface normals and depth rather than edges. Finally, we normalize the depth map to a range of 0 to 1. An example of the control images is shown in \cref{fig:bedlam_in_images}.


\paragraph{Body part masks.}
BEDLAM also provides segmentation masks for body, simulated cloth and environment. However, it doesn't provide masks for body parts or the regions where the cloth is only a skin texture. To get this data, we used the SMPL-X texture map and label the scalp, face, head, body, and feet. We also save the cloth skin texture regions as cloth mask when there is no simulated cloth segmentation.

\section{Experiments}
\label{sec:experiment}
As described in the previous section, we transform BEDLAM images to make them look more photorealistic while preserving the pose and shape of the person in the image using SD. We use an SD fine-tuned model on realistic images \cite{realisticvision}, as we empirically found that the vanilla version of SD almost always generates deformed faces.
A key factor in this process is determining the appropriate noise level to add to the original image. Through our observations, we found that a low noise level is essential for the face, hair, and clothing, as higher noise values can lead to changes in pose and shape or even completely alter the image, as illustrated in \cref{fig:noise_examples}. Thus, we selected a noise value of 0.35 for these regions. Conversely, we observed that low noise levels were insufficient to convert bare feet into shoes, necessitating the use of a higher noise value of 0.5 for the feet. A more detailed explanation of these decisions is provided in \cref{sec:ablation}. We show qualitative results of the dataset in the Sup. Mat.
 
 As pointed by SMPLer-X~\cite{smplerx}, BEDLAM is the only synthetic dataset that produces SOTA performance on multiple benchmark without additional training data. Hence we use BEDLAM  for our experiments and modify it to create Gen-B.
To ensure a fair comparison and avoid results being skewed by additional data, we conduct our experiments without using any extra training datasets. We compare Gen-B and BEDLAM~\cite{black2023bedlam} by training HMR~\cite{kanazawa2018end}, CLIFF~\cite{li2022cliff}, and HMR2.0~\cite{goel2023humans} using both datasets. We evaluate performance using Mean Per Joint Position Error (MPJPE) and Per Vertex Error (PVE) metrics on the 3DPW~\cite{von2018recovering}, EMDB~\cite{kaufmann2023emdb}, and RICH~\cite{huang2022capturing} benchmarks. HMR and CLIFF are trained for 550K iterations with a batch size of 64, as per~\cite{black2023bedlam}, while HMR2.0 is trained for 300K iterations with a batch size of 48 according to its configuration \cite{goel2023humans} and the dataloader of \cite{dwivedi2024tokenhmr}. The results are shown in \cref{tab:experiments}.

\definecolor{BrightGray}{rgb}{0.247,0.262,0.313}
\begin{table*}
\centering
\small
\begin{tblr}{
  width = 0.9\linewidth,
  colspec = {Q[136]Q[173]Q[100]Q[92]Q[109]Q[109]Q[100]Q[109]},
  row{1} = {c},
  row{2} = {c},
  cell{1}{1} = {r=2}{},
  cell{1}{2} = {r=2}{},
  cell{1}{3} = {c=2}{0.192\linewidth},
  cell{1}{5} = {c=2}{0.218\linewidth},
  cell{1}{7} = {c=2}{0.209\linewidth},
  cell{3}{1} = {r=2}{},
  cell{3}{3} = {c,fg=BrightGray},
  cell{3}{4} = {c,fg=BrightGray},
  cell{3}{5} = {c,fg=BrightGray},
  cell{3}{6} = {c,fg=BrightGray},
  cell{3}{7} = {c,fg=BrightGray},
  cell{3}{8} = {c,fg=BrightGray},
  cell{4}{3} = {c,fg=BrightGray},
  cell{4}{4} = {c,fg=BrightGray},
  cell{4}{5} = {c,fg=BrightGray},
  cell{4}{6} = {c,fg=BrightGray},
  cell{4}{7} = {c,fg=BrightGray},
  cell{4}{8} = {c,fg=BrightGray},
  cell{5}{1} = {r=2}{},
  cell{5}{3} = {c},
  cell{5}{4} = {c},
  cell{5}{5} = {c},
  cell{5}{6} = {c},
  cell{5}{7} = {c},
  cell{5}{8} = {c},
  cell{6}{3} = {c},
  cell{6}{4} = {c},
  cell{6}{5} = {c},
  cell{6}{6} = {c},
  cell{6}{7} = {c},
  cell{6}{8} = {c},
  cell{7}{1} = {r=2}{},
  cell{7}{3} = {c,fg=BrightGray},
  cell{7}{4} = {c,fg=BrightGray},
  cell{7}{5} = {c,fg=BrightGray},
  cell{7}{6} = {c,fg=BrightGray},
  cell{7}{7} = {c},
  cell{7}{8} = {c},
  cell{8}{3} = {c,fg=BrightGray},
  cell{8}{4} = {c,fg=BrightGray},
  cell{8}{5} = {c,fg=BrightGray},
  cell{8}{6} = {c},
  cell{8}{7} = {c},
  cell{8}{8} = {c},
  cell{9}{1} = {r=2}{},
  cell{9}{3} = {c},
  cell{9}{4} = {c},
  cell{9}{5} = {c},
  cell{9}{6} = {c},
  cell{9}{7} = {c},
  cell{9}{8} = {c},
  cell{10}{3} = {c},
  cell{10}{4} = {c},
  cell{10}{5} = {c},
  cell{10}{6} = {c},
  cell{10}{7} = {c},
  cell{10}{8} = {c},
  hline{1,3,5,7,9,11} = {-}{},
  hline{2} = {3-8}{},
}
\textbf{Network} & \textbf{ Dataset} & \textbf{3DPW}  &                & \textbf{EMDB}   &                                                     & \textbf{RICH}  &                 \\
                 &                   & \textbf{MPJPE} $\downarrow$ & \textbf{PVE} $\downarrow$   & \textbf{MPJPE} $\downarrow$  & \textbf{PVE} $\downarrow$                                        & \textbf{MPJPE} $\downarrow$ & \textbf{PVE} $\downarrow$    \\
HMR              & BEDLAM            &    93.62            &    107.95            &      142.19           &     166.14                                                &       110.15         &      124.85           \\
                 & Gen-B        &     \textbf{91.40}           &      \textbf{106.42}          &     \textbf{135.56}            &    \textbf{159.20}                                                 &     \textbf{108.00}           &     \textbf{121.58}            \\
CLIFF            & ~BEDLAM           & 80.30          & 94.14          & 115.60          & \textbf{133.60}                                     & 95.66          & 106.71          \\
                 & Gen-B        & \textbf{79.80} & \textbf{94.02} & \textbf{115.37} & 133.77                                              & \textbf{92.80} & \textbf{103.75} \\
HMR2.0           & BEDLAM            & 72.35 & 85.99 & \textbf{88.66}           & \textbf{105.64}                                              &     84.07           &        97.63         \\
                 & Gen-B        & \textbf{71.38}          & \textbf{85.87}          & 90.69  & 108.58 &       \textbf{82.17}         &     \textbf{96.09}            \\
\end{tblr}
\caption{\textbf{Comparison of model performance on the 3DPW, EMDB, and RICH.} Networks trained with Gen-B dataset get an incremental and consistent gain over those trained with BEDLAM.}
\label{tab:experiments}
\end{table*}

\begin{figure}[htbp]
    \centering
    \resizebox{0.7\columnwidth}{!}{ 
    \begin{tikzpicture}
        \begin{axis}[
            xlabel={Networks},
            ylabel={$PVE_{diff}$},
            xtick={1,2,3},
            xticklabels={HMR, CLIFF, HMR2.0},
            legend pos=north east,
            grid=both,
            ymin=-4, ymax=8
        ]
    
        \addplot[
            color=red,
            mark=*,
            ] coordinates {
            (1,1.53)
            (2,0.12)
            (3,0.12)
        };
        \addlegendentry{3DPW}
    
        \addplot[
            color=blue,
            mark=square*,
            ] coordinates {
            (1,6.94)
            (2,-0.17)
            (3,-2.93)
        };
        \addlegendentry{EMDB}
    
        \addplot[
            color=green,
            mark=triangle*,
            ] coordinates {
            (1,3.27)
            (2,2.96)
            (3,1.54)
        };
        \addlegendentry{RICH}
        \end{axis}
    \end{tikzpicture}
    }
    \caption{\textbf{PVE difference comparison.} We observe that better architectures achieve a lower PVE, suggesting they are more effective at disregarding high-frequency information in photorealistic images to achieve accurate mesh predictions.}
    \label{fig:pve_differences}
\end{figure}
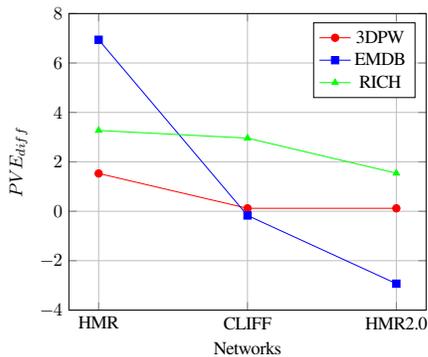

Training HMR on Gen-B improves accuracy on real image datasets, 3DPW, EMDB, and RICH, by $2.37\%$, $4.66\%$, and $1.95\%$, respectively. CLIFF~\cite{li2022cliff} presents the same behavior, where Gen-B gets a reduction in error of $0.6\%$, $0.1\%$, and $2.99\%$ percent for 3DPW, EMDB, and RICH. For HMR2.0, Gen-B performs better by $1.34\%$ and $2.26\%$. However, we perform $2.23\%$ worse in EMDB. We analyze why networks trained with Gen-B show a small but consistent improvement over those trained with BEDLAM on the 3DPW and RICH datasets, but perform similarly on the EMDB dataset. Thus, we compute the Fréchet Inception Distance (FID)~\cite{heusel2017gans} score between the real image datasets and BEDLAM and Gen-B. 
The FID score measures the distance between the distributions of real and synthetic data, with lower scores indicating that the synthetic data is more similar to the real data. The FID score is calculated for each person (or synthetic person) in the scene by cropping around its center. From \cref{tab:fid_bedlam_gen_bedlam}, 
we observe that Gen-B more closely matches the distributions of 3DPW and RICH compared to BEDLAM, while Gen-B and BEDLAM have similar score for the EMDB dataset.
This could explain the mixed results on the EMDB dataset. We also find that our photorealistic images can help old architectures like the conv-net HMR to improve their performance over BEDLAM. However, this advantage diminishes when more information is provided, like in CLIFF, where the camera intrinsics are given, or with advanced architecture, like the vision transformer from HMR2.0, see \cref{fig:pve_differences}. For instance, on the RICH dataset, we observe that the PVE difference between Gen-B and BEDLAM diminishes based on the network architecture: 3.27 with HMR, 2.96 with CLIFF, and 1.54 with HMR2.0. We also observe that HMR2.0 trained only on our synthetic data performs better than HMR2.0b by $12.2\%$, and we are only $1.9\%$ worse than HMR2.0a.

In the next section, we explore how the added noise and different control signals affect the performance of HPS regressors.

    
    
    

\begin{table}
\centering
\small
\begin{tblr}{
  width = \linewidth,
  colspec = {Q[236]Q[217]Q[219]Q[217]},
  hline{1-2,4} = {-}{},
}
       & 3DPW $\downarrow$            & EMDB $\downarrow$            & RICH $\downarrow$           \\
BEDLAM & 137.80          & 130.68          & 133.28          \\
Gen-B  & \textbf{135.79} & \textbf{130.21} & \textbf{131.98} 
\end{tblr}
\caption{\textbf{BEDLAM and Gen-B FID score comparison}. Gen-B distribution is closer to the real datasets than BEDLAM.}
\label{tab:fid_bedlam_gen_bedlam}
\end{table}

\subsection{Ablation Studies}
\label{sec:ablation}

\begin{figure}
\begin{center}
   \includegraphics[width=\linewidth]{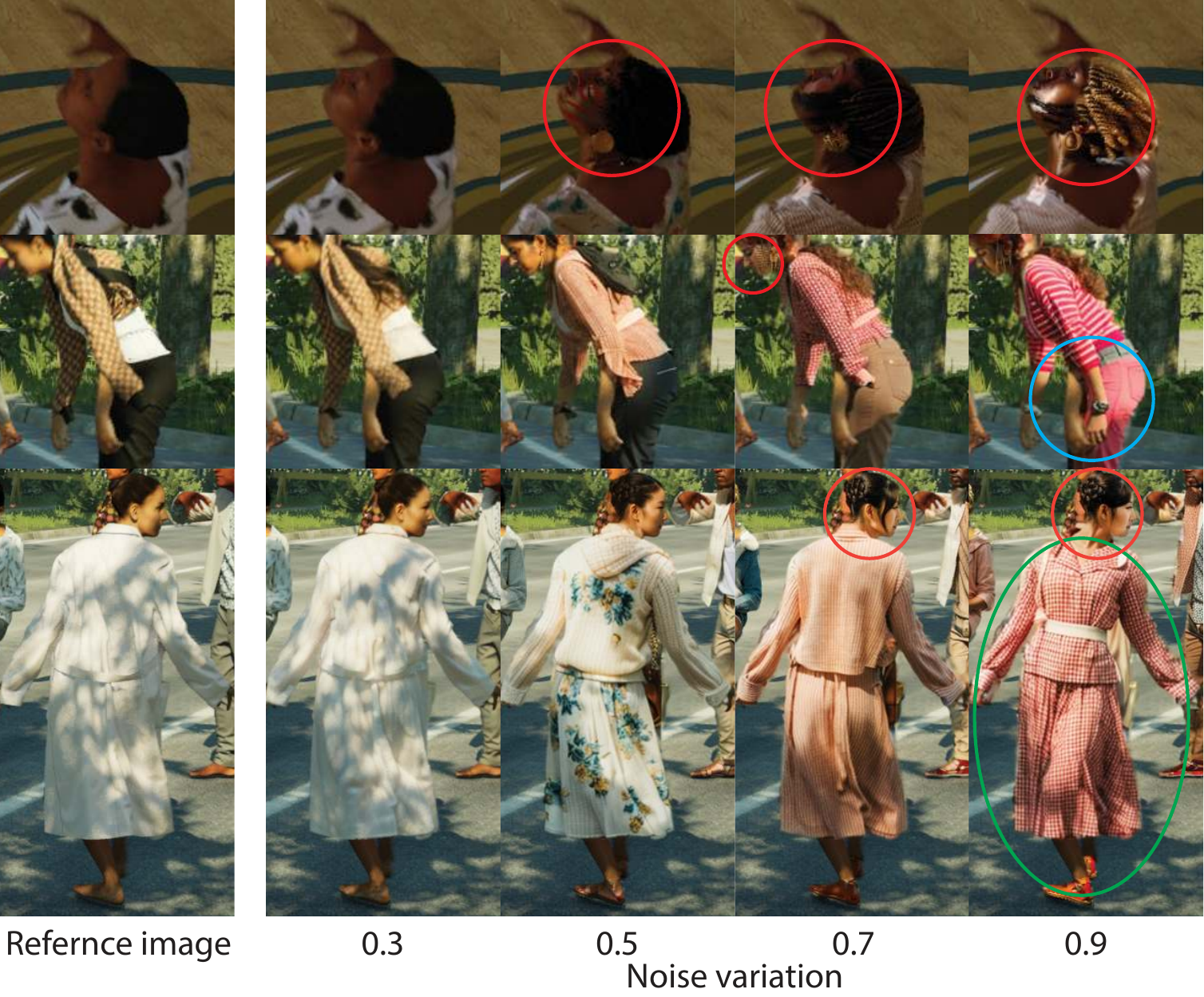}
\end{center}
   \caption{\textbf{Generated images based on different noise values.} We demonstrate how high noise can lead to artifacts on the image. Some of these modifications are: \textcolor{red}{Adding artifacts}, \textcolor{cyan}{Generating a part that does not exist}, \textcolor{green}{Flipping parts of the body}, \textcolor{orange}{modifying pose}, in the last row we can see that the head pose was changed. See Sup. Mat. for more results.}
\label{fig:noise_examples}
\end{figure}

\paragraph{Multi-control}
We observe that generating an image without additional control, even with a single control signal, cannot preserve the shape and pose of the person, as seen in \cref{fig:control_differences} and \cref{fig:noise_examples}. Therefore, more control signals should be used to enforce the shape and pose preservation. In this section, we explore how different control signals, edges, depth, normals, and pose affect the performance of HPS regressors. Thus, we train BEDLAM-CLIFF on a subset of BEDLAM, which consists of 126,597 images and fix the noise level to 0.5 for all experiments. First, we tested how each control signal affects the network's performance, so we generated four datasets, one for each control signal. 
Then, we picked the best result and created another dataset by including the control signal with the second-best results. We did the same for the third and fourth-best results. Comparing the individual performance of each control signal, from \cref{tab:exp_controlnet} we can see that depth information gives the lowest MPJPE: 104.85 showing to be a good control signal to preserve the GT alignment. However, each control signal by itself performs worse than when combined. We can see the reason when we visually inspect the generated images. In \cref{fig:control_differences}, at first glance, the images generated by the control signals look like they preserve the pose and shape from the reference image. However, when we do a close-up, we can observe that some body parts are changed, which introduces noise to the training data. On the other hand, when we combine all the control signals, the generated images follow better the control signals and reference image. However, when the noise is high, or the pose is complex, it is still not guaranteed that the generated image will preserve the alignment; see Sup. Mat.
\begin{table}
\centering
\small
\begin{tblr}{
  width = \linewidth,
  colspec = {Q[607]Q[154]Q[154]},
  cells = {c},
  hline{1-2,9} = {-}{},
}
\textbf{Control Signal}  & \textbf{MPJPE} $\downarrow$ & \textbf{PVE}  $\downarrow$ \\
Edges                    & 110.31        & 129.48       \\
Depth                    & 104.85        & 122.96       \\
Normals                  & 115.27        & 135.04       \\
Pose                     & 108.14        & 127.02       \\
Depth+Pose               & 105.64         & 123.64        \\
Depth+Pose+Edges         & 104.82         & 121.63        \\
Depth+Pose+Edges+Normals & \textbf{98.36}        & \textbf{115.36}       
\end{tblr}
\caption{\textbf{Performance of ControlNet inputs on 3DPW.} We observe that the best result is obtained when combining all the control signals.}
\label{tab:exp_controlnet}
\end{table}

\paragraph{Noise level}
Under the optimal control signal combination to control and preserve the image appearance, we evaluate how the the noise level can affect the HPS network performance. Visually, see \cref{fig:teaser}, we observe that there is a trade-off between faithfulness to the input image (rendered BEDLAM) and the realistic but more diverse image when we add different starting noise levels on the reverse denoising process. We explore whether this realism translates to better performance. We use the same subset and network as the previous experiment and vary the noise level: 0.3, 0.5, 0.7, and 0.9. This is fixed for all the parts.

\begin{table}
\small
\centering
\begin{tblr}{
  cells = {c},
  hline{1-2,6,8} = {-}{},
}
\textbf{Noise level} & \textbf{MPJPE} $\downarrow$ & \textbf{PVE} $\downarrow$ \\
0.3                  & \textbf{97.56}         & \textbf{113.63}        \\
0.5                  & 98.36         & 115.36        \\
0.7                  & 101.29         & 118.37        \\
0.9                  & 104.38         & 121.50       \\
BEDLAM               & 96.64         & 112.81      \\
Gen\mbox{-}B         & \textbf{95.00}         & \textbf{110.82} 
\end{tblr}
\caption{\textbf{Noise variation performance on 3DPW.} Even though more noise means more diversity, it harms the performance of the network as the alignment between the GT image and body mesh is affected. We also found that different parts need different noise level to get better performance than BEDLAM with the generated images (Gen\mbox{-}B).}
\label{tab:exp_noise}
\end{table}

\begin{figure}[ht]
    \centering
    \begin{tikzpicture}
        \begin{axis}[
            xlabel={MPJPE (mm)},
            ylabel={FID Score},
            legend style={
                at={(1,0)},
                anchor=south east,
                legend columns=1,
                column sep=1ex
            },
            ymajorgrids=true,
            grid style=dashed,
             ytick={140,145,150,155,160,165}, 
            xmin=94, xmax=106,
            ymin=140, ymax=165,
            width=0.9\columnwidth,
            height=0.6\columnwidth
        ]
        \addplot[
            color=blue,
            mark=square*,
            ]
            coordinates {
            (97.56,146.25)(98.37,144.56)(101.29,152.04)(104.38,161.58)
            };
            \addlegendentry{Noise level}
        
        \addplot[
            color=red,
            mark=diamond*,
            ]
            coordinates {
            (96.6,145.59)
            };
            \addlegendentry{BEDLAM}
        
        \addplot[
            color=black,
            mark=mystar,
            ]
            coordinates {
            (95.0,145.04)
            };
            \addlegendentry{Gen-B}

        \addplot[
            only marks,
            mark=none,
            nodes near coords,
            point meta=explicit symbolic,
            visualization depends on=\thisrow{label} \as \label]
            table[meta=label] {
                x y label
                97.56 146.25 {0.3}
                98.37 144.56 {0.5}
                101.29 152.04 {0.7}
                104.38 161.58 {0.9}
            };
        \end{axis}
    \end{tikzpicture}
    \caption{\textbf{FID Score vs. MPJPE on 3DPW.} We observe a correlation between the FID score and HMR regressor performance. Higher noise generates more photo-realistic images visually (\cref{fig:noise_examples}), but it drives away from the real dataset distribution and harms the ground truth data, resulting in worse performance.}
    \label{fig:fid_mpjpe}
\end{figure}
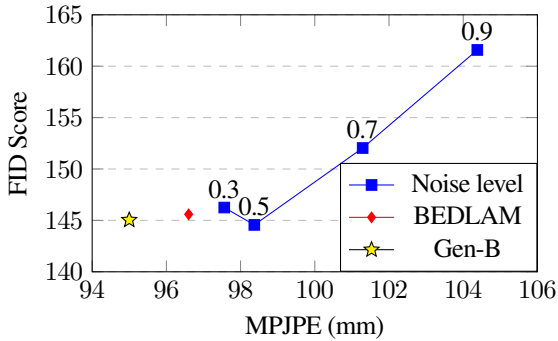

In \cref{fig:fid_mpjpe}, we observe the performance diminishes when more noise is added. At first glance, this is counterintuitive, as images generated with greater amounts of noise levels appear more realistic and diverse than those with lower noise levels. However, the more freedom we allow the network in modifying the images, the less faithful it becomes to the reference image. In the context of HPS, this leads to changes in pose and shape, as evidenced by \cref{tab:exp_noise} and \cref{fig:noise_examples}. This occurs even when control signals such as pose, surface normals, and depth are present. At high noise levels, the model may also hallucinate body parts, such as placing a face on the back of the head or generating an additional hand, as shown in \cref{fig:noise_examples}. These unwanted modifications hinder the performance of HPS regressors, as it's difficult to filter them out when generating a large-scale dataset. This also explains why HPS diffusion models have failed to serve as standalone datasets for training HPS models without fine-tuning on real data, as noted in previous works~\cite{weng2024diffusion, ge2024d}.

However, even at the lowest noise level, the generated images do not improve the performance of the HPS network compared to the original BEDLAM dataset. This raises the question of whether the photorealistic images are indeed closer to the real image distribution compared to BEDLAM. To address this, we calculate the FID score between crops around the people of the synthetic data and the 3DPW real dataset. The results indicate that, while higher noise levels indeed causes a greater deviation from the real dataset distribution, see \cref{fig:fid_mpjpe}. As result, the MPJPE error increases as more noise is added. 

Based on the analysis above, and from \cref{tab:exp_noise} and \cref{fig:fid_mpjpe},  we observe that the best results, which are also closest to the real data distribution, occur at noise levels between 0.3 and 0.5. We also observe, after visual inspection, that certain parts do not change at a low noise level, for example shoes are almost never added to the feet at noise level 0.3, see \cref{fig:teaser} and \cref{fig:noise_examples}. Additionally, incorrect poses and extra body parts begin to appear at a noise level of 0.5 when the pose is heavily occluded, as in the first row of \cref{fig:noise_examples}. Based on these insights, we created another subset where we increase the noise of the face, clothes and hair to 0.35, while the noise for the feet was set to 0.5 to encourage the generation of shoes. This resulted in improved performance compared to the original BEDLAM dataset, with Gen-B achieving an MPJPE of 95.00, compared to 96.64 for BEDLAM. These results underscore that generating well-aligned data using generative models is non-trivial and that careful attention must be paid to each body part to avoid degrading the quality of the data if it is intended for use in HPS.

\section{Conclusion}
\label{sec:conclusions}
Based on our analysis and experiments, we can answer how we can use generative models to create photorealistic images without harming the performance of HPS regressors. Our method shows that enhancing realism improves HPS performance and generalizes well to real-world datasets like 3DPW, EMDB, and RICH. However, if we are not careful, the generated images can be visually appealing but negatively impact HPS performance, as they may alter the pose and shape of the person, from slight pose variations to the creation of extra body parts. After an extensive evaluation, we obtained a method that provides a balance between diversity and faithfulness. We also showed that applying inpainting to each part of the image is not enough, and we need to add extra constraints to enforce pose and shape consistency by using multiple control signals. Otherwise, the generated images will deviate from the source images, which will result in noisy training data and a worse performance.
 
The caveat to the above. While generating realistic-looking humans with perfect ground truth data might seem straightforward, it requires significant hyper-parameter tuning to control noise levels and ensure the control signals don't deviate from the distribution of real images. Additionally, visual inspection is necessary to verify whether the generated images closely follow the reference images. This is difficult at scale and shows why previous attempts to create a photorealistic dataset that performs better than BEDLAM failed.

{
    \small
    \bibliographystyle{ieeenat_fullname}
    \bibliography{main}

\begin{thebibliography}{47}
\providecommand{\natexlab}[1]{#1}
\providecommand{\url}[1]{\texttt{#1}}
\expandafter\ifx\csname urlstyle\endcsname\relax
  \providecommand{\doi}[1]{doi: #1}\else
  \providecommand{\doi}{doi: \begingroup \urlstyle{rm}\Url}\fi

\bibitem[Andriluka et~al.(2014)Andriluka, Pishchulin, Gehler, and Schiele]{andriluka20142d}
Mykhaylo Andriluka, Leonid Pishchulin, Peter Gehler, and Bernt Schiele.
\newblock 2d human pose estimation: New benchmark and state of the art analysis.
\newblock In \emph{Proceedings of the IEEE Conference on computer Vision and Pattern Recognition}, pages 3686--3693, 2014.

\bibitem[Ben-Shabat et~al.(2021)Ben-Shabat, Yu, Saleh, Campbell, Rodriguez-Opazo, Li, and Gould]{ben2021ikea}
Yizhak Ben-Shabat, Xin Yu, Fatemeh Saleh, Dylan Campbell, Cristian Rodriguez-Opazo, Hongdong Li, and Stephen Gould.
\newblock The ikea asm dataset: Understanding people assembling furniture through actions, objects and pose.
\newblock In \emph{Proceedings of the IEEE/CVF Winter Conference on Applications of Computer Vision}, pages 847--859, 2021.

\bibitem[Bhatnagar et~al.(2022)Bhatnagar, Xie, Petrov, Sminchisescu, Theobalt, and Pons-Moll]{bhatnagar2022behave}
Bharat~Lal Bhatnagar, Xianghui Xie, Ilya~A Petrov, Cristian Sminchisescu, Christian Theobalt, and Gerard Pons-Moll.
\newblock Behave: Dataset and method for tracking human object interactions.
\newblock In \emph{Proceedings of the IEEE/CVF Conference on Computer Vision and Pattern Recognition}, pages 15935--15946, 2022.

\bibitem[Black et~al.(2023)Black, Patel, Tesch, and Yang]{black2023bedlam}
Michael~J Black, Priyanka Patel, Joachim Tesch, and Jinlong Yang.
\newblock Bedlam: A synthetic dataset of bodies exhibiting detailed lifelike animated motion.
\newblock In \emph{Proceedings of the IEEE/CVF Conference on Computer Vision and Pattern Recognition}, pages 8726--8737, 2023.

\bibitem[Cai et~al.(2022)Cai, Ren, Zeng, Lin, Yu, Wang, Fan, Gao, Yu, Pan, et~al.]{cai2022humman}
Zhongang Cai, Daxuan Ren, Ailing Zeng, Zhengyu Lin, Tao Yu, Wenjia Wang, Xiangyu Fan, Yang Gao, Yifan Yu, Liang Pan, et~al.
\newblock Humman: Multi-modal 4d human dataset for versatile sensing and modeling.
\newblock In \emph{European Conference on Computer Vision}, pages 557--577. Springer, 2022.

\bibitem[Cai et~al.(2024)Cai, Yin, Zeng, Wei, Sun, Yanjun, Pang, Mei, Zhang, Zhang, et~al.]{smplerx}
Zhongang Cai, Wanqi Yin, Ailing Zeng, Chen Wei, Qingping Sun, Wang Yanjun, Hui~En Pang, Haiyi Mei, Mingyuan Zhang, Lei Zhang, et~al.
\newblock Smpler-x: Scaling up expressive human pose and shape estimation.
\newblock \emph{Advances in Neural Information Processing Systems}, 36, 2024.

\bibitem[Choutas et~al.(2022)Choutas, M{\"u}ller, Huang, Tang, Tzionas, and Black]{choutas2022accurate}
Vasileios Choutas, Lea M{\"u}ller, Chun-Hao~P Huang, Siyu Tang, Dimitrios Tzionas, and Michael~J Black.
\newblock Accurate 3d body shape regression using metric and semantic attributes.
\newblock In \emph{Proceedings of the IEEE/CVF Conference on Computer Vision and Pattern Recognition}, pages 2718--2728, 2022.

\bibitem[Cuevas-Velasquez et~al.(2024)Cuevas-Velasquez, Hewitt, Aliakbarian, and Baltru{\v{s}}aitis]{cuevas2024simpleego}
Hanz Cuevas-Velasquez, Charlie Hewitt, Sadegh Aliakbarian, and Tadas Baltru{\v{s}}aitis.
\newblock Simpleego: Predicting probabilistic body pose from egocentric cameras.
\newblock In \emph{2024 International Conference on 3D Vision (3DV)}, pages 1446--1455. IEEE, 2024.

\bibitem[Dwivedi et~al.(2024)Dwivedi, Sun, Patel, Feng, and Black]{dwivedi2024tokenhmr}
Sai~Kumar Dwivedi, Yu Sun, Priyanka Patel, Yao Feng, and Michael~J Black.
\newblock Tokenhmr: Advancing human mesh recovery with a tokenized pose representation.
\newblock In \emph{Proceedings of the IEEE/CVF Conference on Computer Vision and Pattern Recognition}, pages 1323--1333, 2024.

\bibitem[{Epic Games}(2022)]{unreal}
{Epic Games}.
\newblock {Unreal Engine 5}.
\newblock \url{https://www.unrealengine.com}, 2022.

\bibitem[Gabeur et~al.(2019)Gabeur, Franco, Martin, Schmid, and Rogez]{gabeur2019moulding}
Valentin Gabeur, Jean-S{\'e}bastien Franco, Xavier Martin, Cordelia Schmid, and Gregory Rogez.
\newblock Moulding humans: Non-parametric 3d human shape estimation from single images.
\newblock In \emph{Proceedings of the IEEE/CVF international conference on computer vision}, pages 2232--2241, 2019.

\bibitem[Ge et~al.(2024)Ge, Wang, Chen, Liu, Chen, Wang, and Shen]{ge2024d}
Yongtao Ge, Wenjia Wang, Yongfan Chen, Yang Liu, Hao Chen, Xuan Wang, and Chunhua Shen.
\newblock 3d human reconstruction in the wild with synthetic data using generative models, 2024.

\bibitem[Goel et~al.(2023)Goel, Pavlakos, Rajasegaran, Kanazawa, and Malik]{goel2023humans}
Shubham Goel, Georgios Pavlakos, Jathushan Rajasegaran, Angjoo Kanazawa, and Jitendra Malik.
\newblock Humans in 4d: Reconstructing and tracking humans with transformers.
\newblock In \emph{Proceedings of the IEEE/CVF International Conference on Computer Vision}, pages 14783--14794, 2023.

\bibitem[He et~al.(2022)He, Sun, Yu, Xue, Zhang, Torr, Bai, and Qi]{he2022synthetic}
Ruifei He, Shuyang Sun, Xin Yu, Chuhui Xue, Wenqing Zhang, Philip Torr, Song Bai, and Xiaojuan Qi.
\newblock Is synthetic data from generative models ready for image recognition?
\newblock \emph{arXiv preprint arXiv:2210.07574}, 2022.

\bibitem[Heusel et~al.(2017)Heusel, Ramsauer, Unterthiner, Nessler, and Hochreiter]{heusel2017gans}
Martin Heusel, Hubert Ramsauer, Thomas Unterthiner, Bernhard Nessler, and Sepp Hochreiter.
\newblock Gans trained by a two time-scale update rule converge to a local nash equilibrium.
\newblock \emph{Advances in neural information processing systems}, 30, 2017.

\bibitem[Hewitt et~al.(2023)Hewitt, Baltru{\v{s}}aitis, Wood, Petikam, Florentin, and Velasquez]{hewitt2023procedural}
Charlie Hewitt, Tadas Baltru{\v{s}}aitis, Erroll Wood, Lohit Petikam, Louis Florentin, and Hanz~Cuevas Velasquez.
\newblock Procedural humans for computer vision.
\newblock \emph{arXiv preprint arXiv:2301.01161}, 2023.

\bibitem[Ho et~al.(2020)Ho, Jain, and Abbeel]{ho2020denoising}
Jonathan Ho, Ajay Jain, and Pieter Abbeel.
\newblock Denoising diffusion probabilistic models.
\newblock \emph{Advances in neural information processing systems}, 33:\penalty0 6840--6851, 2020.

\bibitem[Huang et~al.(2022)Huang, Yi, H{\"o}schle, Safroshkin, Alexiadis, Polikovsky, Scharstein, and Black]{huang2022capturing}
Chun-Hao~P Huang, Hongwei Yi, Markus H{\"o}schle, Matvey Safroshkin, Tsvetelina Alexiadis, Senya Polikovsky, Daniel Scharstein, and Michael~J Black.
\newblock Capturing and inferring dense full-body human-scene contact.
\newblock In \emph{Proceedings of the IEEE/CVF Conference on Computer Vision and Pattern Recognition}, pages 13274--13285, 2022.

\bibitem[Iqbal et~al.(2017)Iqbal, Milan, and Gall]{iqbal2017posetrack}
Umar Iqbal, Anton Milan, and Juergen Gall.
\newblock Posetrack: Joint multi-person pose estimation and tracking.
\newblock In \emph{Proceedings of the IEEE Conference on Computer Vision and Pattern Recognition}, pages 2011--2020, 2017.

\bibitem[Joo et~al.(2017)Joo, Simon, Li, Liu, Tan, Gui, Banerjee, Godisart, Nabbe, Matthews, et~al.]{joo2017panoptic}
Hanbyul Joo, Tomas Simon, Xulong Li, Hao Liu, Lei Tan, Lin Gui, Sean Banerjee, Timothy Godisart, Bart Nabbe, Iain Matthews, et~al.
\newblock Panoptic studio: A massively multiview system for social interaction capture. dec 2016.
\newblock \emph{URL http://arxiv. org/abs/1612.03153}, 2017.

\bibitem[Kanazawa et~al.(2018)Kanazawa, Black, Jacobs, and Malik]{kanazawa2018end}
Angjoo Kanazawa, Michael~J Black, David~W Jacobs, and Jitendra Malik.
\newblock End-to-end recovery of human shape and pose.
\newblock In \emph{Proceedings of the IEEE conference on computer vision and pattern recognition}, pages 7122--7131, 2018.

\bibitem[Kaufmann et~al.(2023)Kaufmann, Song, Guo, Shen, Jiang, Tang, Z{\'a}rate, and Hilliges]{kaufmann2023emdb}
Manuel Kaufmann, Jie Song, Chen Guo, Kaiyue Shen, Tianjian Jiang, Chengcheng Tang, Juan~Jos{\'e} Z{\'a}rate, and Otmar Hilliges.
\newblock Emdb: The electromagnetic database of global 3d human pose and shape in the wild.
\newblock In \emph{Proceedings of the IEEE/CVF International Conference on Computer Vision}, pages 14632--14643, 2023.

\bibitem[Kocabas et~al.(2021)Kocabas, Huang, Tesch, M{\"u}ller, Hilliges, and Black]{kocabas2021spec}
Muhammed Kocabas, Chun-Hao~P Huang, Joachim Tesch, Lea M{\"u}ller, Otmar Hilliges, and Michael~J Black.
\newblock Spec: Seeing people in the wild with an estimated camera.
\newblock In \emph{Proceedings of the IEEE/CVF International Conference on Computer Vision}, pages 11035--11045, 2021.

\bibitem[Li et~al.(2022)Li, Liu, Zhang, Xu, and Yan]{li2022cliff}
Zhihao Li, Jianzhuang Liu, Zhensong Zhang, Songcen Xu, and Youliang Yan.
\newblock Cliff: Carrying location information in full frames into human pose and shape estimation.
\newblock In \emph{European Conference on Computer Vision}, pages 590--606. Springer, 2022.

\bibitem[Loper et~al.(2015)Loper, Mahmood, Romero, Pons-Moll, and Black]{loper2015smpl}
Matthew Loper, Naureen Mahmood, Javier Romero, Gerard Pons-Moll, and Michael~J Black.
\newblock Smpl: A skinned multi-person linear model.
\newblock \emph{TOG}, 34\penalty0 (6):\penalty0 1--16, 2015.

\bibitem[Lugmayr et~al.(2022)Lugmayr, Danelljan, Romero, Yu, Timofte, and Van~Gool]{lugmayrinpainting}
Andreas Lugmayr, Martin Danelljan, Andres Romero, Fisher Yu, Radu Timofte, and Luc Van~Gool.
\newblock Repaint: Inpainting using denoising diffusion probabilistic models.
\newblock In \emph{Proceedings of the IEEE/CVF conference on computer vision and pattern recognition}, pages 11461--11471, 2022.

\bibitem[Mahmood et~al.(2019)Mahmood, Ghorbani, Troje, Pons-Moll, and Black]{mahmood2019amass}
Naureen Mahmood, Nima Ghorbani, Nikolaus~F Troje, Gerard Pons-Moll, and Michael~J Black.
\newblock Amass: Archive of motion capture as surface shapes.
\newblock In \emph{Proceedings of the IEEE/CVF international conference on computer vision}, pages 5442--5451, 2019.

\bibitem[Martin-Martin et~al.(2021)Martin-Martin, Patel, Rezatofighi, Shenoi, Gwak, Frankel, Sadeghian, and Savarese]{martin2021jrdb}
Roberto Martin-Martin, Mihir Patel, Hamid Rezatofighi, Abhijeet Shenoi, JunYoung Gwak, Eric Frankel, Amir Sadeghian, and Silvio Savarese.
\newblock Jrdb: A dataset and benchmark of egocentric robot visual perception of humans in built environments.
\newblock \emph{IEEE transactions on pattern analysis and machine intelligence}, 45\penalty0 (6):\penalty0 6748--6765, 2021.

\bibitem[Mehta et~al.(2017)Mehta, Rhodin, Casas, Fua, Sotnychenko, Xu, and Theobalt]{mehta2017monocular}
Dushyant Mehta, Helge Rhodin, Dan Casas, Pascal Fua, Oleksandr Sotnychenko, Weipeng Xu, and Christian Theobalt.
\newblock Monocular 3d human pose estimation in the wild using improved cnn supervision.
\newblock In \emph{2017 international conference on 3D vision (3DV)}, pages 506--516. IEEE, 2017.

\bibitem[Mehta et~al.(2018)Mehta, Sotnychenko, Mueller, Xu, Sridhar, Pons-Moll, and Theobalt]{mehta2018single}
Dushyant Mehta, Oleksandr Sotnychenko, Franziska Mueller, Weipeng Xu, Srinath Sridhar, Gerard Pons-Moll, and Christian Theobalt.
\newblock Single-shot multi-person 3d pose estimation from monocular rgb.
\newblock In \emph{2018 International Conference on 3D Vision (3DV)}, pages 120--130. IEEE, 2018.

\bibitem[Meng et~al.(2022)Meng, He, Song, Song, Wu, Zhu, and Ermon]{meng2021sdedit}
Chenlin Meng, Yutong He, Yang Song, Jiaming Song, Jiajun Wu, Jun{-}Yan Zhu, and Stefano Ermon.
\newblock Sdedit: Guided image synthesis and editing with stochastic differential equations.
\newblock In \emph{The Tenth International Conference on Learning Representations, {ICLR} 2022, Virtual Event, April 25-29, 2022}. OpenReview.net, 2022.

\bibitem[Patel et~al.(2021)Patel, Huang, Tesch, Hoffmann, Tripathi, and Black]{patel2021agora}
Priyanka Patel, Chun-Hao~P Huang, Joachim Tesch, David~T Hoffmann, Shashank Tripathi, and Michael~J Black.
\newblock Agora: Avatars in geography optimized for regression analysis.
\newblock In \emph{Proceedings of the IEEE/CVF Conference on Computer Vision and Pattern Recognition}, pages 13468--13478, 2021.

\bibitem[{Realistic Vision Team}(2023)]{realisticvision}
{Realistic Vision Team}.
\newblock {Realistic Vision V5.1 (VAE)}.
\newblock \url{https://civitai.com/models/4201?modelVersionId=130072}, 2023.

\bibitem[Rogez and Schmid(2016)]{rogez2016mocap}
Gr{\'e}gory Rogez and Cordelia Schmid.
\newblock Mocap-guided data augmentation for 3d pose estimation in the wild.
\newblock \emph{Advances in neural information processing systems}, 29, 2016.

\bibitem[Rombach et~al.(2022)Rombach, Blattmann, Lorenz, Esser, and Ommer]{rombach2022high}
Robin Rombach, Andreas Blattmann, Dominik Lorenz, Patrick Esser, and Bj{\"o}rn Ommer.
\newblock High-resolution image synthesis with latent diffusion models.
\newblock In \emph{Proceedings of the IEEE/CVF conference on computer vision and pattern recognition}, pages 10684--10695, 2022.

\bibitem[Ruiz et~al.(2023)Ruiz, Li, Jampani, Pritch, Rubinstein, and Aberman]{ruiz2023dreambooth}
Nataniel Ruiz, Yuanzhen Li, Varun Jampani, Yael Pritch, Michael Rubinstein, and Kfir Aberman.
\newblock Dreambooth: Fine tuning text-to-image diffusion models for subject-driven generation.
\newblock In \emph{Proceedings of the IEEE/CVF conference on computer vision and pattern recognition}, pages 22500--22510, 2023.

\bibitem[Sengupta et~al.(2020)Sengupta, Budvytis, and Cipolla]{STRAPS2018BMVC}
Akash Sengupta, Ignas Budvytis, and Roberto Cipolla.
\newblock Synthetic training for accurate 3d human pose and shape estimation in the wild.
\newblock In \emph{British Machine Vision Conference (BMVC)}, 2020.

\bibitem[Sigal et~al.(2010)Sigal, Balan, and Black]{sigal2010humaneva}
Leonid Sigal, Alexandru~O Balan, and Michael~J Black.
\newblock Humaneva: Synchronized video and motion capture dataset and baseline algorithm for evaluation of articulated human motion.
\newblock \emph{International journal of computer vision}, 87\penalty0 (1):\penalty0 4--27, 2010.

\bibitem[Song et~al.(2020)Song, Meng, and Ermon]{song2020denoising}
Jiaming Song, Chenlin Meng, and Stefano Ermon.
\newblock Denoising diffusion implicit models.
\newblock \emph{arXiv preprint arXiv:2010.02502}, 2020.

\bibitem[Trumble et~al.(2017)Trumble, Gilbert, Malleson, Hilton, and Collomosse]{trumble2017total}
Matthew Trumble, Andrew Gilbert, Charles Malleson, Adrian Hilton, and John~P Collomosse.
\newblock Total capture: 3d human pose estimation fusing video and inertial sensors.
\newblock In \emph{BMVC}, pages 1--13. London, UK, 2017.

\bibitem[Varol et~al.(2017)Varol, Romero, Martin, Mahmood, Black, Laptev, and Schmid]{varol17_surreal}
G{\"u}l Varol, Javier Romero, Xavier Martin, Naureen Mahmood, Michael~J. Black, Ivan Laptev, and Cordelia Schmid.
\newblock Learning from synthetic humans.
\newblock In \emph{CVPR}, 2017.

\bibitem[Von~Marcard et~al.(2018)Von~Marcard, Henschel, Black, Rosenhahn, and Pons-Moll]{von2018recovering}
Timo Von~Marcard, Roberto Henschel, Michael~J Black, Bodo Rosenhahn, and Gerard Pons-Moll.
\newblock Recovering accurate 3d human pose in the wild using imus and a moving camera.
\newblock In \emph{Proceedings of the European conference on computer vision (ECCV)}, pages 601--617, 2018.

\bibitem[Weng et~al.(2024)Weng, Bravo-S{\'a}nchez, and Yeung-Levy]{weng2024diffusion}
Zhenzhen Weng, Laura Bravo-S{\'a}nchez, and Serena Yeung-Levy.
\newblock Diffusion-hpc: Synthetic data generation for human mesh recovery in challenging domains.
\newblock In \emph{2024 International Conference on 3D Vision (3DV)}, pages 257--267. IEEE, 2024.

\bibitem[Wu et~al.(2023)Wu, Zhao, Chen, Gu, Zhao, He, Zhou, Shou, and Shen]{wu2023datasetdm}
Weijia Wu, Yuzhong Zhao, Hao Chen, Yuchao Gu, Rui Zhao, Yefei He, Hong Zhou, Mike~Zheng Shou, and Chunhua Shen.
\newblock Datasetdm: Synthesizing data with perception annotations using diffusion models.
\newblock \emph{Advances in Neural Information Processing Systems}, 36:\penalty0 54683--54695, 2023.

\bibitem[Zanfir et~al.(2020)Zanfir, Oneata, Popa, Zanfir, and Sminchisescu]{zanfir2020human}
Mihai Zanfir, Elisabeta Oneata, Alin-Ionut Popa, Andrei Zanfir, and Cristian Sminchisescu.
\newblock Human synthesis and scene compositing.
\newblock In \emph{Proceedings of the AAAI Conference on Artificial Intelligence}, pages 12749--12756, 2020.

\bibitem[Zhang et~al.(2023)Zhang, Rao, and Agrawala]{zhang2023adding}
Lvmin Zhang, Anyi Rao, and Maneesh Agrawala.
\newblock Adding conditional control to text-to-image diffusion models.
\newblock In \emph{Proceedings of the IEEE/CVF International Conference on Computer Vision}, pages 3836--3847, 2023.

\bibitem[Zhao et~al.(2023)Zhao, Chen, Chen, Bao, Hao, Yuan, and Wong]{zhao2023uni}
Shihao Zhao, Dongdong Chen, Yen-Chun Chen, Jianmin Bao, Shaozhe Hao, Lu Yuan, and Kwan-Yee~K Wong.
\newblock Uni-controlnet: All-in-one control to text-to-image diffusion models.
\newblock \emph{arXiv preprint arXiv:2305.16322}, 2023.

\end{thebibliography}
}
\clearpage
\setcounter{page}{1}
\maketitlesupplementary

\section{Pre-processing}
\label{sec:preproc_hypermaram}
\paragraph{Cropping.} We separately process the head, hair, clothes, and shoes for each person in the image. To select the person, we use the label indices that BEDLAM provides. For each index, we use the segmentation masks from BEDLAM, or the ones we generated, and crop the image around that mask. The crop size is $512\times512$, as we use Stable-Diffusion~\cite{rombach2022high}. 

\begin{figure}[htbp!]
    \centering
    \begin{subfigure}{0.48\linewidth}
        \centering
        \includegraphics[width=\linewidth]{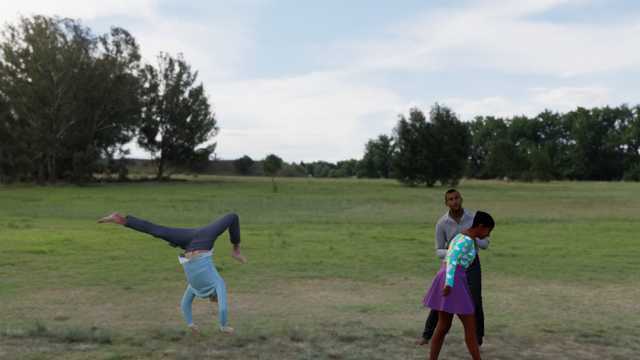}
        \caption{Color image.}
    \end{subfigure}
    \hfill
    \begin{subfigure}{0.48\linewidth}
        \centering
        \includegraphics[width=\linewidth]{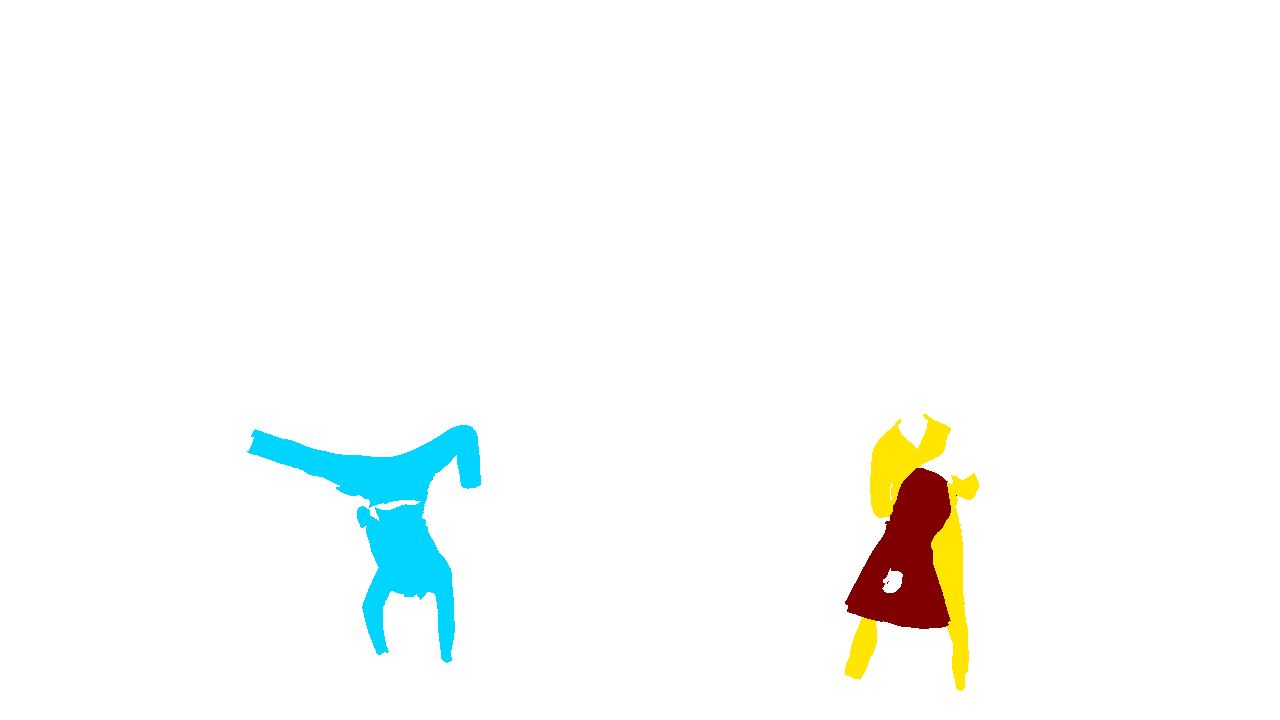}
        \caption{Clothes masks.}
    \end{subfigure}
    
    \vspace{1em} 
    
    \begin{subfigure}{0.48\linewidth}
        \centering
        \includegraphics[width=\linewidth]{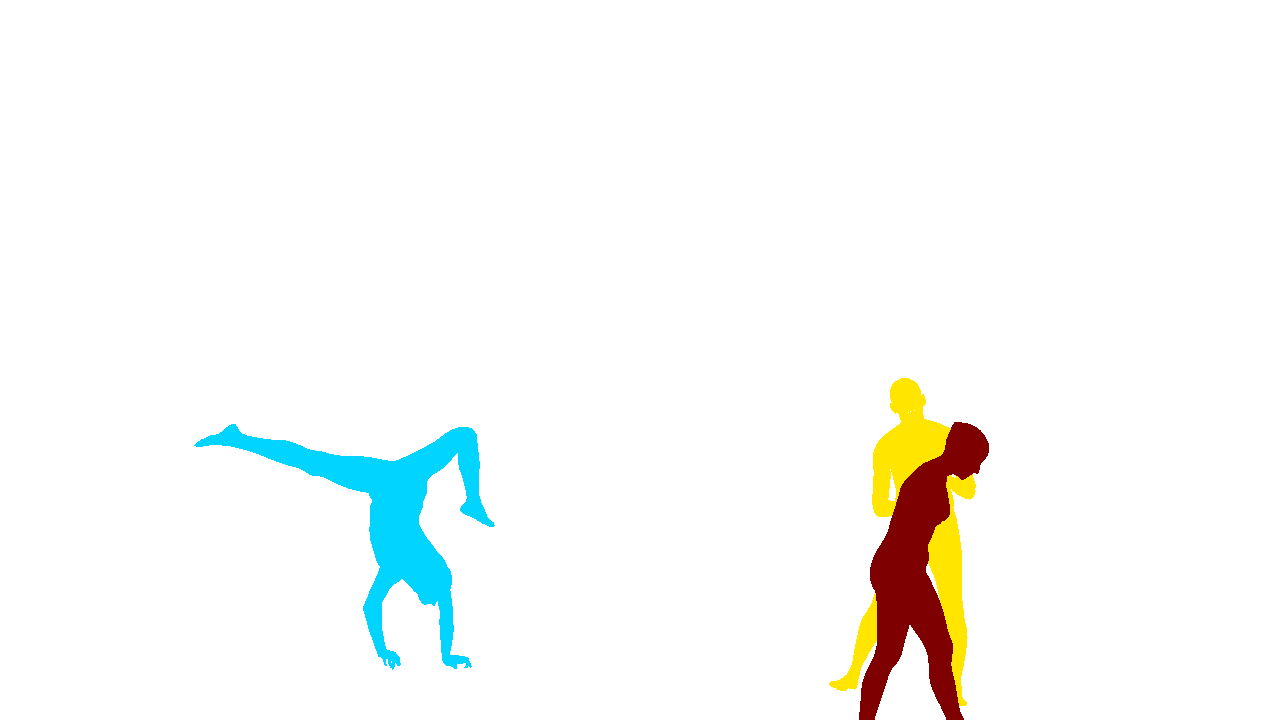}
        \caption{SMPL body masks.}
        \label{fig:smpl_masks_occluded}
    \end{subfigure}
    \hfill
    \begin{subfigure}{0.48\linewidth}
        \centering
        \includegraphics[width=\linewidth]{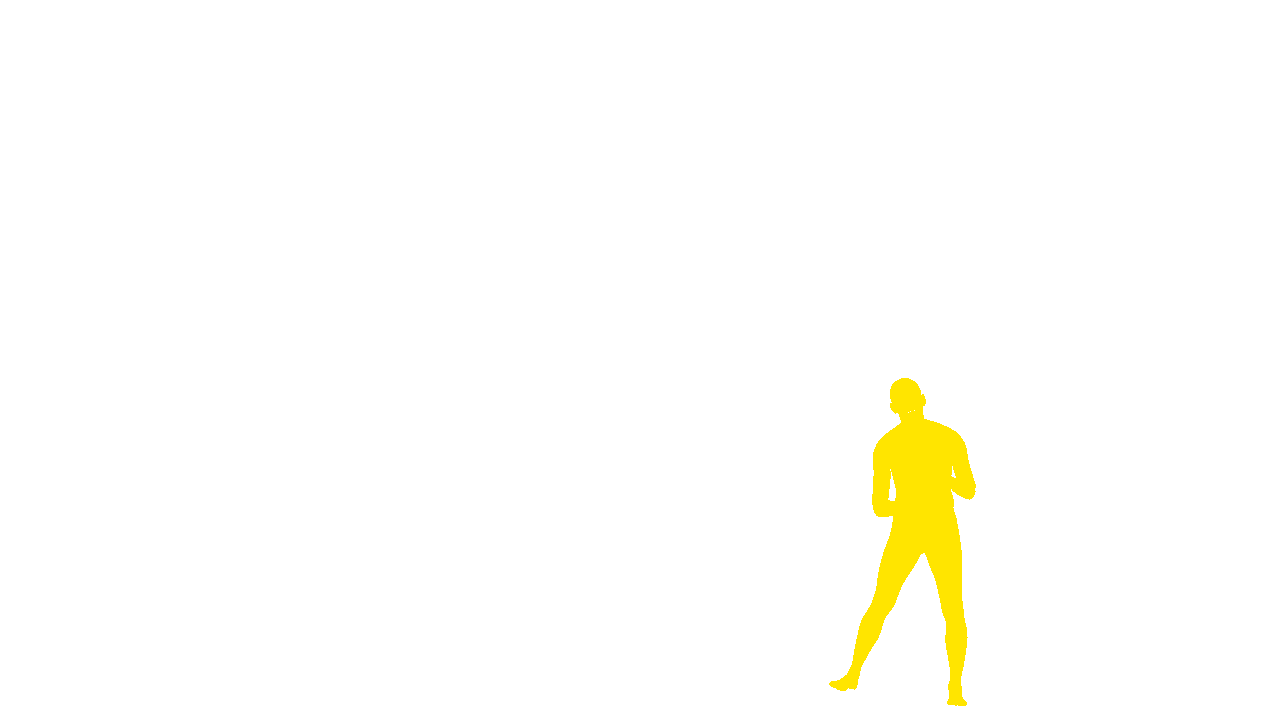}
        \caption{SMPL body mask without occlusion of yellow body.}
        \label{fig:smpl_masks_no_occluded}
    \end{subfigure}
    
    \vspace{1em} 

    \begin{subfigure}{0.48\linewidth}
        \centering
        \includegraphics[width=\linewidth]{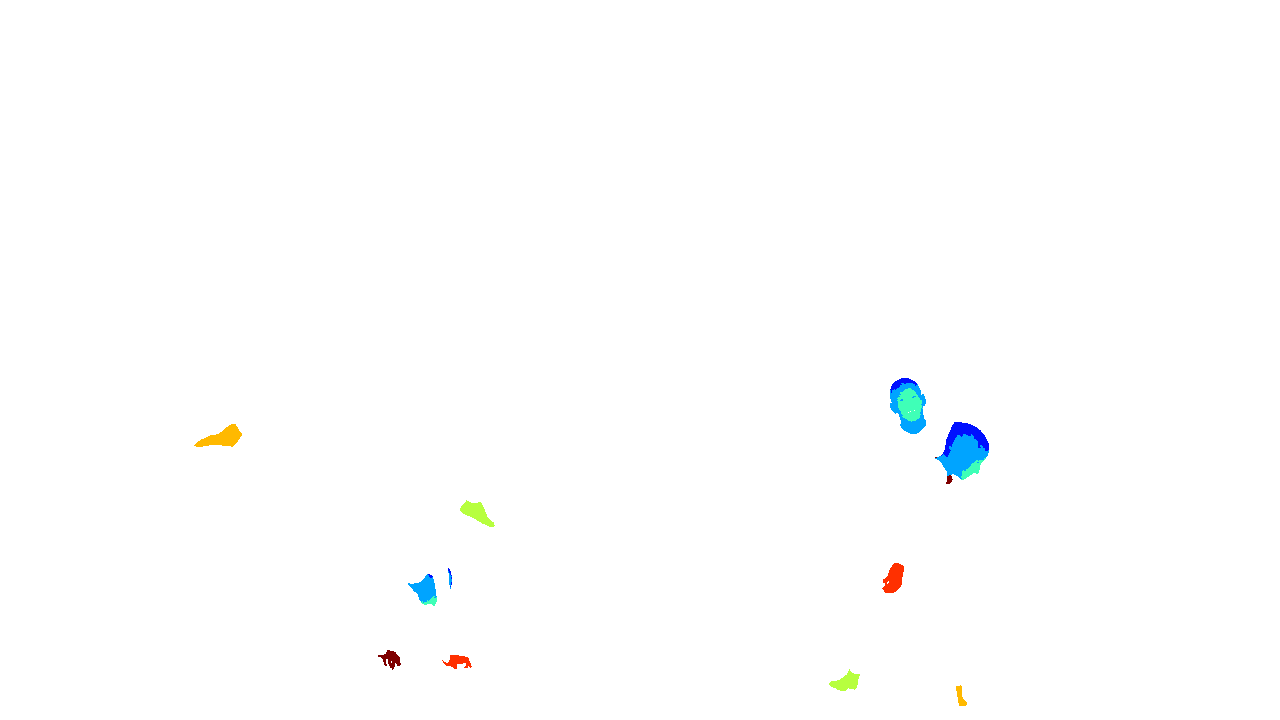}
        \caption{Hair, head, hands, and feet masks.}
        \label{fig:colored_part_masks}
    \end{subfigure}
    \hfill
    \begin{subfigure}{0.48\linewidth}
        \centering
        \includegraphics[width=\linewidth]{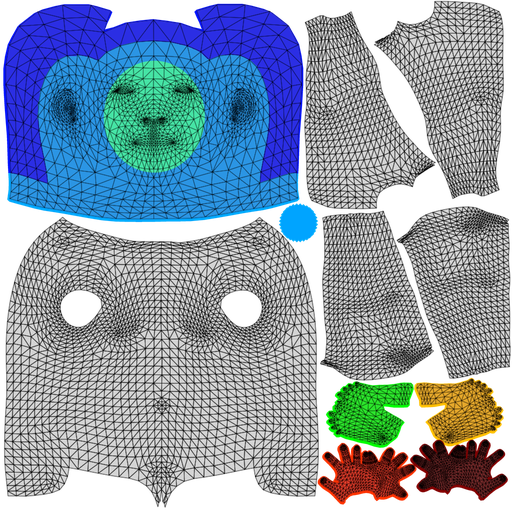}
        \caption{UV map with labels.}
    \end{subfigure}

    \caption{Color image and clothes masks (a) and (b) are provided by the BEDLAM dataset~\cite{black2023bedlam}. We use the SMPL-X mesh to render the labeled body masks (c), body masks without occlusion (d), and (e) body parts. The body parts were obtained by segmenting the UV map (f).}
    \label{fig:sup_mat_masks}
\end{figure}

\paragraph{What bodies should we process?} The BEDLAM dataset includes images of people occluded by objects from the 3D environment or other people. To avoid modifying those regions in the image, which would add noise to the GT training data, we only consider a body valid if more than $80\%$ of it is visible. We calculate the visibility percentage by dividing the total number of pixels of the body with occlusion by the number of pixels without occlusion. The SMPL masks with and without occlusion can be seen in \cref{fig:smpl_masks_occluded,fig:smpl_masks_no_occluded}. 

When generating the hair, we observed that SD tends to create faces on the back of the head if the face is barely visible. To avoid this, we do not process the hair mask if a significant portion of the face is not visible. In practice, we do not generate the hair for bodies where the rendered face mask (cyan color in \cref{fig:colored_part_masks}) is less than 100 pixels. We use the same threshold when processing the head.

\paragraph{2D pose as a control signal.} We also observed that when the face is barely visible, either due to occlusion or because it is facing away from the camera, we need to remove the 2D joints associated with the face before using the 2D pose as a control signal, as it causes the diffusion model to struggle in discerning the front and back of people. If these joints are included, the generated image is more likely to show a face on the back of the head or create the appearance of a head on the occluding person, making it look like they have two heads.

\section{HPS Predictions}
\begin{figure*}[htbp!]
\centering
\begin{minipage}{0.3\linewidth}
    \centering
    \includegraphics[width=\linewidth]{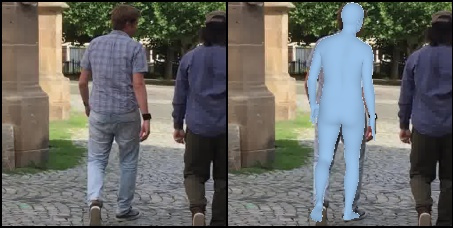}
\end{minipage}
\begin{minipage}{0.3\linewidth}
    \centering
    \includegraphics[width=\linewidth]{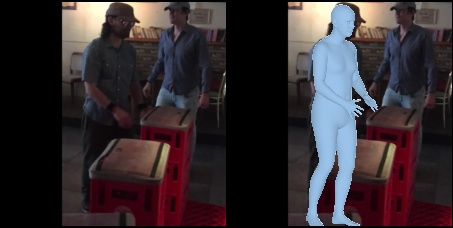}
\end{minipage}
\begin{minipage}{0.3\linewidth}
    \centering
    \includegraphics[width=\linewidth]{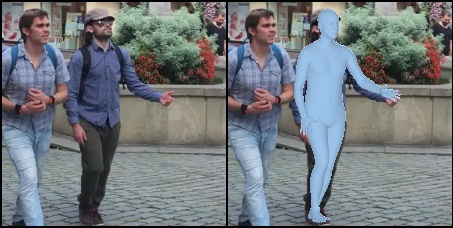}
\end{minipage}

\vspace{0.2cm} 

\begin{minipage}{0.3\linewidth}
    \centering
    \includegraphics[width=\linewidth]{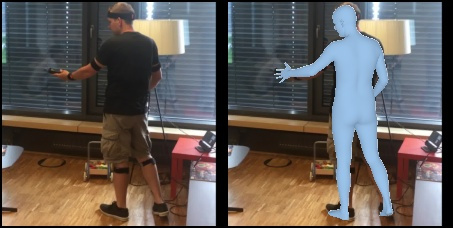}
\end{minipage}
\begin{minipage}{0.3\linewidth}
    \centering
    \includegraphics[width=\linewidth]{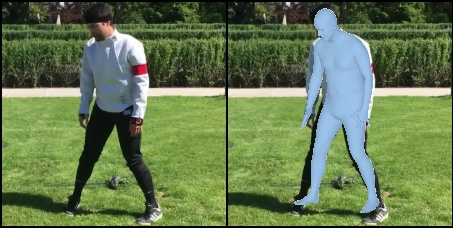}
\end{minipage}
\begin{minipage}{0.3\linewidth}
    \centering
    \includegraphics[width=\linewidth]{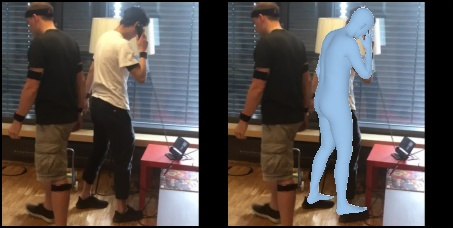}
\end{minipage}

\vspace{0.2cm} 

\begin{minipage}{0.3\linewidth}
    \centering
    \includegraphics[width=\linewidth]{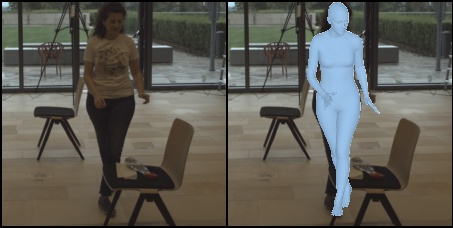}
\end{minipage}
\begin{minipage}{0.3\linewidth}
    \centering
    \includegraphics[width=\linewidth]{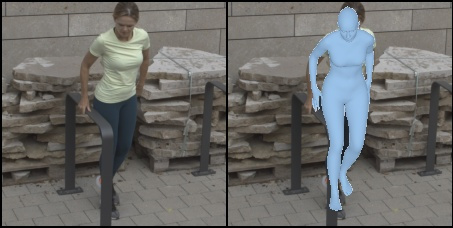}
\end{minipage}
\begin{minipage}{0.3\linewidth}
    \centering
    \includegraphics[width=\linewidth]{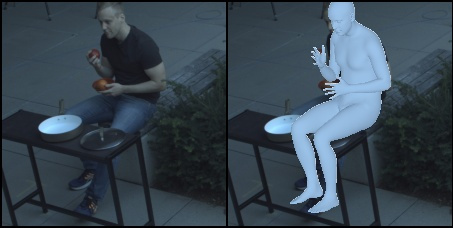}
\end{minipage}

\caption{HMR outputs for the 3DPW, RICH, and EMDB test set, each corresponding to a row, respectively.}
\label{fig:hmr_output}
\end{figure*}

\begin{figure*}[htbp!]
\centering
\begin{minipage}{0.3\linewidth}
    \centering
    \includegraphics[width=\linewidth]{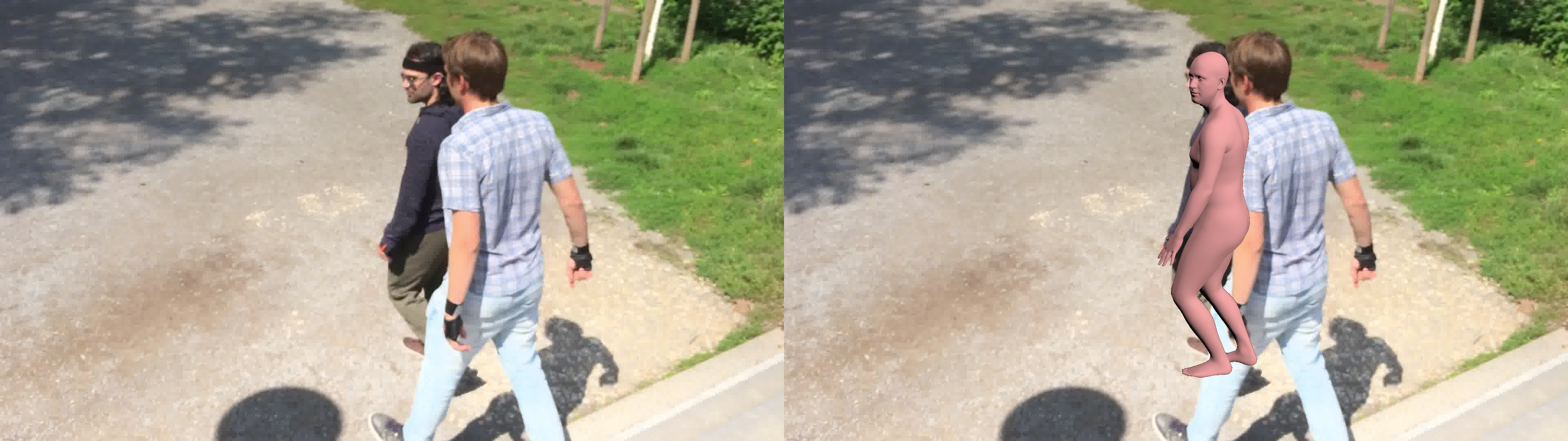}
\end{minipage}
\begin{minipage}{0.3\linewidth}
    \centering
    \includegraphics[width=\linewidth]{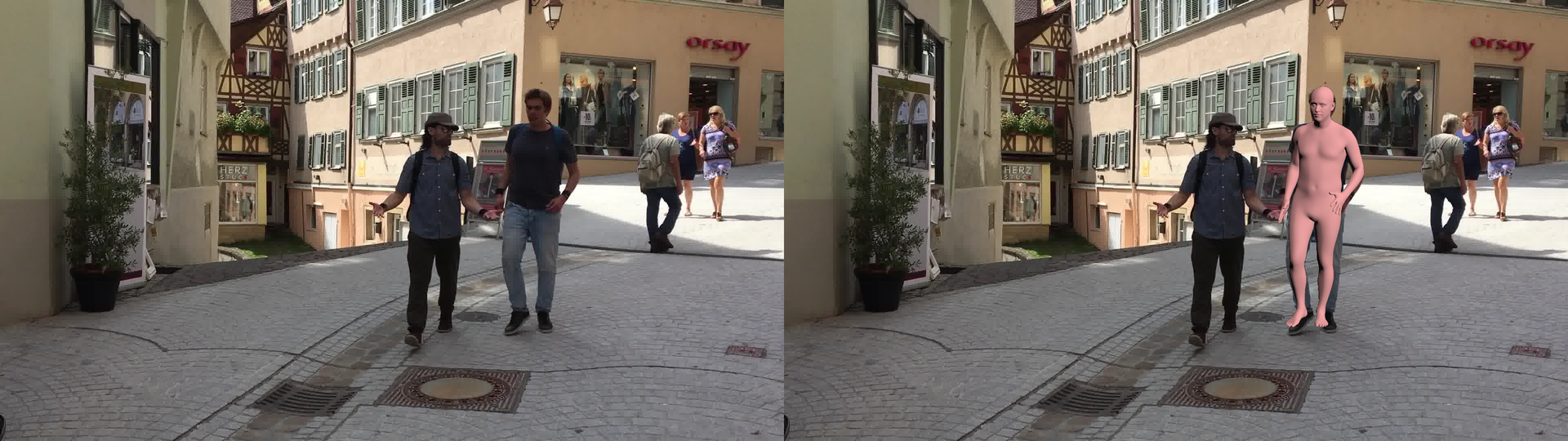}
\end{minipage}
\begin{minipage}{0.3\linewidth}
    \centering
    \includegraphics[width=\linewidth]{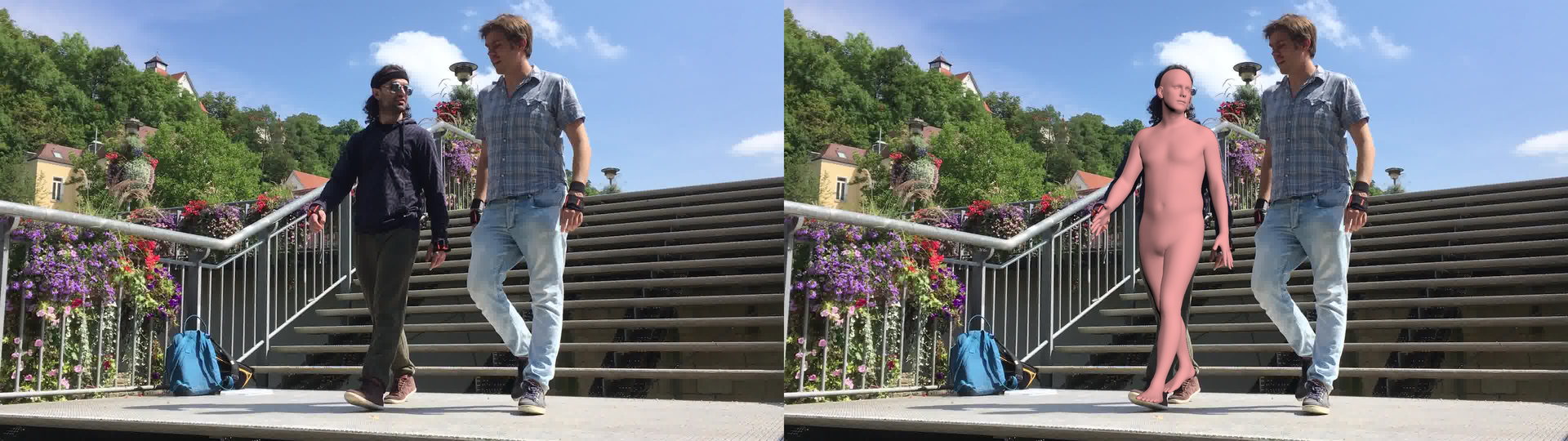}
\end{minipage}

\vspace{0.2cm} 

\begin{minipage}{0.3\linewidth}
    \centering
    \includegraphics[width=\linewidth]{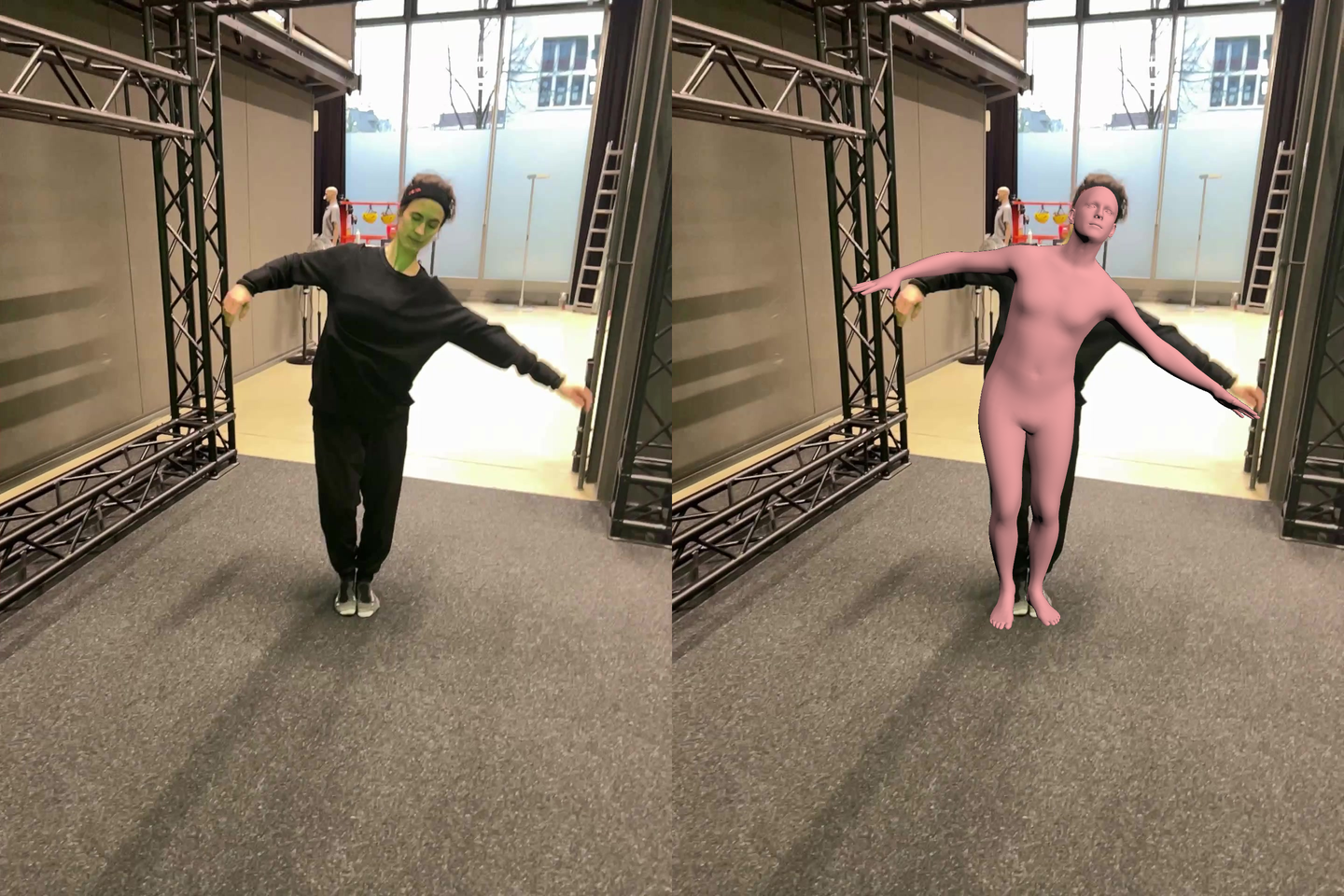}
\end{minipage}
\begin{minipage}{0.3\linewidth}
    \centering
    \includegraphics[width=\linewidth]{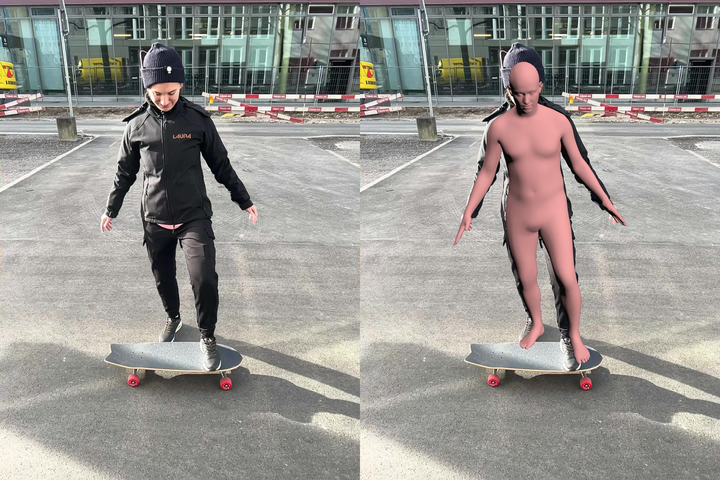}
\end{minipage}
\begin{minipage}{0.3\linewidth}
    \centering
    \includegraphics[width=\linewidth]{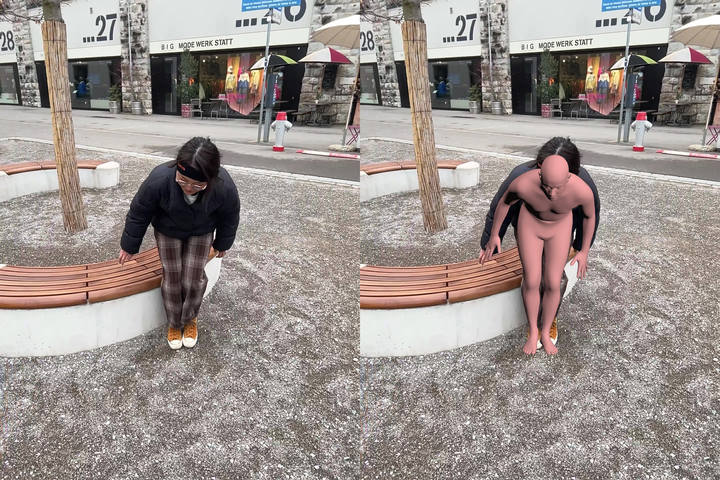}
\end{minipage}

\vspace{0.2cm} 

\begin{minipage}{0.3\linewidth}
    \centering
    \includegraphics[width=\linewidth]{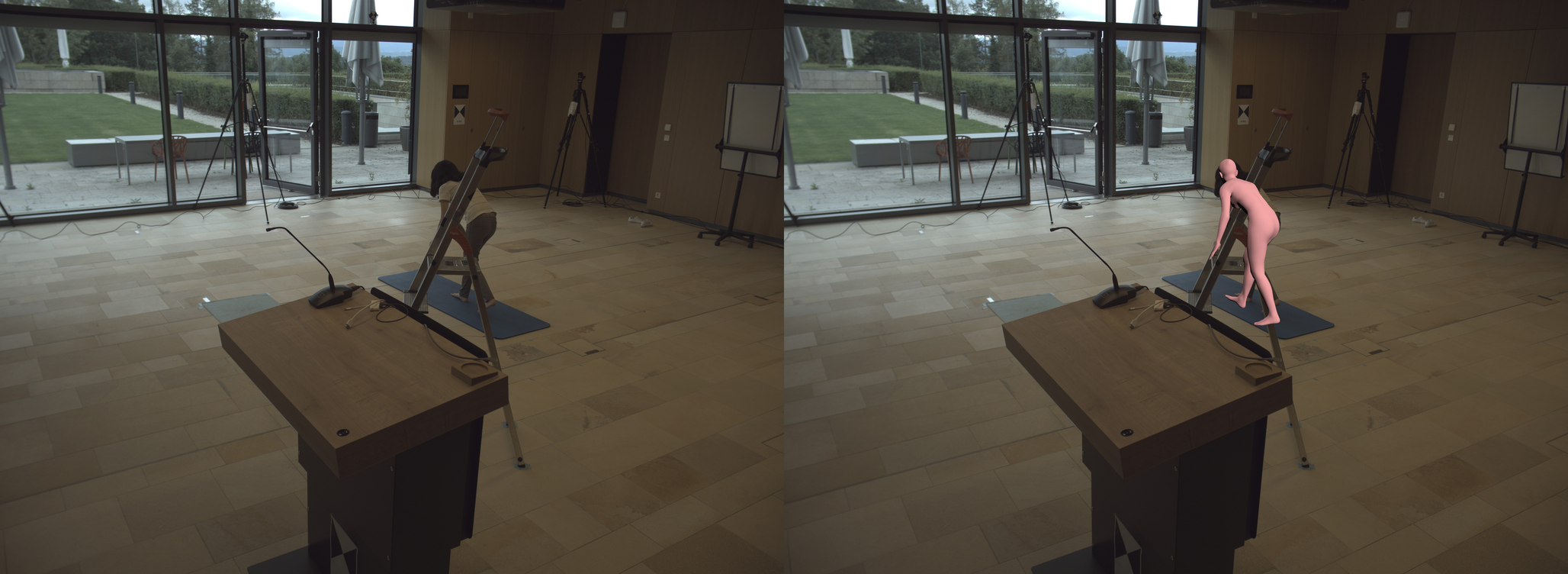}
\end{minipage}
\begin{minipage}{0.3\linewidth}
    \centering
    \includegraphics[width=\linewidth]{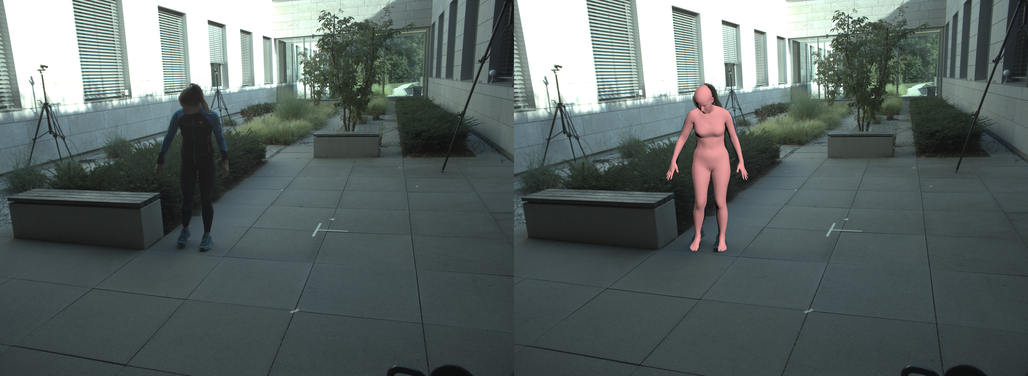}
\end{minipage}
\begin{minipage}{0.3\linewidth}
    \centering
    \includegraphics[width=\linewidth]{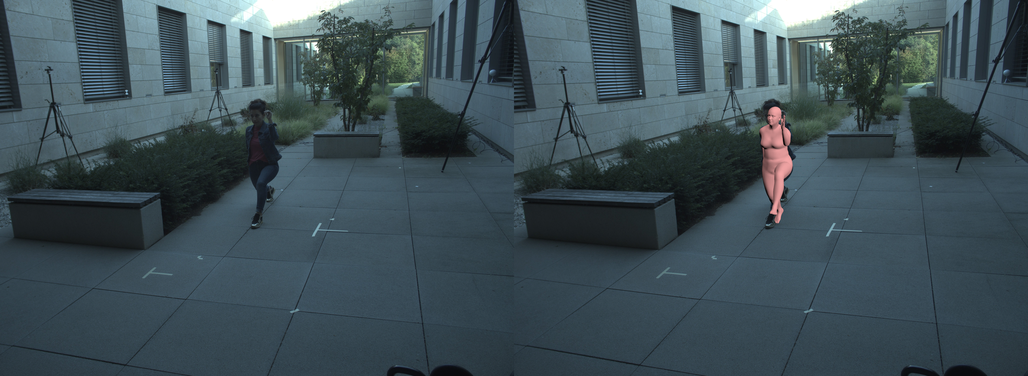}
\end{minipage}

\caption{CLIFF outputs for the 3DPW, RICH, and EMDB test set, each corresponding to a row, respectively.}
\label{fig:cliff_output}
\end{figure*}

\begin{figure*}[htbp!]
\centering
\begin{minipage}{0.3\linewidth}
    \centering
    \includegraphics[width=\linewidth]{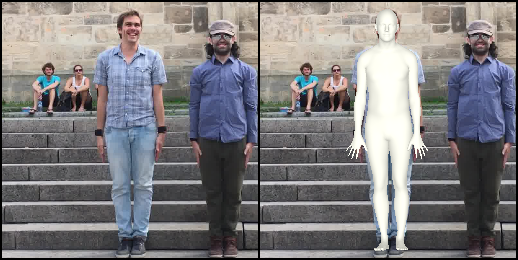}
\end{minipage}
\begin{minipage}{0.3\linewidth}
    \centering
    \includegraphics[width=\linewidth]{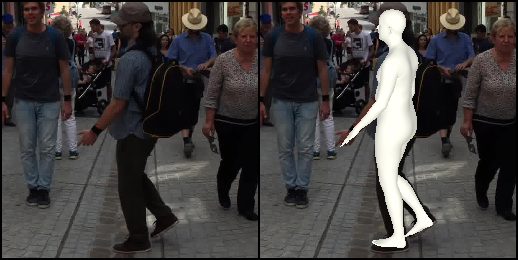}
\end{minipage}
\begin{minipage}{0.3\linewidth}
    \centering
    \includegraphics[width=\linewidth]{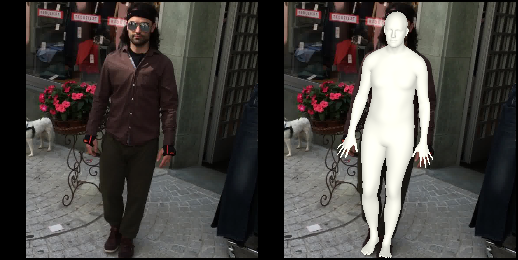}
\end{minipage}

\vspace{0.2cm} 

\begin{minipage}{0.3\linewidth}
    \centering
    \includegraphics[width=\linewidth]{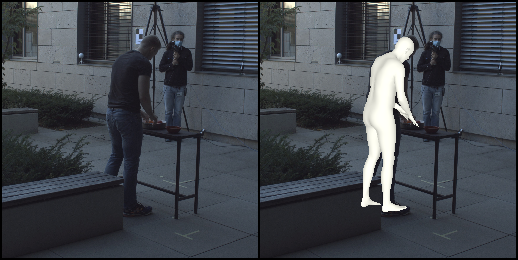}
\end{minipage}
\begin{minipage}{0.3\linewidth}
    \centering
    \includegraphics[width=\linewidth]{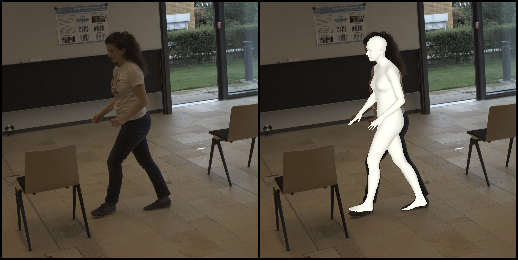}
\end{minipage}
\begin{minipage}{0.3\linewidth}
    \centering
    \includegraphics[width=\linewidth]{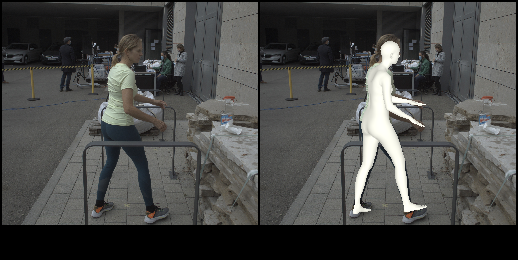}
\end{minipage}

\vspace{0.2cm} 

\begin{minipage}{0.3\linewidth}
    \centering
    \includegraphics[width=\linewidth]{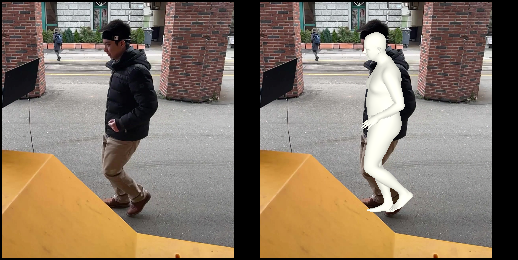}
\end{minipage}
\begin{minipage}{0.3\linewidth}
    \centering
    \includegraphics[width=\linewidth]{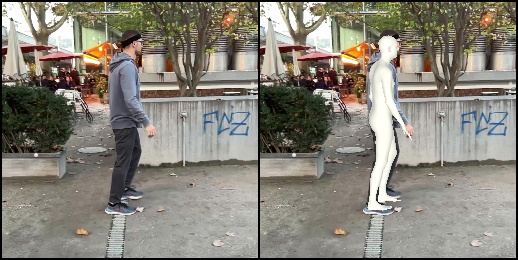}
\end{minipage}
\begin{minipage}{0.3\linewidth}
    \centering
    \includegraphics[width=\linewidth]{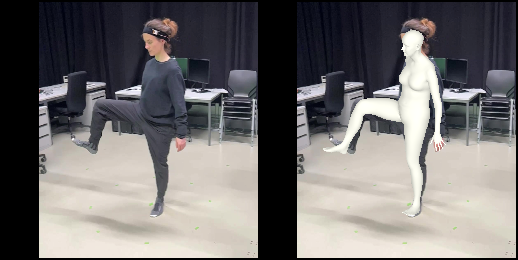}
\end{minipage}

\caption{HMR2.0 outputs for the 3DPW, RICH, and EMDB test set, each corresponding to a row, respectively.}
\label{fig:hmr2_output}
\end{figure*}

In \cref{fig:hmr_output,fig:cliff_output,fig:hmr2_output}, we provide qualitative results for HMR, CLIFF, and HMR2.0 predictions trained on our Gen-B dataset on 3DPW, EMDB, and RICH datasets.

\section{Noise Level Errors}
\begin{figure*}
\begin{center}
   \includegraphics[width=\linewidth]{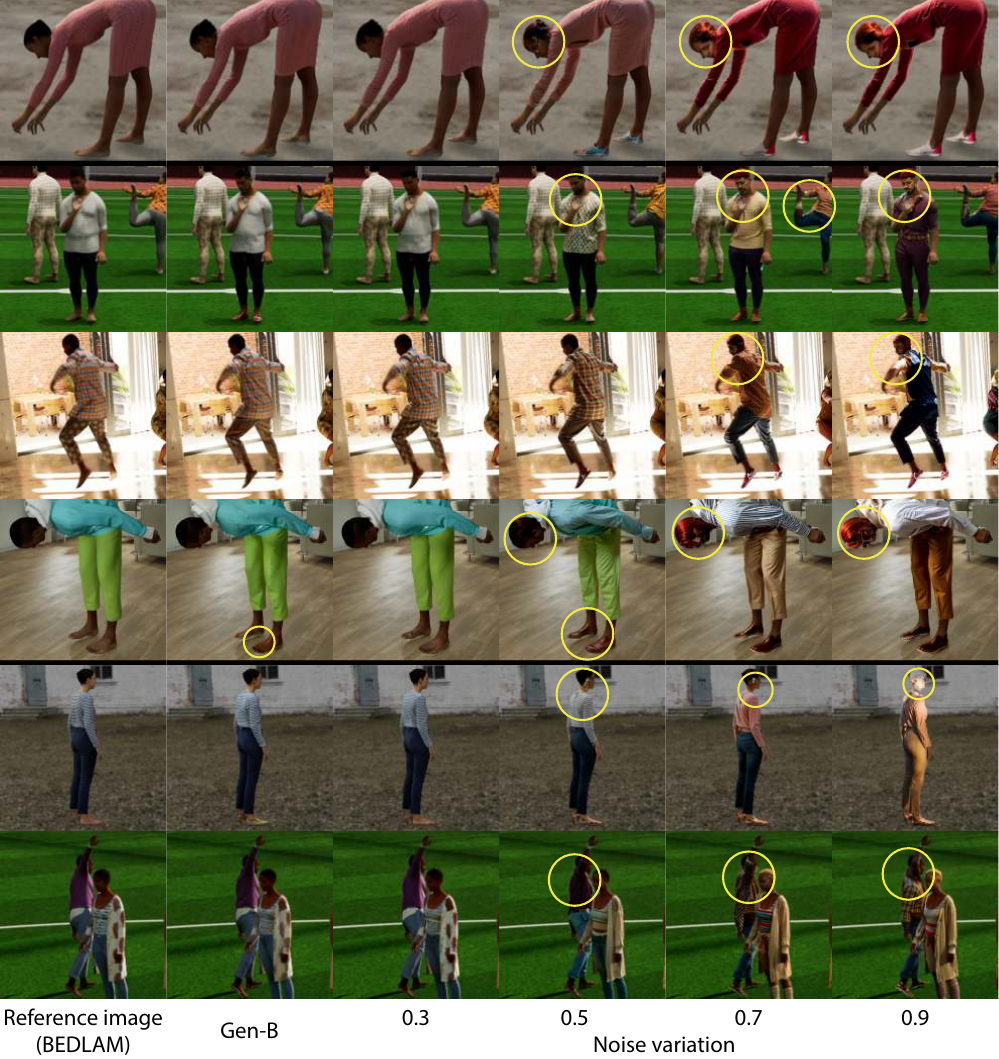}
\end{center}
   \caption{\textbf{Noise errors.} We present examples with different noise levels and also Gen-B output. We can see that higher noise levels allow for more diversity in generated people, clothes, hair, and shoes. However, this also increases the likelihood of artifacts, such as those seen in row 5, or unintended pose changes, as shown in rows 1 to 4. It can also generate extra body parts, as in the last row. Gen-B is not exempt from these artifacts, for instance, low noise in the shoes sometimes results in finger-like shapes at the tips}
\label{fig:sup_mat_noise_errors}
\end{figure*}

In \cref{fig:sup_mat_noise_errors}, we show more artifacts that can be generated while varying the noise levels.

\section{Qualitative Results of Gen-B}
\label{sec:Qualitative_results}

\begin{figure*}
\centering
\resizebox{0.9\linewidth}{!}{%
    \begin{minipage}{\linewidth}
        \centering
        \begin{subfigure}[t]{0.48\linewidth}
            \includegraphics[width=\linewidth]{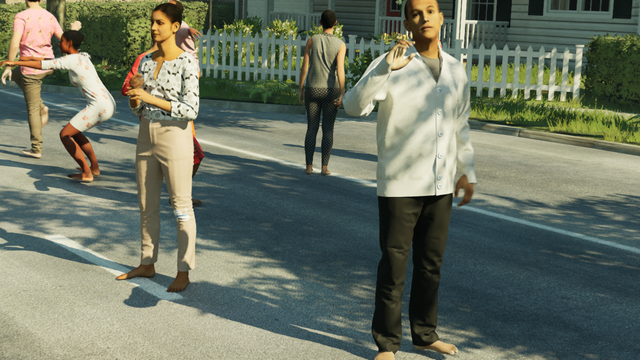}
        \end{subfigure}
        \begin{subfigure}[t]{0.48\linewidth}
            \includegraphics[width=\linewidth]{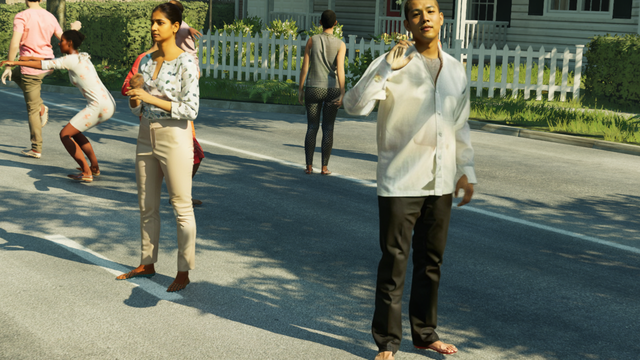}
        \end{subfigure}

        \begin{subfigure}[t]{0.48\linewidth}
            \includegraphics[width=\linewidth]{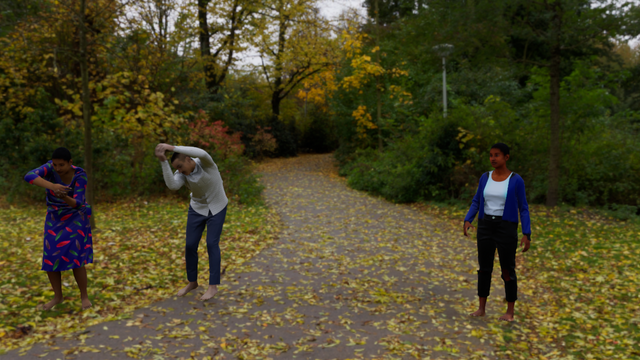}
        \end{subfigure}
        \begin{subfigure}[t]{0.48\linewidth}
            \includegraphics[width=\linewidth]{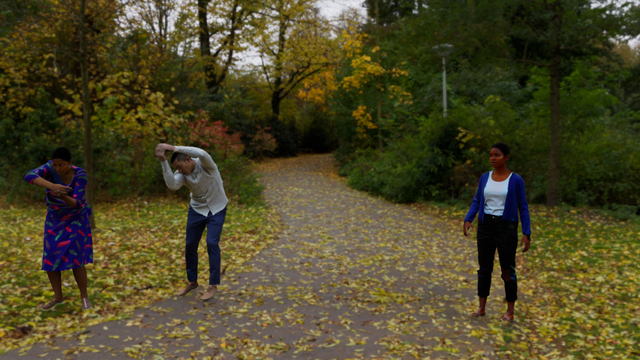}
        \end{subfigure}

        \begin{subfigure}[t]{0.48\linewidth}
            \includegraphics[width=\linewidth]{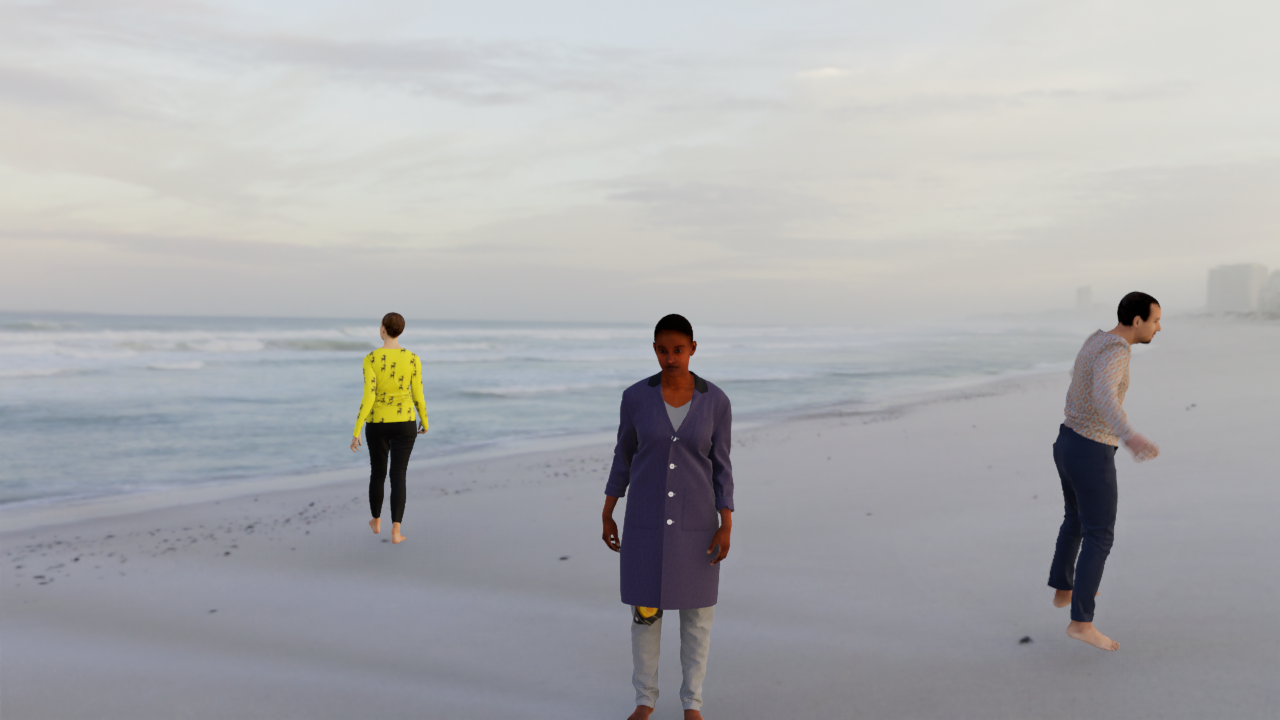}
        \end{subfigure}
        \begin{subfigure}[t]{0.48\linewidth}
            \includegraphics[width=\linewidth]{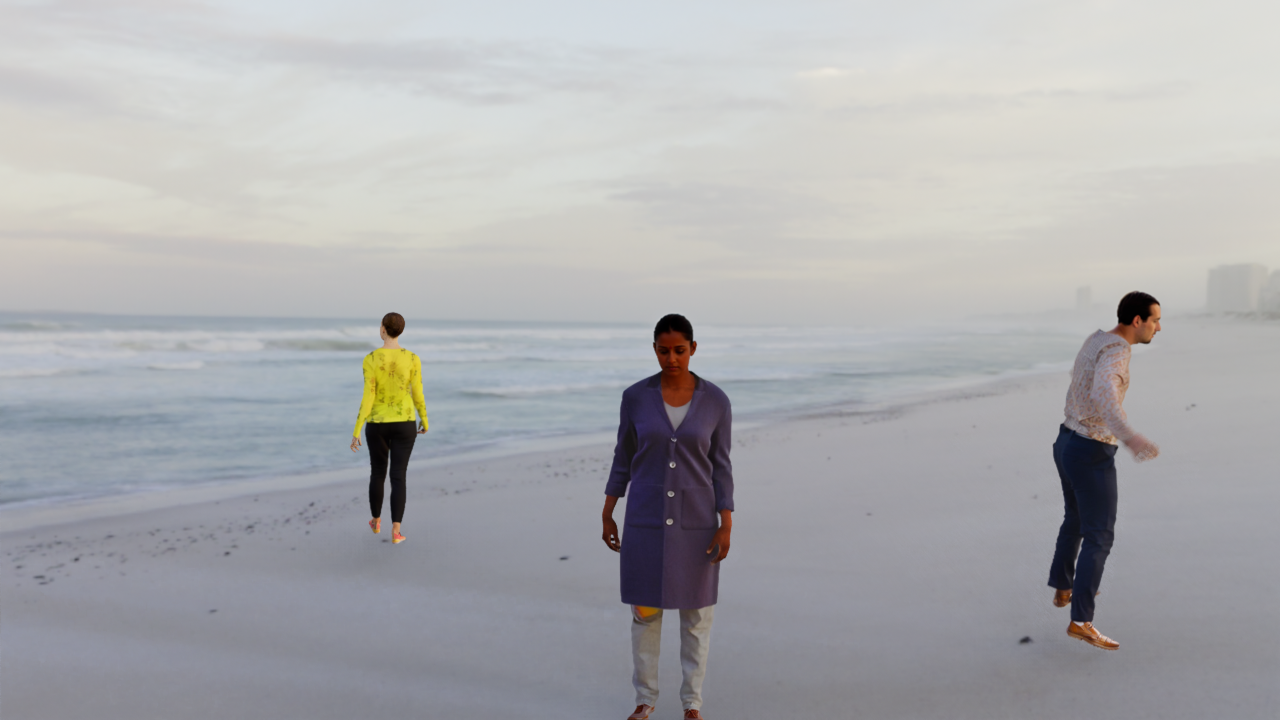}
        \end{subfigure}

        \begin{subfigure}[t]{0.48\linewidth}
            \includegraphics[width=\linewidth]{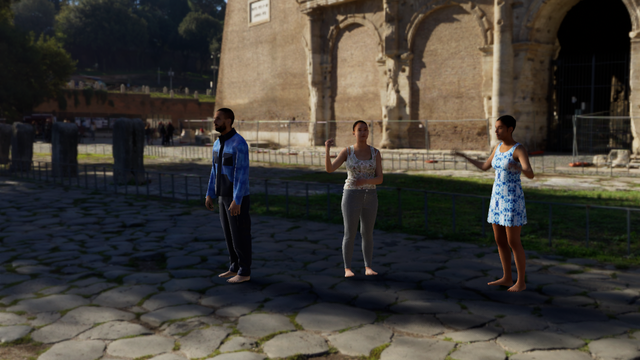}
        \end{subfigure}
        \begin{subfigure}[t]{0.48\linewidth}
            \includegraphics[width=\linewidth]{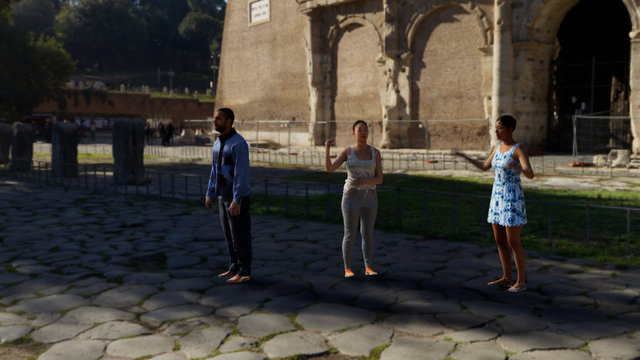}
        \end{subfigure}

        \begin{subfigure}[t]{0.48\linewidth}
            \includegraphics[width=\linewidth]{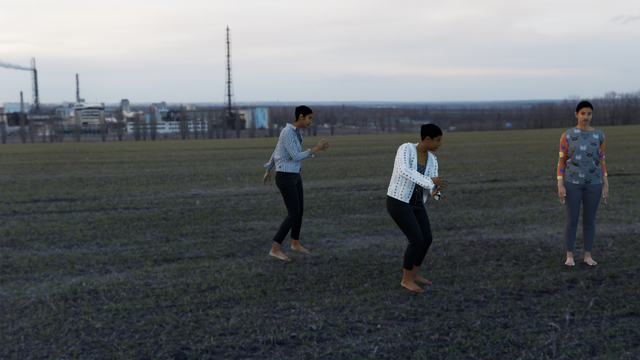}
        \end{subfigure}
        \begin{subfigure}[t]{0.48\linewidth}
            \includegraphics[width=\linewidth]{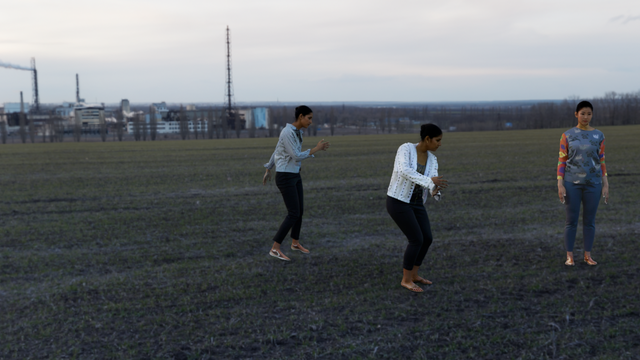}
        \end{subfigure}
    \end{minipage}%
}
\caption{Bedlam vs Gen-B.}
\label{fig:bedlam_gen_examples}
\end{figure*}

\begin{figure*}
\centering
\resizebox{0.9\linewidth}{!}{%
    \begin{minipage}{\linewidth}
        \centering
        \begin{subfigure}[t]{0.48\linewidth}
            \includegraphics[width=\linewidth]{img/ori_bedlam/_is_cluster_fast_scratch_hcuevas_20221010_3-10_500_batch01hand_zoom_suburb_dseq_000000_seq_000000_0080.png}
        \end{subfigure}
        \begin{subfigure}[t]{0.48\linewidth}
            \includegraphics[width=\linewidth]{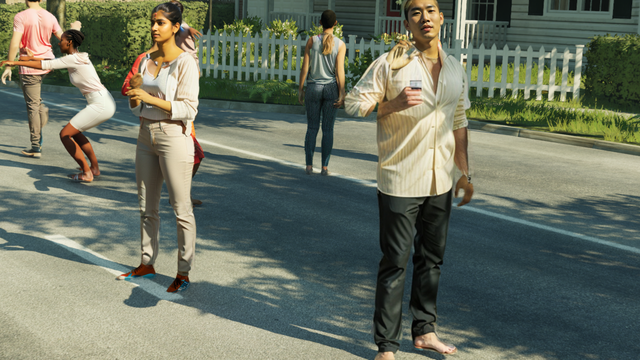}
        \end{subfigure}
        \begin{subfigure}[t]{0.48\linewidth}
            \includegraphics[width=\linewidth]{img/ori_bedlam/_is_cluster_fast_scratch_hcuevas_20221010_3_1000_batch01hand_6fpsseq_000021_seq_000021_0210.png}
        \end{subfigure}
        \begin{subfigure}[t]{0.48\linewidth}
            \includegraphics[width=\linewidth]{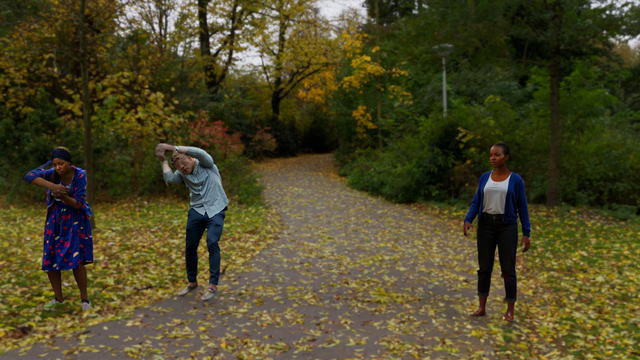}
        \end{subfigure}
        \begin{subfigure}[t]{0.48\linewidth}
            \includegraphics[width=\linewidth]{img/ori_bedlam/_is_cluster_fast_scratch_hcuevas_20221010_3_1000_batch01hand_6fpsseq_000016_seq_000016_0115.png}
        \end{subfigure}
        \begin{subfigure}[t]{0.48\linewidth}
            \includegraphics[width=\linewidth]{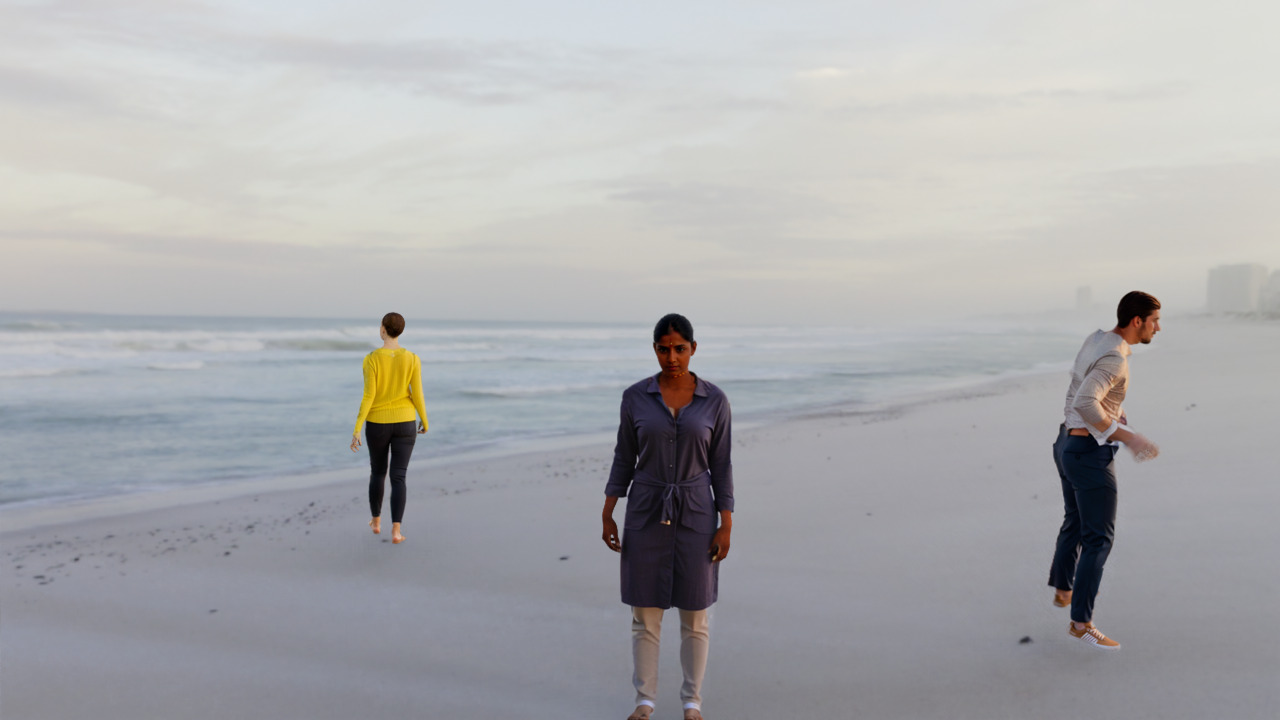}
        \end{subfigure}
        \begin{subfigure}[t]{0.48\linewidth}
            \includegraphics[width=\linewidth]{img/ori_bedlam/_is_cluster_fast_scratch_hcuevas_20221010_3_1000_batch01hand_6fpsseq_000022_seq_000022_0030.png}
        \end{subfigure}
        \begin{subfigure}[t]{0.48\linewidth}
            \includegraphics[width=\linewidth]{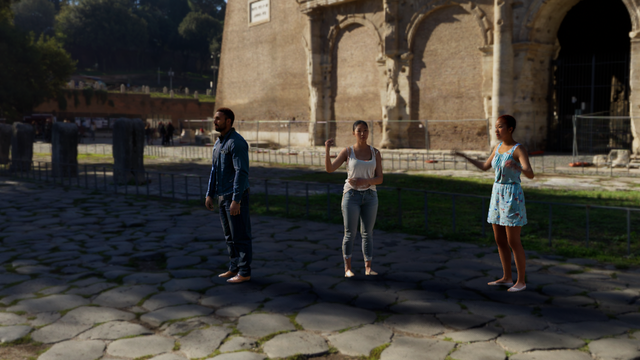}
        \end{subfigure}
        \begin{subfigure}[t]{0.48\linewidth}
            \includegraphics[width=\linewidth]{img/ori_bedlam/_is_cluster_fast_scratch_hcuevas_20221010_3_1000_batch01hand_6fpsseq_000033_seq_000033_0190.png}
        \end{subfigure}
        \begin{subfigure}[t]{0.48\linewidth}
            \includegraphics[width=\linewidth]{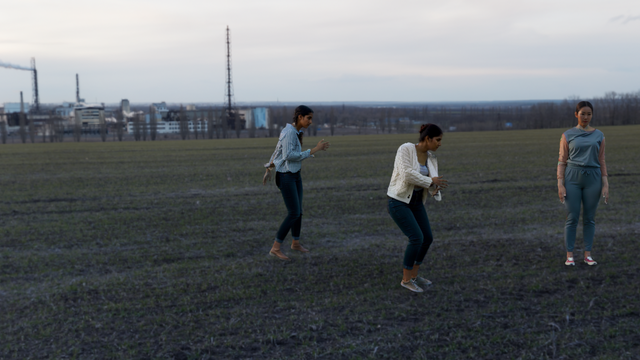}
        \end{subfigure}
    \end{minipage}%
}
   \caption{Bedlam vs. Gen-B with edges as the control signal.}
\label{fig:control_edges_example}
\end{figure*}

\begin{figure*}
\centering
\resizebox{0.9\linewidth}{!}{%
    \begin{minipage}{\linewidth}
        \centering
        \begin{subfigure}[t]{0.48\linewidth}
            \includegraphics[width=\linewidth]{img/ori_bedlam/_is_cluster_fast_scratch_hcuevas_20221010_3-10_500_batch01hand_zoom_suburb_dseq_000000_seq_000000_0080.png}
        \end{subfigure}
        \begin{subfigure}[t]{0.48\linewidth}
            \includegraphics[width=\linewidth]{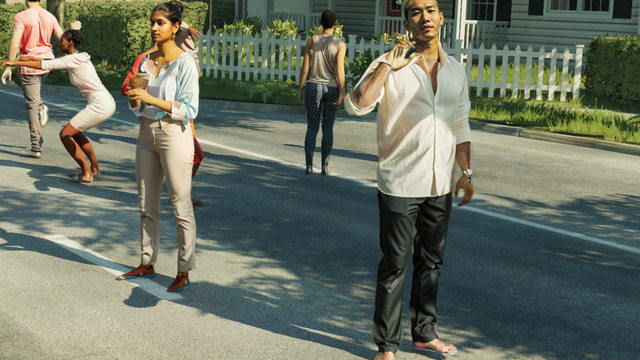}
        \end{subfigure}
        \begin{subfigure}[t]{0.48\linewidth}
            \includegraphics[width=\linewidth]{img/ori_bedlam/_is_cluster_fast_scratch_hcuevas_20221010_3_1000_batch01hand_6fpsseq_000021_seq_000021_0210.png}
        \end{subfigure}
        \begin{subfigure}[t]{0.48\linewidth}
            \includegraphics[width=\linewidth]{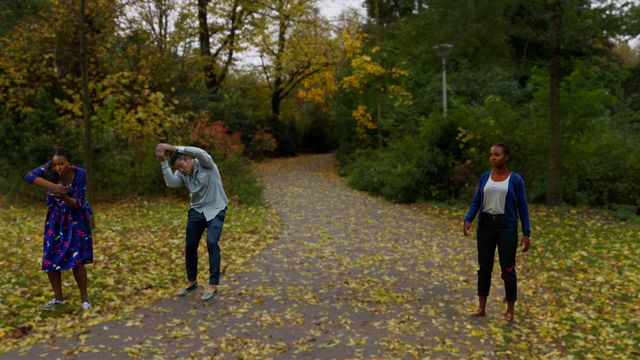}
        \end{subfigure}
        \begin{subfigure}[t]{0.48\linewidth}
            \includegraphics[width=\linewidth]{img/ori_bedlam/_is_cluster_fast_scratch_hcuevas_20221010_3_1000_batch01hand_6fpsseq_000016_seq_000016_0115.png}
        \end{subfigure}
        \begin{subfigure}[t]{0.48\linewidth}
            \includegraphics[width=\linewidth]{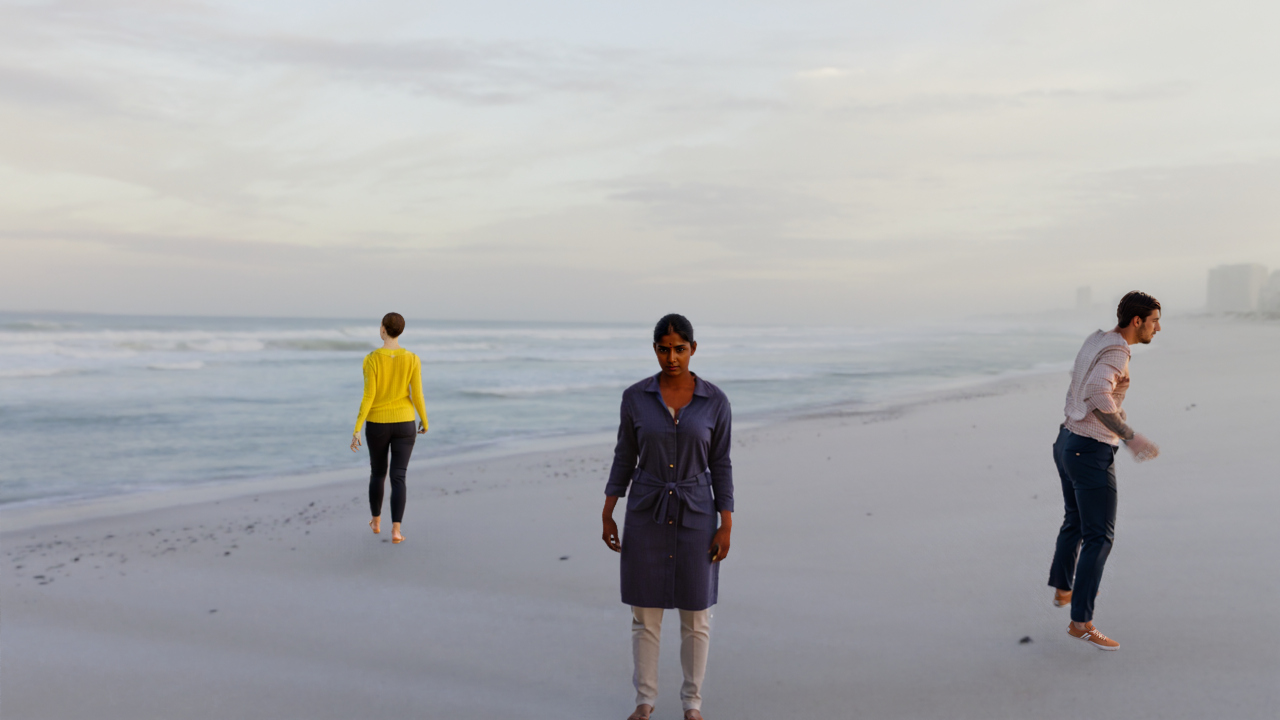}
        \end{subfigure}
        \begin{subfigure}[t]{0.48\linewidth}
            \includegraphics[width=\linewidth]{img/ori_bedlam/_is_cluster_fast_scratch_hcuevas_20221010_3_1000_batch01hand_6fpsseq_000022_seq_000022_0030.png}
        \end{subfigure}
        \begin{subfigure}[t]{0.48\linewidth}
            \includegraphics[width=\linewidth]{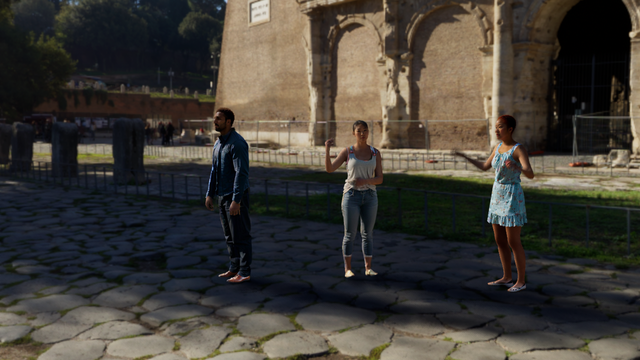}
        \end{subfigure}
        \begin{subfigure}[t]{0.48\linewidth}
            \includegraphics[width=\linewidth]{img/ori_bedlam/_is_cluster_fast_scratch_hcuevas_20221010_3_1000_batch01hand_6fpsseq_000033_seq_000033_0190.png}
        \end{subfigure}
        \begin{subfigure}[t]{0.48\linewidth}
            \includegraphics[width=\linewidth]{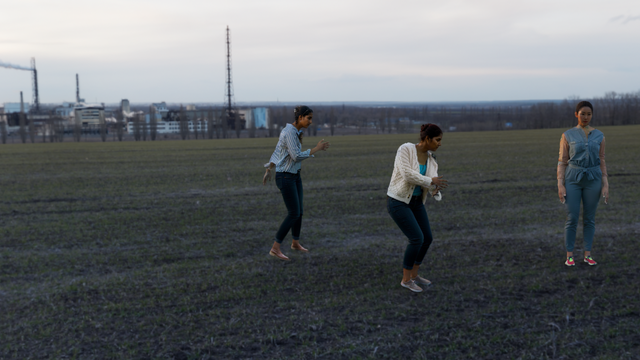}
        \end{subfigure}
    \end{minipage}%
}
   \caption{Bedlam vs. Gen-B with depth as the control signal.}
\label{fig:control_depth_example}
\end{figure*}

\begin{figure*}
\centering
\resizebox{0.9\linewidth}{!}{%
    \begin{minipage}{\linewidth}
        \centering
        \begin{subfigure}[t]{0.48\linewidth}
            \includegraphics[width=\linewidth]{img/ori_bedlam/_is_cluster_fast_scratch_hcuevas_20221010_3-10_500_batch01hand_zoom_suburb_dseq_000000_seq_000000_0080.png}
        \end{subfigure}
        \begin{subfigure}[t]{0.48\linewidth}
            \includegraphics[width=\linewidth]{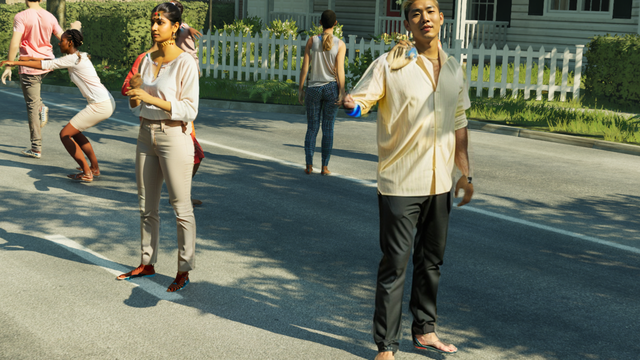}
        \end{subfigure}
        \begin{subfigure}[t]{0.48\linewidth}
            \includegraphics[width=\linewidth]{img/ori_bedlam/_is_cluster_fast_scratch_hcuevas_20221010_3_1000_batch01hand_6fpsseq_000021_seq_000021_0210.png}
        \end{subfigure}
        \begin{subfigure}[t]{0.48\linewidth}
            \includegraphics[width=\linewidth]{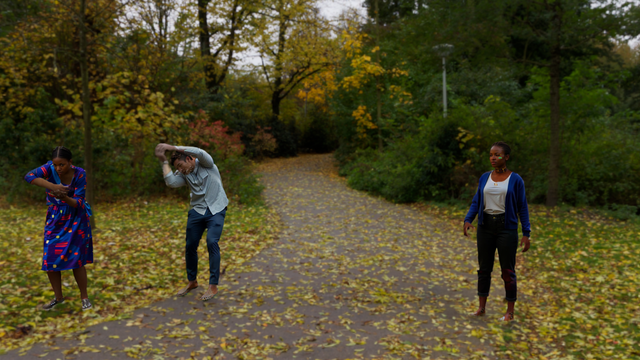}
        \end{subfigure}
        \begin{subfigure}[t]{0.48\linewidth}
            \includegraphics[width=\linewidth]{img/ori_bedlam/_is_cluster_fast_scratch_hcuevas_20221010_3_1000_batch01hand_6fpsseq_000016_seq_000016_0115.png}
        \end{subfigure}
        \begin{subfigure}[t]{0.48\linewidth}
            \includegraphics[width=\linewidth]{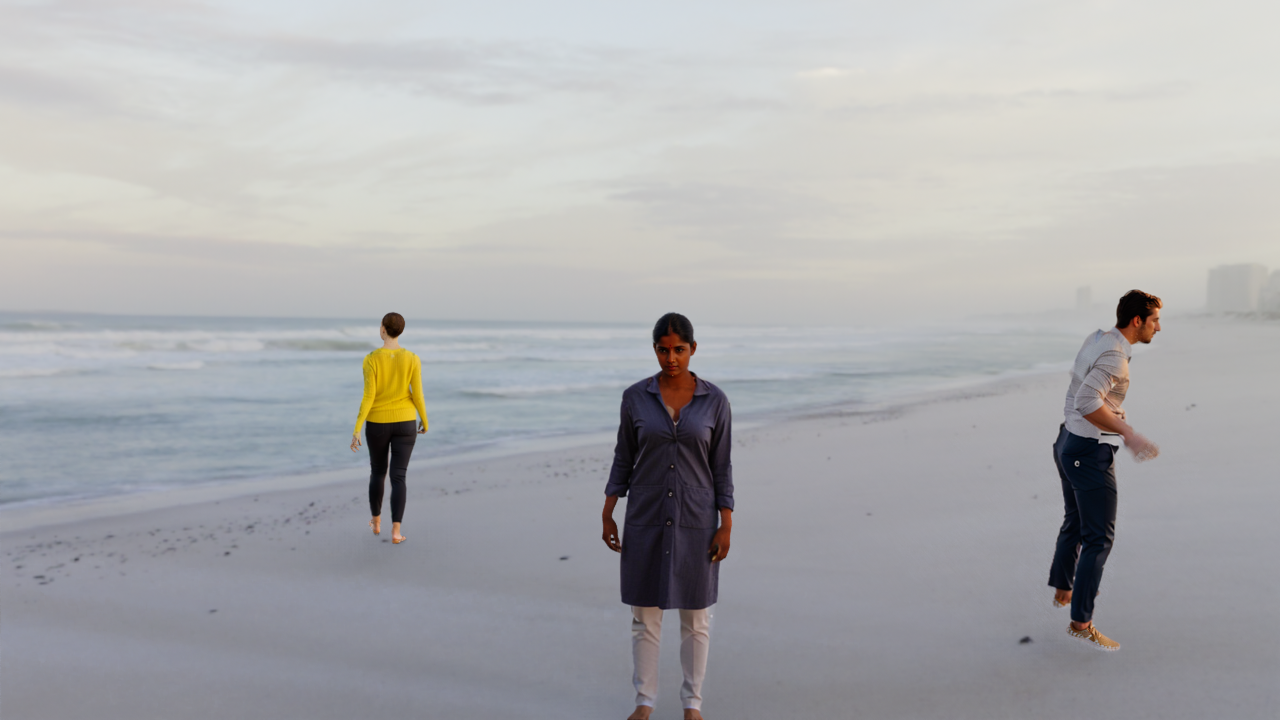}
        \end{subfigure}
        \begin{subfigure}[t]{0.48\linewidth}
            \includegraphics[width=\linewidth]{img/ori_bedlam/_is_cluster_fast_scratch_hcuevas_20221010_3_1000_batch01hand_6fpsseq_000022_seq_000022_0030.png}
        \end{subfigure}
        \begin{subfigure}[t]{0.48\linewidth}
            \includegraphics[width=\linewidth]{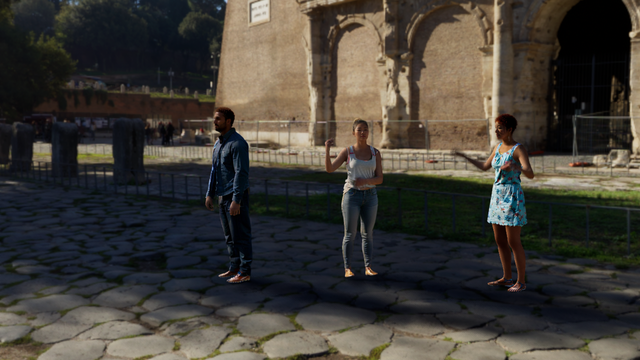}
        \end{subfigure}
        \begin{subfigure}[t]{0.48\linewidth}
            \includegraphics[width=\linewidth]{img/ori_bedlam/_is_cluster_fast_scratch_hcuevas_20221010_3_1000_batch01hand_6fpsseq_000033_seq_000033_0190.png}
        \end{subfigure}
        \begin{subfigure}[t]{0.48\linewidth}
            \includegraphics[width=\linewidth]{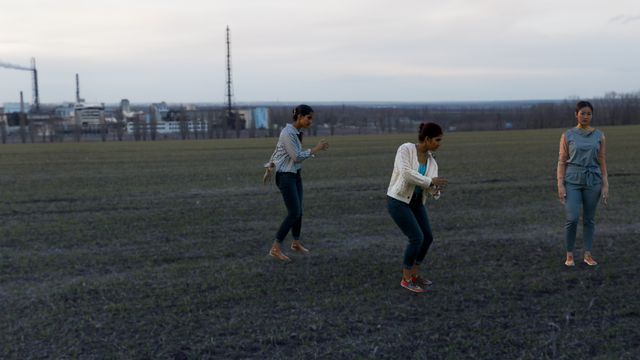}
        \end{subfigure}
    \end{minipage}%
}
   \caption{Bedlam vs. Gen-B with normals as the control signal.}
\label{fig:control_normals_example}
\end{figure*}

\begin{figure*}
\centering
\resizebox{0.9\linewidth}{!}{%
    \begin{minipage}{\linewidth}
        \centering
        \begin{subfigure}[t]{0.48\linewidth}
            \includegraphics[width=\linewidth]{img/ori_bedlam/_is_cluster_fast_scratch_hcuevas_20221010_3-10_500_batch01hand_zoom_suburb_dseq_000000_seq_000000_0080.png}
        \end{subfigure}
        \begin{subfigure}[t]{0.48\linewidth}
            \includegraphics[width=\linewidth]{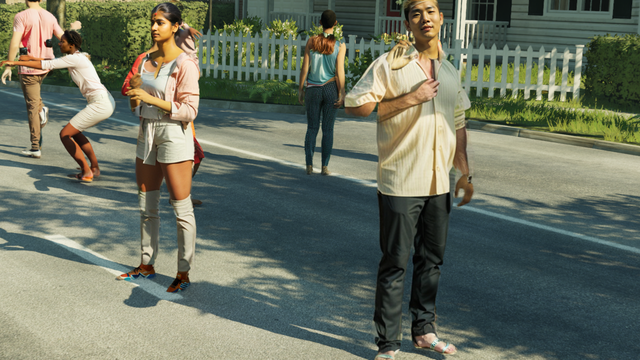}
        \end{subfigure}
        \begin{subfigure}[t]{0.48\linewidth}
            \includegraphics[width=\linewidth]{img/ori_bedlam/_is_cluster_fast_scratch_hcuevas_20221010_3_1000_batch01hand_6fpsseq_000021_seq_000021_0210.png}
        \end{subfigure}
        \begin{subfigure}[t]{0.48\linewidth}
            \includegraphics[width=\linewidth]{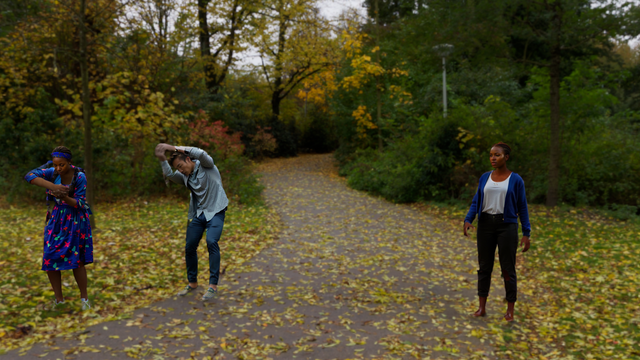}
        \end{subfigure}
        \begin{subfigure}[t]{0.48\linewidth}
            \includegraphics[width=\linewidth]{img/ori_bedlam/_is_cluster_fast_scratch_hcuevas_20221010_3_1000_batch01hand_6fpsseq_000016_seq_000016_0115.png}
        \end{subfigure}
        \begin{subfigure}[t]{0.48\linewidth}
            \includegraphics[width=\linewidth]{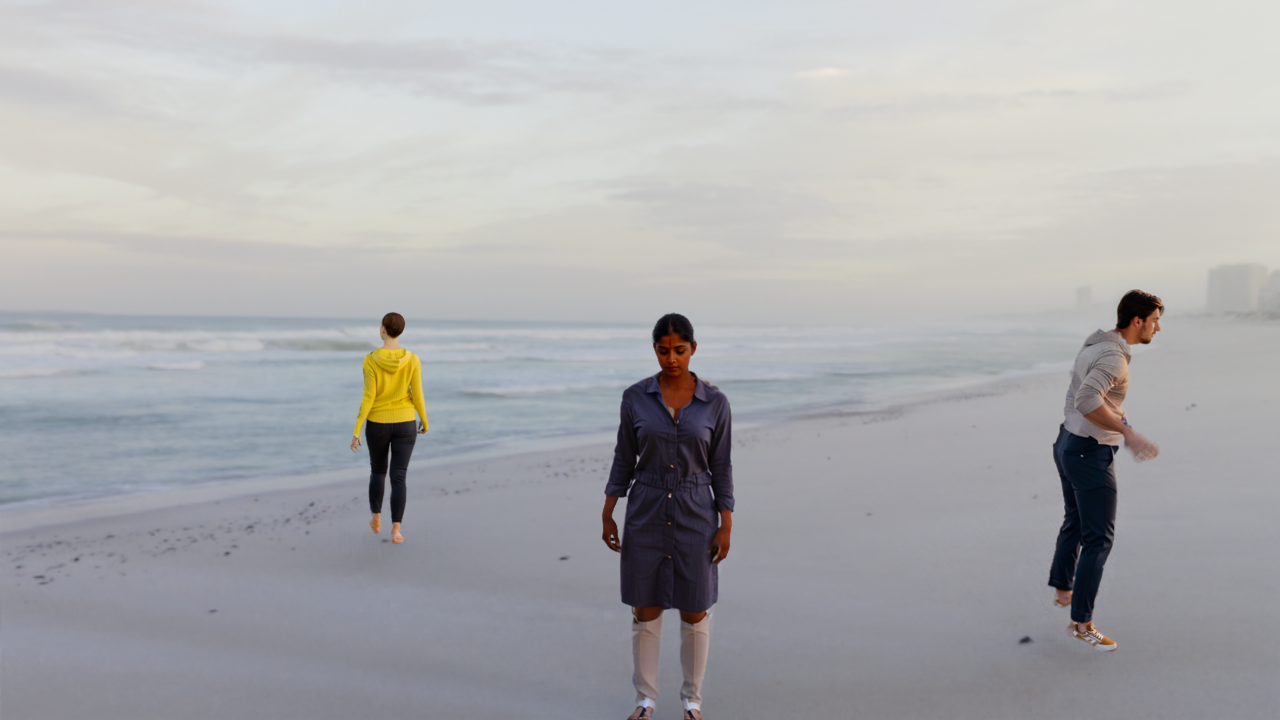}
        \end{subfigure}
        \begin{subfigure}[t]{0.48\linewidth}
            \includegraphics[width=\linewidth]{img/ori_bedlam/_is_cluster_fast_scratch_hcuevas_20221010_3_1000_batch01hand_6fpsseq_000022_seq_000022_0030.png}
        \end{subfigure}
        \begin{subfigure}[t]{0.48\linewidth}
            \includegraphics[width=\linewidth]{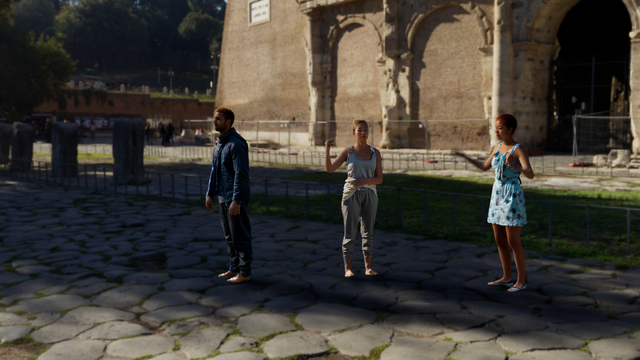}
        \end{subfigure}
        \begin{subfigure}[t]{0.48\linewidth}
            \includegraphics[width=\linewidth]{img/ori_bedlam/_is_cluster_fast_scratch_hcuevas_20221010_3_1000_batch01hand_6fpsseq_000033_seq_000033_0190.png}
        \end{subfigure}
        \begin{subfigure}[t]{0.48\linewidth}
            \includegraphics[width=\linewidth]{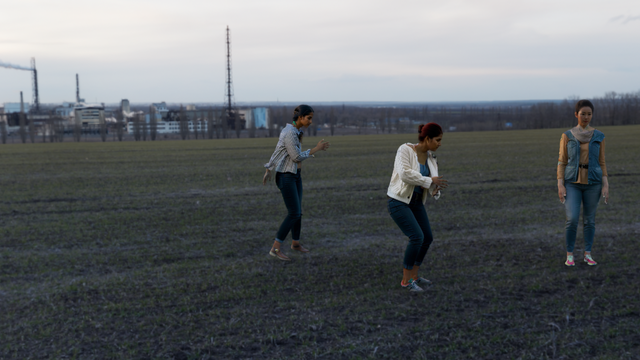}
        \end{subfigure}
    \end{minipage}%
}
   \caption{Bedlam vs. Gen-B with pose as the control signal.}
\label{fig:control_pose_example}
\end{figure*}

\begin{figure*}
\centering
\resizebox{0.9\linewidth}{!}{%
    \begin{minipage}{\linewidth}
        \centering
        \begin{subfigure}[t]{0.48\linewidth}
            \includegraphics[width=\linewidth]{img/ori_bedlam/_is_cluster_fast_scratch_hcuevas_20221010_3-10_500_batch01hand_zoom_suburb_dseq_000000_seq_000000_0080.png}
        \end{subfigure}
        \begin{subfigure}[t]{0.48\linewidth}
            \includegraphics[width=\linewidth]{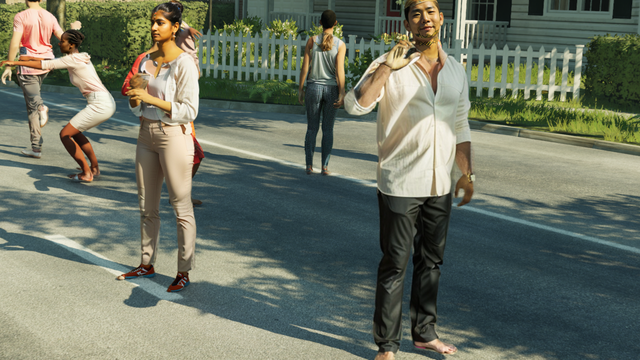}
        \end{subfigure}
        \begin{subfigure}[t]{0.48\linewidth}
            \includegraphics[width=\linewidth]{img/ori_bedlam/_is_cluster_fast_scratch_hcuevas_20221010_3_1000_batch01hand_6fpsseq_000021_seq_000021_0210.png}
        \end{subfigure}
        \begin{subfigure}[t]{0.48\linewidth}
            \includegraphics[width=\linewidth]{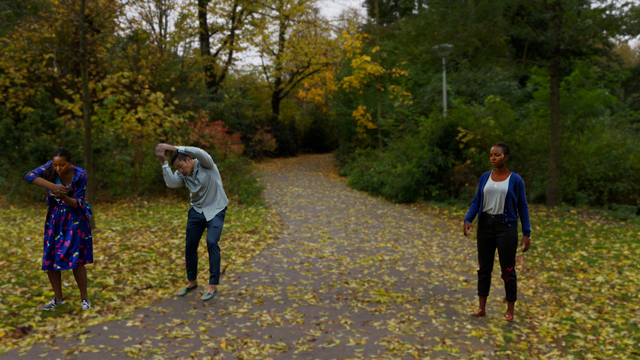}
        \end{subfigure}
        \begin{subfigure}[t]{0.48\linewidth}
            \includegraphics[width=\linewidth]{img/ori_bedlam/_is_cluster_fast_scratch_hcuevas_20221010_3_1000_batch01hand_6fpsseq_000016_seq_000016_0115.png}
        \end{subfigure}
        \begin{subfigure}[t]{0.48\linewidth}
            \includegraphics[width=\linewidth]{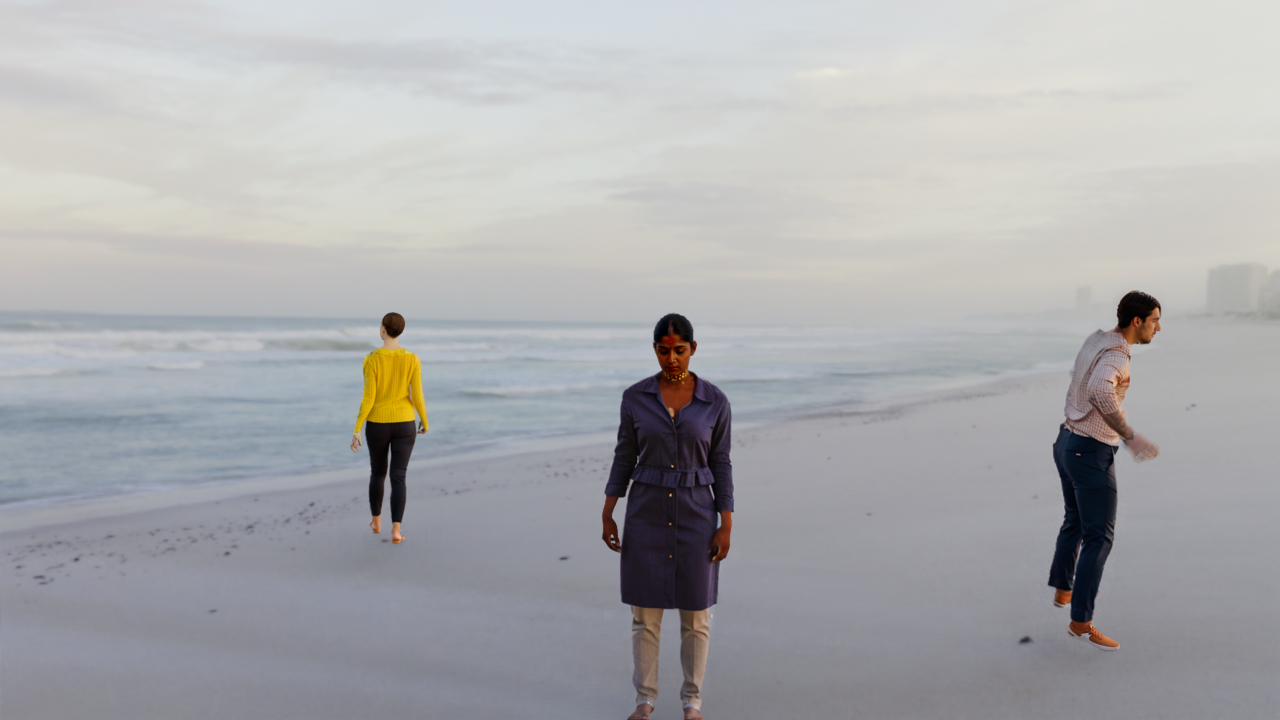}
        \end{subfigure}
        \begin{subfigure}[t]{0.48\linewidth}
            \includegraphics[width=\linewidth]{img/ori_bedlam/_is_cluster_fast_scratch_hcuevas_20221010_3_1000_batch01hand_6fpsseq_000022_seq_000022_0030.png}
        \end{subfigure}
        \begin{subfigure}[t]{0.48\linewidth}
            \includegraphics[width=\linewidth]{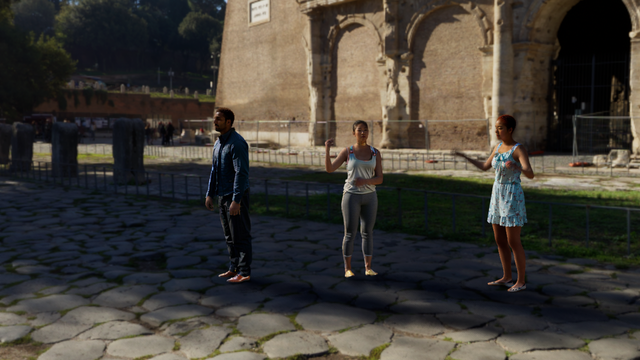}
        \end{subfigure}
        \begin{subfigure}[t]{0.48\linewidth}
            \includegraphics[width=\linewidth]{img/ori_bedlam/_is_cluster_fast_scratch_hcuevas_20221010_3_1000_batch01hand_6fpsseq_000033_seq_000033_0190.png}
        \end{subfigure}
        \begin{subfigure}[t]{0.48\linewidth}
            \includegraphics[width=\linewidth]{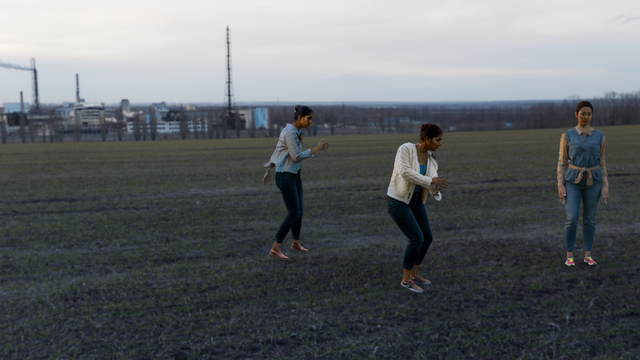}
        \end{subfigure}
    \end{minipage}%
}
   \caption{Bedlam vs. Gen-B with depth and pose as the control signals.}
\label{fig:control_depth_pose_example}
\end{figure*}

\begin{figure*}
\centering
\resizebox{0.9\linewidth}{!}{%
    \begin{minipage}{\linewidth}
        \centering
        \begin{subfigure}[t]{0.48\linewidth}
            \includegraphics[width=\linewidth]{img/ori_bedlam/_is_cluster_fast_scratch_hcuevas_20221010_3-10_500_batch01hand_zoom_suburb_dseq_000000_seq_000000_0080.png}
        \end{subfigure}
        \begin{subfigure}[t]{0.48\linewidth}
            \includegraphics[width=\linewidth]{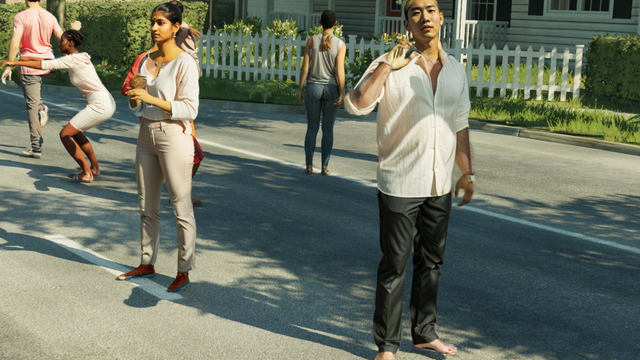}
        \end{subfigure}
        \begin{subfigure}[t]{0.48\linewidth}
            \includegraphics[width=\linewidth]{img/ori_bedlam/_is_cluster_fast_scratch_hcuevas_20221010_3_1000_batch01hand_6fpsseq_000021_seq_000021_0210.png}
        \end{subfigure}
        \begin{subfigure}[t]{0.48\linewidth}
            \includegraphics[width=\linewidth]{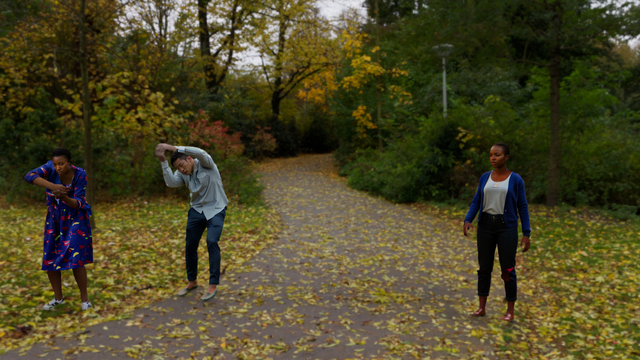}
        \end{subfigure}
        \begin{subfigure}[t]{0.48\linewidth}
            \includegraphics[width=\linewidth]{img/ori_bedlam/_is_cluster_fast_scratch_hcuevas_20221010_3_1000_batch01hand_6fpsseq_000016_seq_000016_0115.png}
        \end{subfigure}
        \begin{subfigure}[t]{0.48\linewidth}
            \includegraphics[width=\linewidth]{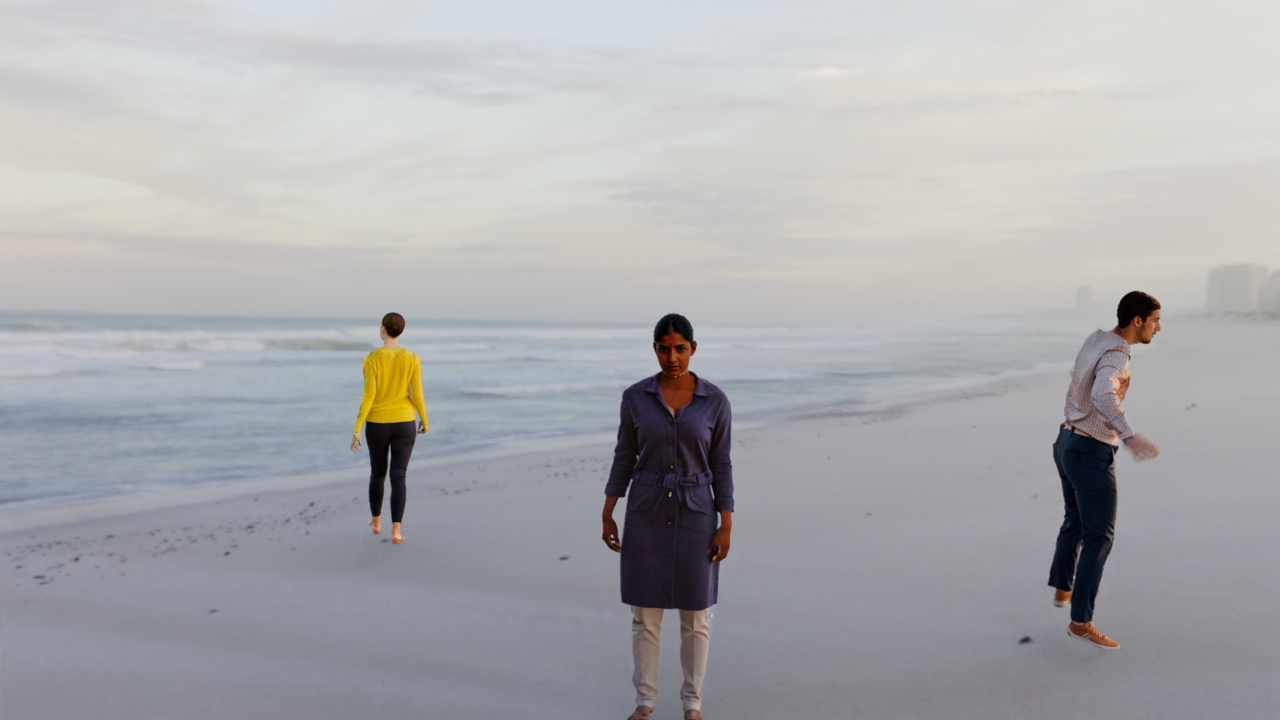}
        \end{subfigure}
        \begin{subfigure}[t]{0.48\linewidth}
            \includegraphics[width=\linewidth]{img/ori_bedlam/_is_cluster_fast_scratch_hcuevas_20221010_3_1000_batch01hand_6fpsseq_000022_seq_000022_0030.png}
        \end{subfigure}
        \begin{subfigure}[t]{0.48\linewidth}
            \includegraphics[width=\linewidth]{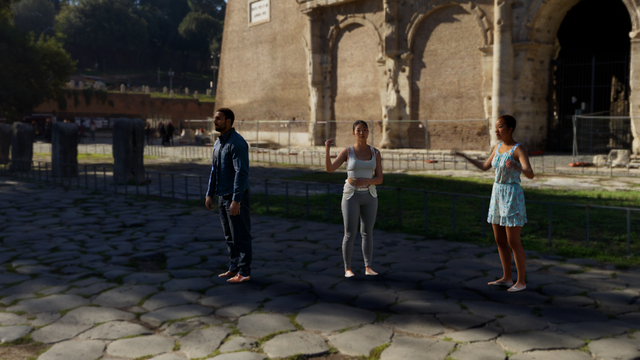}
        \end{subfigure}
        \begin{subfigure}[t]{0.48\linewidth}
            \includegraphics[width=\linewidth]{img/ori_bedlam/_is_cluster_fast_scratch_hcuevas_20221010_3_1000_batch01hand_6fpsseq_000033_seq_000033_0190.png}
        \end{subfigure}
        \begin{subfigure}[t]{0.48\linewidth}
            \includegraphics[width=\linewidth]{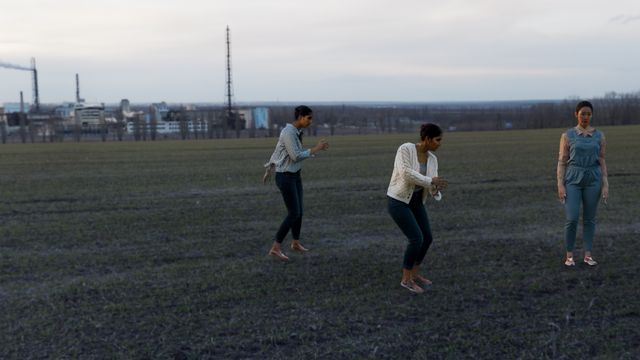}
        \end{subfigure}
    \end{minipage}%
}
   \caption{Bedlam vs. Gen-B with depth, pose, and edges as the control signals.}
\label{fig:control_depth_pose_edges_example}
\end{figure*}

\begin{figure*}
\centering
\resizebox{0.9\linewidth}{!}{%
    \begin{minipage}{\linewidth}
        \centering
        \begin{subfigure}[t]{0.48\linewidth}
            \includegraphics[width=\linewidth]{img/ori_bedlam/_is_cluster_fast_scratch_hcuevas_20221010_3-10_500_batch01hand_zoom_suburb_dseq_000000_seq_000000_0080.png}
        \end{subfigure}
        \begin{subfigure}[t]{0.48\linewidth}
            \includegraphics[width=\linewidth]{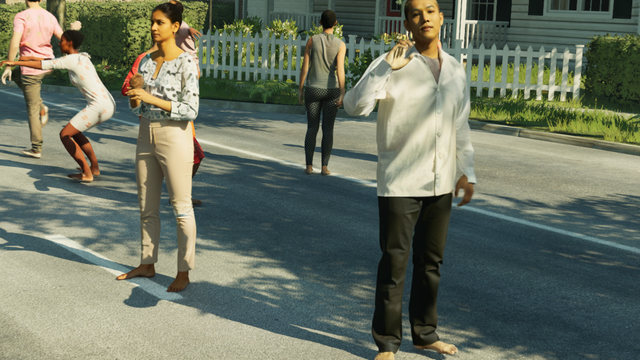}
        \end{subfigure}
        \begin{subfigure}[t]{0.48\linewidth}
            \includegraphics[width=\linewidth]{img/ori_bedlam/_is_cluster_fast_scratch_hcuevas_20221010_3_1000_batch01hand_6fpsseq_000021_seq_000021_0210.png}
        \end{subfigure}
        \begin{subfigure}[t]{0.48\linewidth}
            \includegraphics[width=\linewidth]{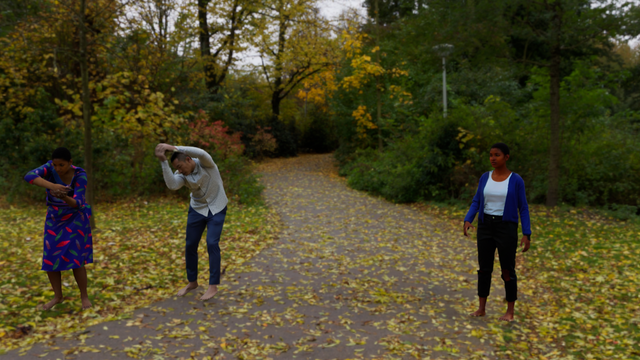}
        \end{subfigure}
        \begin{subfigure}[t]{0.48\linewidth}
            \includegraphics[width=\linewidth]{img/ori_bedlam/_is_cluster_fast_scratch_hcuevas_20221010_3_1000_batch01hand_6fpsseq_000016_seq_000016_0115.png}
        \end{subfigure}
        \begin{subfigure}[t]{0.48\linewidth}
            \includegraphics[width=\linewidth]{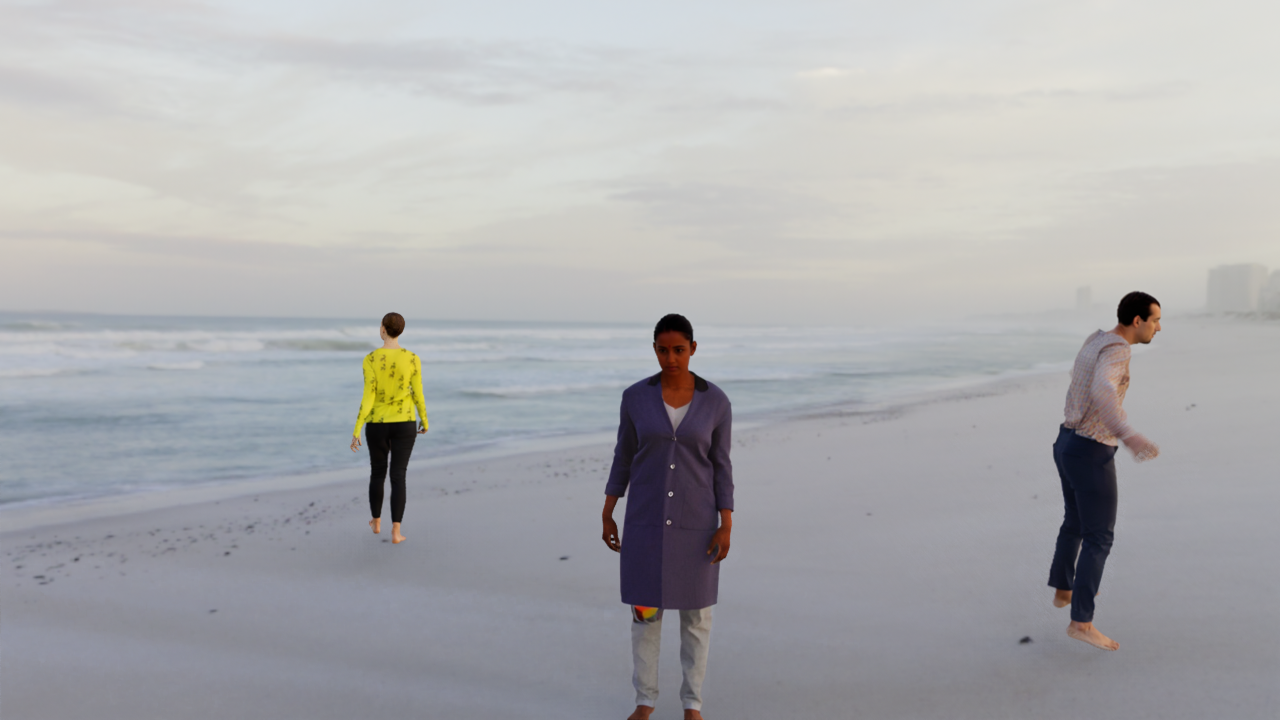}
        \end{subfigure}
        \begin{subfigure}[t]{0.48\linewidth}
            \includegraphics[width=\linewidth]{img/ori_bedlam/_is_cluster_fast_scratch_hcuevas_20221010_3_1000_batch01hand_6fpsseq_000022_seq_000022_0030.png}
        \end{subfigure}
        \begin{subfigure}[t]{0.48\linewidth}
            \includegraphics[width=\linewidth]{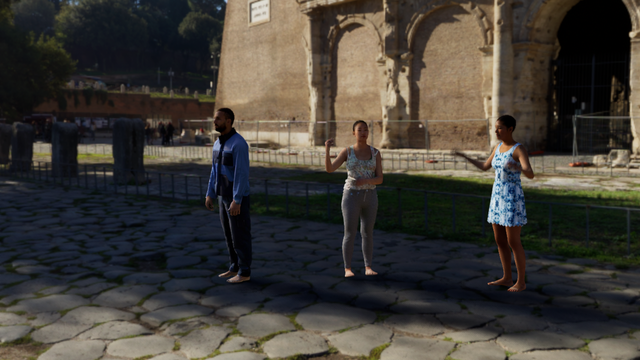}
        \end{subfigure}
        \begin{subfigure}[t]{0.48\linewidth}
            \includegraphics[width=\linewidth]{img/ori_bedlam/_is_cluster_fast_scratch_hcuevas_20221010_3_1000_batch01hand_6fpsseq_000033_seq_000033_0190.png}
        \end{subfigure}
        \begin{subfigure}[t]{0.48\linewidth}
            \includegraphics[width=\linewidth]{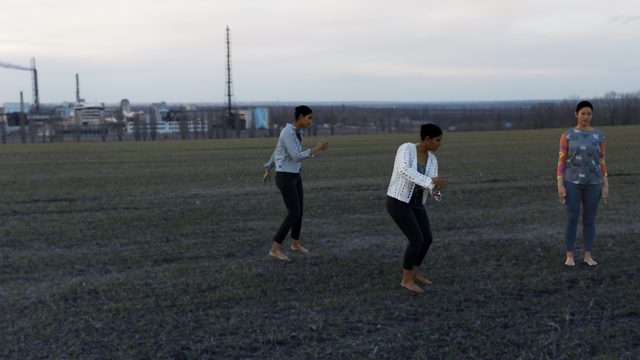}
        \end{subfigure}
    \end{minipage}%
}
   \caption{Bedlam vs. Gen-B generated with a noise level of 0.3.}
\label{fig:noise_03_example}
\end{figure*}

\begin{figure*}
\centering
\resizebox{0.9\linewidth}{!}{%
    \begin{minipage}{\linewidth}
        \centering
        \begin{subfigure}[t]{0.48\linewidth}
            \includegraphics[width=\linewidth]{img/ori_bedlam/_is_cluster_fast_scratch_hcuevas_20221010_3-10_500_batch01hand_zoom_suburb_dseq_000000_seq_000000_0080.png}
        \end{subfigure}
        \begin{subfigure}[t]{0.48\linewidth}
            \includegraphics[width=\linewidth]{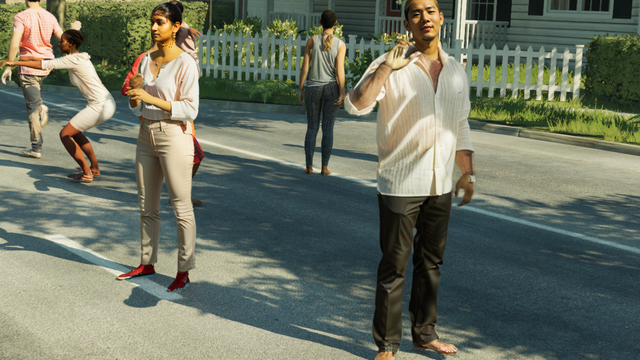}
        \end{subfigure}
        \begin{subfigure}[t]{0.48\linewidth}
            \includegraphics[width=\linewidth]{img/ori_bedlam/_is_cluster_fast_scratch_hcuevas_20221010_3_1000_batch01hand_6fpsseq_000021_seq_000021_0210.png}
        \end{subfigure}
        \begin{subfigure}[t]{0.48\linewidth}
            \includegraphics[width=\linewidth]{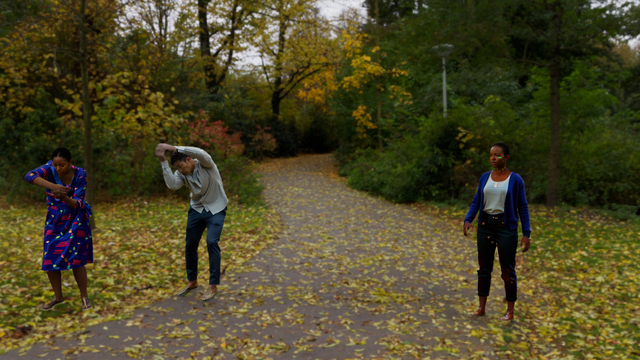}
        \end{subfigure}
        \begin{subfigure}[t]{0.48\linewidth}
            \includegraphics[width=\linewidth]{img/ori_bedlam/_is_cluster_fast_scratch_hcuevas_20221010_3_1000_batch01hand_6fpsseq_000016_seq_000016_0115.png}
        \end{subfigure}
        \begin{subfigure}[t]{0.48\linewidth}
            \includegraphics[width=\linewidth]{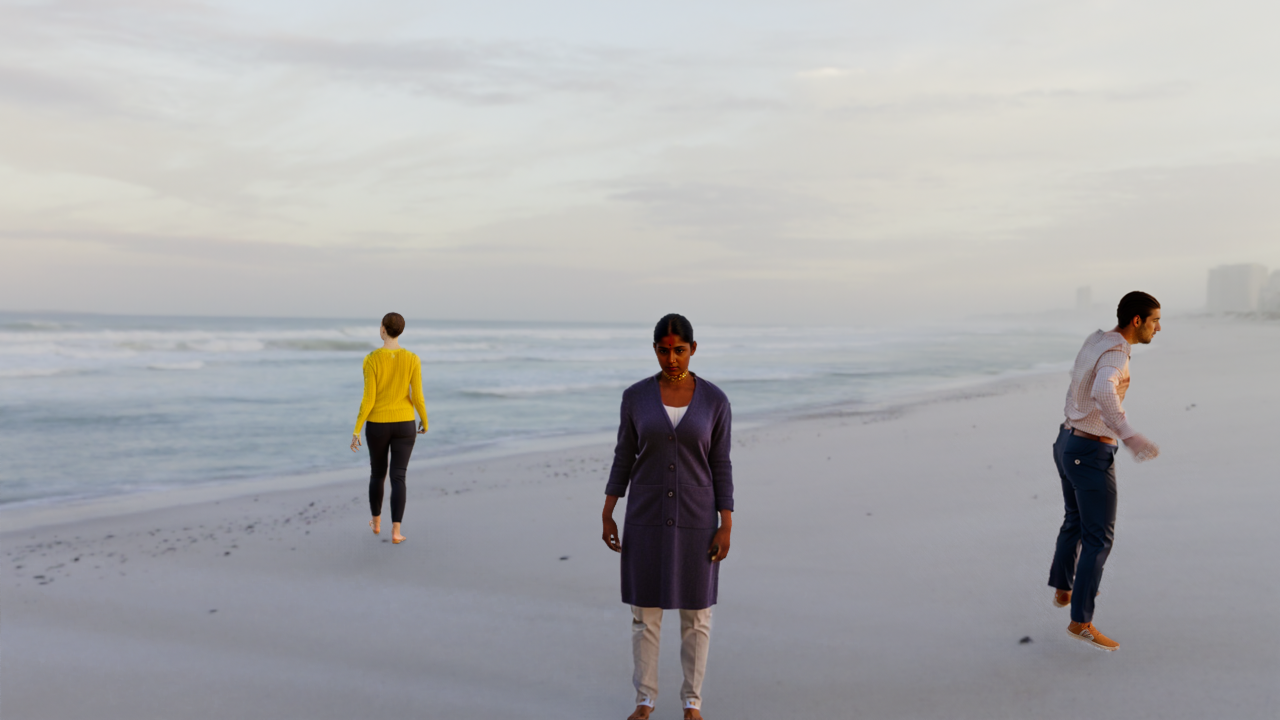}
        \end{subfigure}
        \begin{subfigure}[t]{0.48\linewidth}
            \includegraphics[width=\linewidth]{img/ori_bedlam/_is_cluster_fast_scratch_hcuevas_20221010_3_1000_batch01hand_6fpsseq_000022_seq_000022_0030.png}
        \end{subfigure}
        \begin{subfigure}[t]{0.48\linewidth}
            \includegraphics[width=\linewidth]{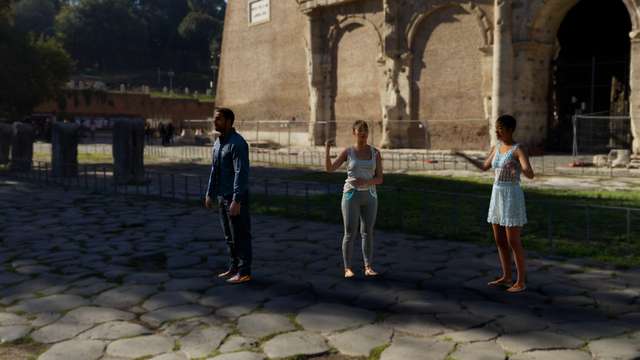}
        \end{subfigure}
        \begin{subfigure}[t]{0.48\linewidth}
            \includegraphics[width=\linewidth]{img/ori_bedlam/_is_cluster_fast_scratch_hcuevas_20221010_3_1000_batch01hand_6fpsseq_000033_seq_000033_0190.png}
        \end{subfigure}
        \begin{subfigure}[t]{0.48\linewidth}
            \includegraphics[width=\linewidth]{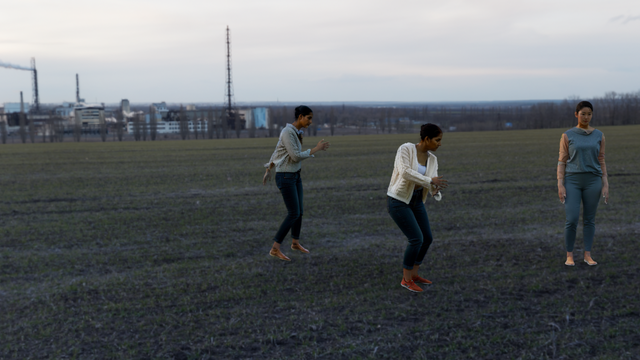}
        \end{subfigure}
    \end{minipage}%
}
   \caption{Bedlam vs. Gen-B generated with a noise level of 0.5 (or Gen-B with depth, pose, and edges as the control signals).}
\label{fig:noise_05_example}
\end{figure*}

\begin{figure*}
\centering
\resizebox{0.9\linewidth}{!}{%
    \begin{minipage}{\linewidth}
        \centering
        \begin{subfigure}[t]{0.48\linewidth}
            \includegraphics[width=\linewidth]{img/ori_bedlam/_is_cluster_fast_scratch_hcuevas_20221010_3-10_500_batch01hand_zoom_suburb_dseq_000000_seq_000000_0080.png}
        \end{subfigure}
        \begin{subfigure}[t]{0.48\linewidth}
            \includegraphics[width=\linewidth]{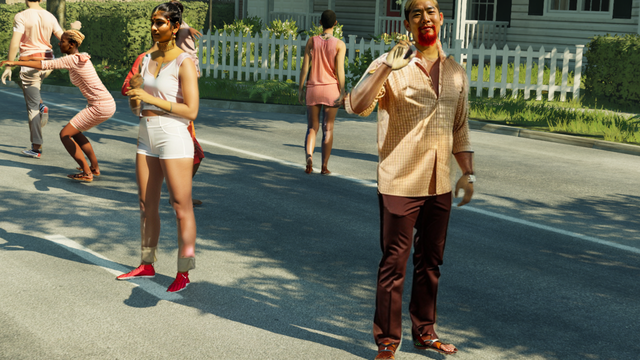}
        \end{subfigure}
        \begin{subfigure}[t]{0.48\linewidth}
            \includegraphics[width=\linewidth]{img/ori_bedlam/_is_cluster_fast_scratch_hcuevas_20221010_3_1000_batch01hand_6fpsseq_000021_seq_000021_0210.png}
        \end{subfigure}
        \begin{subfigure}[t]{0.48\linewidth}
            \includegraphics[width=\linewidth]{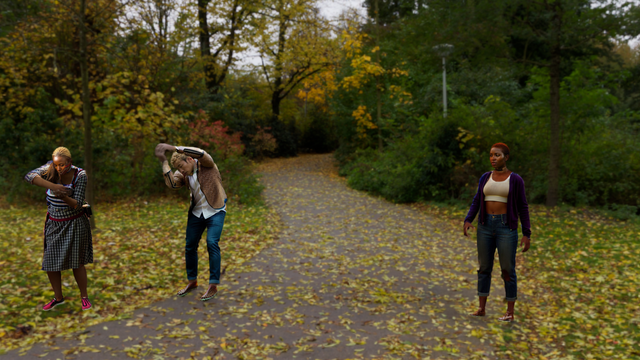}
        \end{subfigure}
        \begin{subfigure}[t]{0.48\linewidth}
            \includegraphics[width=\linewidth]{img/ori_bedlam/_is_cluster_fast_scratch_hcuevas_20221010_3_1000_batch01hand_6fpsseq_000016_seq_000016_0115.png}
        \end{subfigure}
        \begin{subfigure}[t]{0.48\linewidth}
            \includegraphics[width=\linewidth]{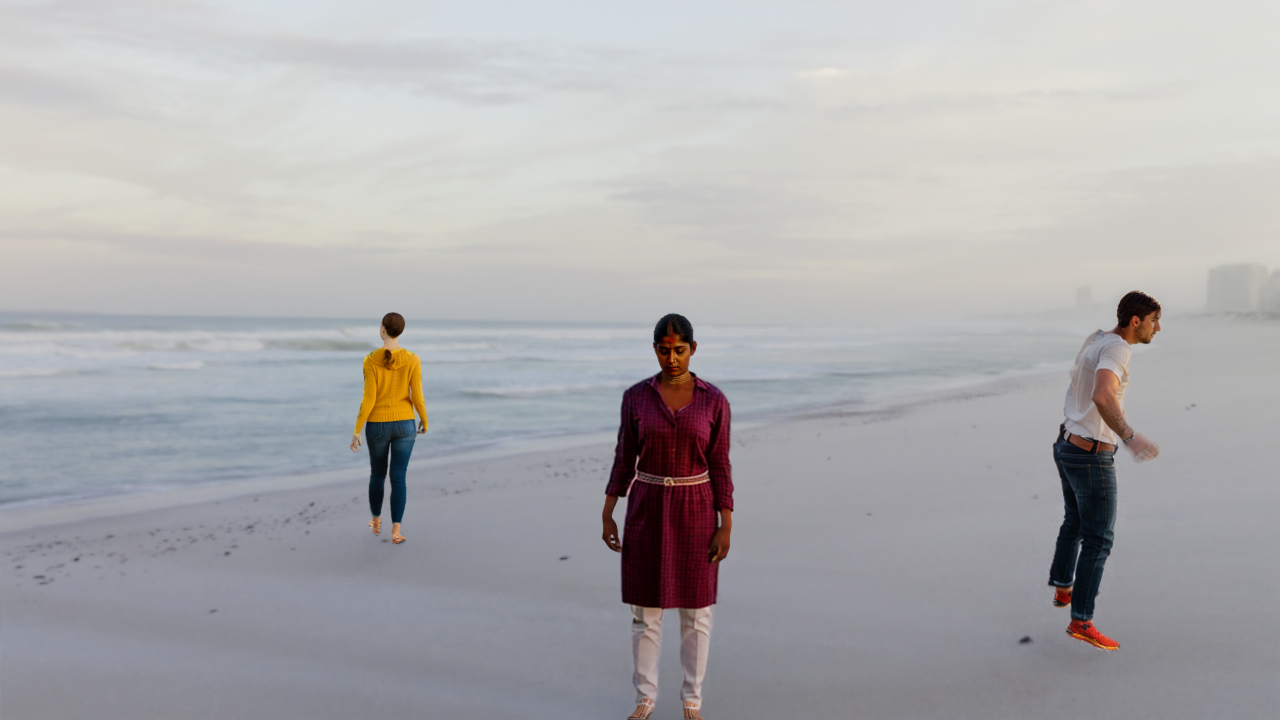}
        \end{subfigure}
        \begin{subfigure}[t]{0.48\linewidth}
            \includegraphics[width=\linewidth]{img/ori_bedlam/_is_cluster_fast_scratch_hcuevas_20221010_3_1000_batch01hand_6fpsseq_000022_seq_000022_0030.png}
        \end{subfigure}
        \begin{subfigure}[t]{0.48\linewidth}
            \includegraphics[width=\linewidth]{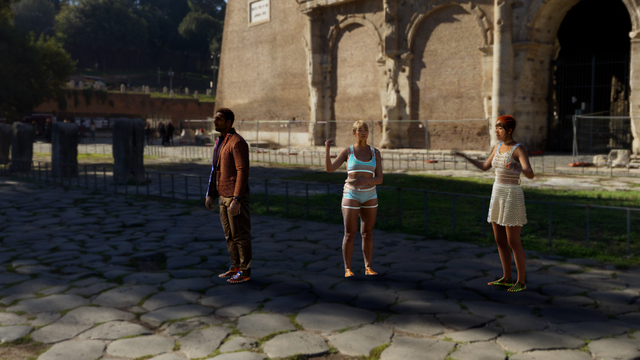}
        \end{subfigure}
        \begin{subfigure}[t]{0.48\linewidth}
            \includegraphics[width=\linewidth]{img/ori_bedlam/_is_cluster_fast_scratch_hcuevas_20221010_3_1000_batch01hand_6fpsseq_000033_seq_000033_0190.png}
        \end{subfigure}
        \begin{subfigure}[t]{0.48\linewidth}
            \includegraphics[width=\linewidth]{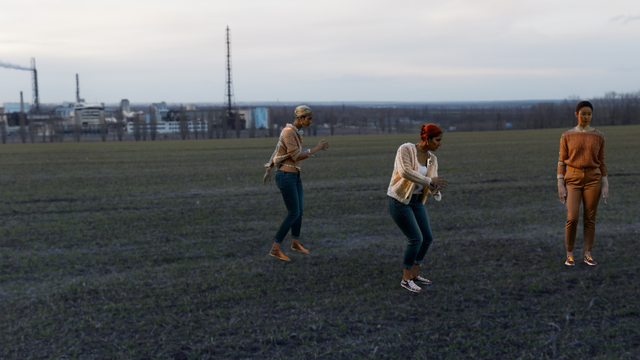}
        \end{subfigure}
    \end{minipage}%
}
   \caption{Bedlam vs. Gen-B generated with a noise level of 0.7.}
\label{fig:noise_07_example}
\end{figure*}

\begin{figure*}
\centering
\resizebox{0.9\linewidth}{!}{%
    \begin{minipage}{\linewidth}
        \centering
        \begin{subfigure}[t]{0.48\linewidth}
            \includegraphics[width=\linewidth]{img/ori_bedlam/_is_cluster_fast_scratch_hcuevas_20221010_3-10_500_batch01hand_zoom_suburb_dseq_000000_seq_000000_0080.png}
        \end{subfigure}
        \begin{subfigure}[t]{0.48\linewidth}
            \includegraphics[width=\linewidth]{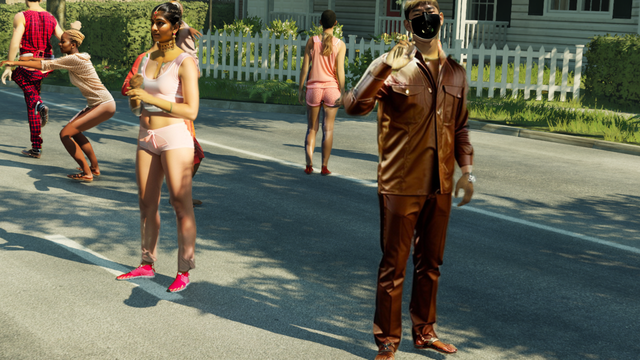}
        \end{subfigure}
        \begin{subfigure}[t]{0.48\linewidth}
            \includegraphics[width=\linewidth]{img/ori_bedlam/_is_cluster_fast_scratch_hcuevas_20221010_3_1000_batch01hand_6fpsseq_000021_seq_000021_0210.png}
        \end{subfigure}
        \begin{subfigure}[t]{0.48\linewidth}
            \includegraphics[width=\linewidth]{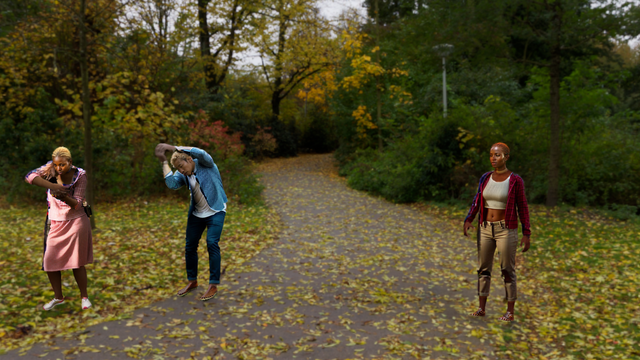}
        \end{subfigure}
        \begin{subfigure}[t]{0.48\linewidth}
            \includegraphics[width=\linewidth]{img/ori_bedlam/_is_cluster_fast_scratch_hcuevas_20221010_3_1000_batch01hand_6fpsseq_000016_seq_000016_0115.png}
        \end{subfigure}
        \begin{subfigure}[t]{0.48\linewidth}
            \includegraphics[width=\linewidth]{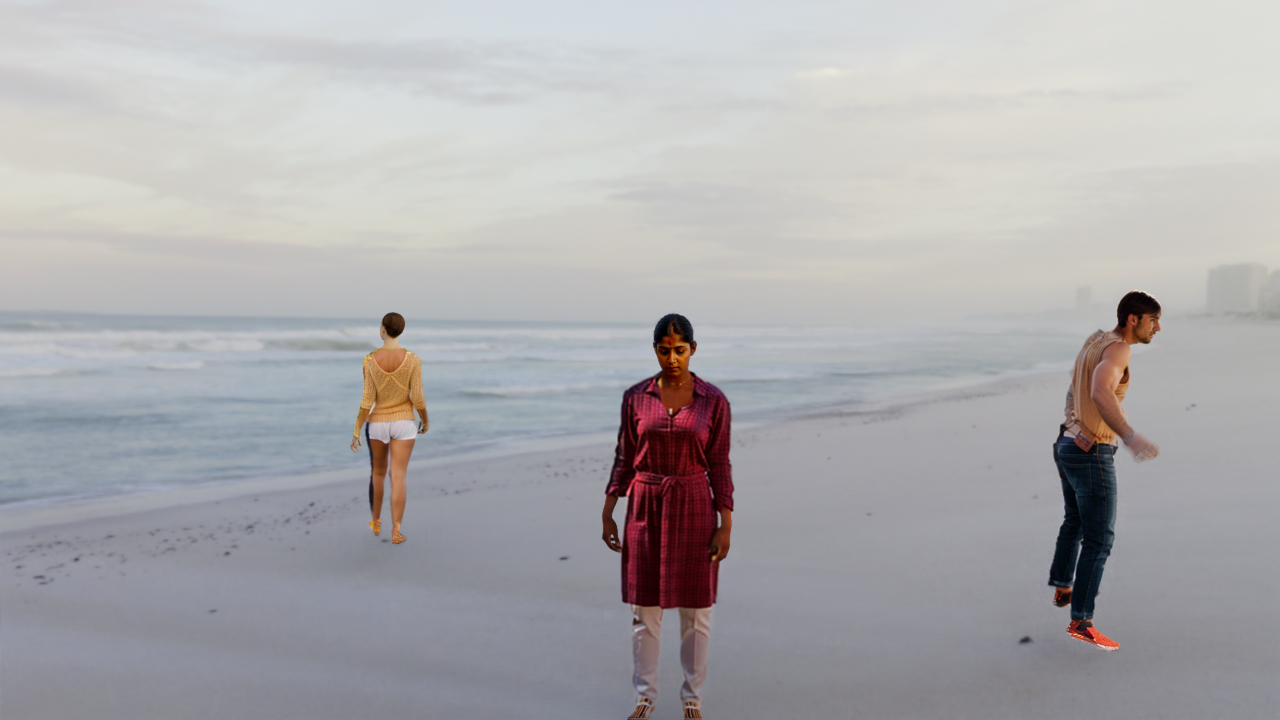}
        \end{subfigure}
        \begin{subfigure}[t]{0.48\linewidth}
            \includegraphics[width=\linewidth]{img/ori_bedlam/_is_cluster_fast_scratch_hcuevas_20221010_3_1000_batch01hand_6fpsseq_000022_seq_000022_0030.png}
        \end{subfigure}
        \begin{subfigure}[t]{0.48\linewidth}
            \includegraphics[width=\linewidth]{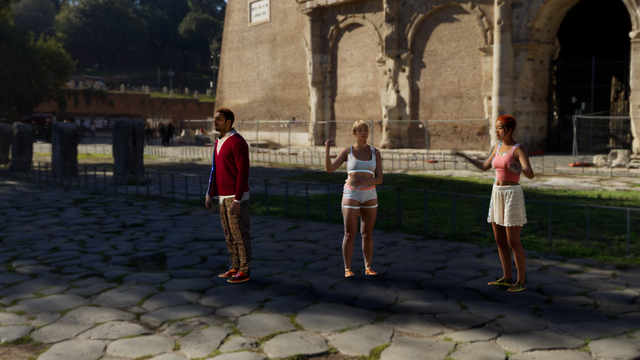}
        \end{subfigure}
        \begin{subfigure}[t]{0.48\linewidth}
            \includegraphics[width=\linewidth]{img/ori_bedlam/_is_cluster_fast_scratch_hcuevas_20221010_3_1000_batch01hand_6fpsseq_000033_seq_000033_0190.png}
        \end{subfigure}
        \begin{subfigure}[t]{0.48\linewidth}
            \includegraphics[width=\linewidth]{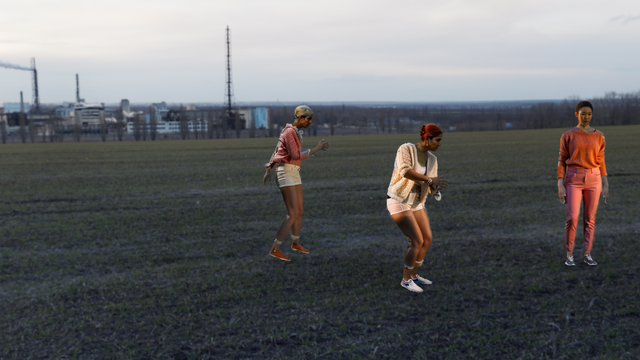}
        \end{subfigure}
    \end{minipage}%
}
   \caption{Bedlam vs. Gen-B generated with a noise level of 0.9.}
\label{fig:noise_09_example}
\end{figure*}

In \cref{fig:bedlam_gen_examples},  we show a side-by-side comparison between Gen-B images and the original BEDLAM images. In \cref{fig:control_edges_example,fig:control_depth_example,fig:control_normals_example,fig:control_pose_example,fig:control_depth_pose_example,fig:control_depth_pose_edges_example}, we show the same comparison, but for the different control signal outputs. Finally, in \cref{fig:noise_03_example,fig:noise_05_example,fig:noise_07_example,fig:noise_09_example}, we compare the images generated at different noise levels. Please refer to the attached videos, \texttt{gen\_bedlam\_comp.mp4}, \texttt{control\_videos}, and \texttt{noise\_videos}, for a better visual comparison.


\end{document}